\documentclass[ nonblindrev]{informs3}

\OneAndAHalfSpacedXI 

\usepackage{endnotes}
\let\footnote=\endnote
\let\enotesize=\normalsize
\def\notesname{Endnotes}%
\def\makeenmark{\hbox to1.275em{\theenmark.\enskip\hss}}
\def\enoteformat{\rightskip0pt\leftskip0pt\parindent=1.275em
  \leavevmode\llap{\makeenmark}}

\usepackage[colorlinks=true, linkcolor=blue, citecolor=blue, urlcolor=blue]{hyperref}
\providecommand{\online}[1]{Section~\ref{#1} of the e-companion}

\newcommand{\ELP}{\textup{ELP}}
\newcommand{\ALP}[1]{\textup{ALP(\ensuremath{#1})}}
\newcommand{\RLP}[2]{\textup{RLP(\ensuremath{#1},\ensuremath{#2})}}
\newcommand{\DUAL}[1]{\textup{Dual(\ensuremath{#1})}}
\newcommand{\RDUAL}[2]{\textup{R-Dual(\ensuremath{#1}, \ensuremath{#2})}}

\newcommand{\nmax}{n_{\max}}                 

\usepackage{amssymb}
\usepackage{bibunits}
\usepackage{natbib}
 \bibpunct[, ]{(}{)}{,}{a}{}{,}%
 \def\bibfont{\small}%
\usepackage{graphicx}
\usepackage{subcaption} 
\usepackage{multirow}
\usepackage{rotating}
\usepackage{soul}
\usepackage{amssymb} 
\usepackage{xcolor}
\usepackage{mathrsfs}

\usepackage[most]{tcolorbox}

\usepackage{subcaption}

\usepackage[ruled,linesnumbered,noend]{algorithm2e}
\usepackage[section]{placeins}

\graphicspath
{{Images/}}
\usepackage{ stmaryrd }
\usepackage{pdfpages}
\usepackage{graphicx}
\usepackage{algorithmic}
\usepackage{subcaption}
\usepackage{array} 
\usepackage{enumitem} 
\usepackage{xcolor}
\usepackage{soul} 
\sethlcolor{green!20!white} 

\usepackage{caption}    
\usepackage{float}      
\TheoremsNumberedThrough     
\ECRepeatTheorems

\EquationsNumberedThrough    

\MANUSCRIPTNO{}
\makeatletter
\newcommand\EC@renumber[1]{%
  \@ifundefined{c@#1}{}{%
    \setcounter{#1}{0}%
    \expandafter\let\csname EC@old@#1\expandafter\endcsname\csname the#1\endcsname
    \expandafter\edef\csname the#1\endcsname{%
      EC.\expandafter\noexpand\csname EC@old@#1\endcsname}%
  }%
}
\def\ECHowTheorems{%
  \EC@renumber{theorem}\EC@renumber{property}\EC@renumber{lemma}%
  \EC@renumber{proposition}\EC@renumber{corollary}\EC@renumber{claim}%
  \EC@renumber{conjecture}\EC@renumber{hypothesis}\EC@renumber{assumption}%
  \EC@renumber{remark}\EC@renumber{example}\EC@renumber{problem}%
  \EC@renumber{definition}\EC@renumber{question}\EC@renumber{answer}%
  \EC@renumber{exercise}%
}
\makeatother

\begin{document}


\makeatletter
\@ifclasswith{informs3}{blindrev}
  {\newcommand{\myrunauthor}{Authors' names blinded for peer review}}
  {\newcommand{\myrunauthor}{Fathi and Lee}}
\makeatother

\RUNAUTHOR{\myrunauthor}


\RUNTITLE{ADP with Basis Function Learning}

\TITLE{\textbf{ Approximate Dynamic Programming with Basis Function Learning  }}


\ARTICLEAUTHORS{%
\AUTHOR{Mohammadreza Fathi, Chi-Guhn Lee}
\AFF{Dynamic Optimization and Reinforcement Learning (DORL) Lab, Department of Mechanical and Industrial Engineering, University of Toronto. \\ \EMAIL{m.fathi@mail.utoronto.ca}, \EMAIL{chiguhn.lee@utoronto.ca}}
}

\ABSTRACT{%
This work presents a novel approach for approximating the value function in Markov decision processes (MDPs). Our approach is grounded in the approximate linear programming (ALP) framework, wherein basis functions (BFs) play an important role. Traditionally, BFs have been pre-defined, randomly generated, or restricted to specific structures, which both limits the accuracy that can be achieved and leaves no mechanism for improving a given approximation. We instead learn BFs from the model itself, by repeatedly applying the Bellman operator to the current value estimate. We introduce the Bellman operator basis function (BOBF) learning method, and prove that each learned BF improves the approximation by a quantifiable amount and that the method converges to the optimal value function. Its shrunken counterpart (SBOBF) retains this guarantee while capping the number of BFs, and therefore the computational effort per iteration. For large-scale problems, where the number of constraints is prohibitive, we develop hybrid variants (HBOBF and SHBOBF) that combine constraint sampling with BF learning, and we show that the improvement and finite-termination guarantees are preserved under explicitly stated conditions. Numerical experiments on (continuous) perishable inventory and queueing control problems demonstrate that our BF learning techniques significantly outperform existing methods in the literature.

}%



\maketitle

%

\begin{bibunit}[informs2014]

\section{Introduction}\label{sec: Introduction}

Exact linear programming (ELP) provides a powerful framework for solving Markov decision process (MDP) problems (\cite{puterman_markov_1994}, Section 6.9). Its well-established theoretical properties have contributed to its extensive application in both operations research (\cite{adelman2008relaxations,ye2011simplex,wang2020randomized}) and artificial intelligence (\cite{wang2007dual, nachum2020reinforcement,kamoutsi2021efficient}). However, ELP becomes computationally intractable as the state and action spaces grow.

The approximate linear program (ALP) framework was introduced to improve the scalability of ELP by reducing the number of decision variables  (\cite{de_farias_linear_2003}). ALP has since been applied across diverse domains, including network revenue management (\cite{ke2019approximate}), real option management (\cite{nadarajah2015relaxations}), queueing control (\cite{samiedaluie2024approximate}), marketing analytics (\mbox{\cite{nadarajah2020network}}), and energy storage management (\cite{lin_revisiting_2020}). 

ALP employs a vector of basis functions (BFs) as the foundation of its approximation scheme. In most cases, BFs are predefined using domain knowledge or constructed by imposing specific structures, such as affine functions, as commonly seen in the literature. This limitation is not unique to the ALP framework. It is prevalent across the broader field of value function approximation (see \cite{learning2005basis, kveton2006learning}, and \cite{kretchmar1999using} for example). However, such predefined or parameterized BFs often result in poor approximation accuracy and substantial optimality gaps.

Despite the critical role of BFs in ALP, the development of high-quality BFs has received limited attention. Recently, \cite{pakiman2025selfguided} proposed a universal random approach for generating BFs, which is both cost-effective and easy to sample. However, their method requires a large number of BFs to guarantee convergence to the optimal value function, potentially exceeding the size of the state space in some cases. Moreover, the performance of the optimal policy derived from their approach is highly sensitive to the choice of a suitable state-relevance distribution, which determines the relative importance of each state in the approximation. In contrast, the convergence of our exact BF learning methods holds for any state-relevance distribution that is positive component-wise, with no simulation-based calibration required.

Our study is closely related to the broader class of value-based reinforcement learning (RL) algorithms, including Q-learning (\cite{watkins_q-learning_1992}), double Q-Learning (\cite{hasselt2010double}), actor-critic RL (\cite{barto2012neuronlike}), deep Q-networks (\cite{mnih2015human}), double Q-networks \cite{van2016deep}, and soft actor-critic RL (\cite{haarnoja2018soft}). Early RL methods typically offer convergence guarantees but suffer from limited scalability. In contrast, more recent approaches rely heavily on large-scale neural networks to achieve strong empirical performance, yet they often lack theoretical convergence guarantees.

Several studies address value function approximation through the lens of approximate dynamic programming (ADP). For instance, \cite{jiang2015approximate} proposed a method for approximating monotone value functions. Their approach estimates the value of sampled states and generalizes these estimates to unobserved states by exploiting the monotonicity of the value function. The method is proven to converge under certain conditions and achieves near-optimal solutions. \cite{haskell2019universal} introduced empirical value learning algorithms for continuous-state MDPs that approximate the Bellman operator using simulation-based samples. Their approach combines this with universal function approximators constructed from randomly selected basis functions and provides convergence guarantees. In another application, \cite{maxwell2010approximate} developed an ADP method for ambulance redeployment to maximize timely responses to emergency calls. This method approximates the value function using a low-dimensional parametric model, tunes parameters through simulation, and demonstrates significant improvements over static benchmarks in real-world settings.

BF generation for general MDP problems, beyond the ALP framework, has been explored through various methodologies, including directed graph Laplacian (\cite{johns_constructing_2007}), Krylov sub-spaces (\cite{petrik_analysis_2007}), neighborhood component analysis (\cite{keller_automatic_2006}), diffusion models (\cite{mahadevan2007proto}), and Bellman residual BFs (\cite{parr_analyzing_2007}). Our BF learning approach is most similar to \cite{parr_analyzing_2007}. However, instead of relying on the Bellman residual, we utilize the Bellman operator.

Another stream of research seeks to eliminate the need for defining high-quality BFs in ALP. For example, \cite{desai_approximate_2012} reformulated ALP to allow constraint violations, introducing the concept of a violation budget to control the weighted sum of violations. By adjusting this budget parameter, the performance of the learned policy can be experimentally improved. However, guaranteeing tight optimality bounds under their approach requires prior knowledge of the state-visit distribution associated with the optimal policy. Building on this idea, \cite{bhat2012non} integrated kernel methods into ALP, thereby removing the necessity of explicitly defining BFs. Their approach demonstrated promising results in the Tetris game. Nevertheless, similar to \cite{desai_approximate_2012}, it relies on an idealized sampling distribution to ensure near-optimality. Furthermore, their formulation results in a quadratic program, which is computationally more expensive than a linear programming problem.

In contrast, our approach learns the BFs of ALP from the model itself, by applying the Bellman operator to the current value estimate, and it does so in two regimes. When the entire state--action pair can be enforced, we prove that the learned approximation converges to the optimal value function $V^*$ and that the returned estimate is $\epsilon$-optimal upon termination. To the best of our knowledge, this is the first application-agnostic BF learning methodology for ALP in stochastic discounted MDPs carrying such a guarantee. When the state and action spaces are too large for that, we retain a sampled constraint set and prove that each learned BF still yields a certified improvement and that the procedure terminates after finitely many iterations. In both regimes, the distinguishing feature is the ability to improve a given approximation rather than merely to select one from a fixed span.

The remainder of this paper is organized as follows. Section \ref{sec: MDP formulation} formalizes the problem and introduces the underlying notation. Section \ref{sec: BOBF} presents our solution methods, Bellman operator basis functions (BOBFs) and Shrunken BOBF (SBOBF), along with their theoretical guarantees on convergence, performance, and optimality. Section \ref{sec: Constraint Sampling} extends these frameworks by incorporating constraint sampling, yielding the hybrid methods HBOBF and SHBOBF, and provides theoretical insights into their improvement, termination, and residual guarantees. Section \ref{sec: Numerical Studies} reports numerical experiments on perishable inventory and queueing control problems, and Section \ref{sec: Conclusion} concludes the paper. Conditions for extending the results to continuous and compact spaces, supplementary discussion, and the omitted proofs are collected in the online appendix.

\section{MDP Formulation} \label{sec: MDP formulation}

Consider an infinite-horizon MDP with real-valued state set $\mathcal{S}$ and real-valued action set $\mathcal{A}$. All theoretical results in this study are established for finite $\mathcal{S}$, with $|\mathcal{S}|=N$, and finite $\mathcal{A}$. Most of them extend to the case in which $\mathcal{S}$ and $\mathcal{A}$ are continuous and compact. \online{app: continuous} collects the conditions required for the continuous case. 

For a state-action pair $(s,a) \in (\mathcal{S}\times\mathcal{A})$, let $c(s,a)$ denote the immediate cost, which is assumed to be bounded and non-negative, and let $p(s'|s,a)$ represent the transition probability to state $s' \in \mathcal{S}$. A policy $\pi: \mathcal{S} \mapsto \mathcal{A}$ assigns an action $a\in \mathcal{A}$ to each state $s \in \mathcal{S}$. The expected discounted cost of following the policy $\pi$ is defined as \begin{align*}
\mathcal{V} ^\pi(s) := \mathbb{E}_p\left[\sum_{t=0}^\infty \gamma^t c(s_t,\pi(s_t)) \bigg|s_0 = s \right], \qquad \forall s \in \mathcal{S},
\end{align*}
where $\gamma \in (0,1) $ denotes the discount factor, $s_t$ is the state visited at time $t$ under policy $\pi$, and the expectation $\mathbb{E}_p$ is taken with respect to the transition probability $p(\cdot|s_t,\pi(s_t) )$ induced by $\pi$. The optimal value function $V^*$ is given by \[
V^* := \min_\pi\left\{ \mathcal{V}^\pi\right\},
\] where the minimization is performed component-wise. For each state $s \in \mathcal{S}$, $V^*(s)$ represents the long-run discounted cost of following an optimal policy $\pi^* := \argmin_\pi \left\{ \mathcal{V}^\pi\right\}$. It is well established that $V^*$ is unique (\citealp{puterman_markov_1994}, Section 6.2).

For any value estimate $v \in \mathbb{R}^N$ the Bellman operator $\mathcal{T}: \mathbb{R}^N \mapsto \mathbb{R}^N$ is defined as \begin{align}
(\mathcal{T}v)(s) := \min_{a \in \mathcal{A}} \left\{c(s,a) + \gamma \sum_{s'\in \mathcal{S}}p(s'|s,a)v(s') \right\}
\label{eq: Bellman Operator Definition}\end{align}  for all $s \in \mathcal{S}$. Similarly, for a policy $\pi$, the operator $\mathcal{T}_\pi: \mathbb{R}^N \mapsto \mathbb{R}^N$ is defined as \begin{align}
    (\mathcal{T}_\pi v)(s) := c(s,\pi(s)) + \gamma \sum_{s' \in \mathcal{S} }p(s'|s,\pi(s)) v(s'), \label{eq: Bellman operator for policy pi}
\end{align} for all $s \in \mathcal{S}$. Similarly, for a fixed state $s \in \mathcal{S}$, we sometimes use the notation $\mathcal{T}_a$ for the same operator defined in (\ref{eq: Bellman operator for policy pi}) with $\pi(s) = a$.

Throughout, the discount factor is held fixed at some $\gamma\in(0,1)$. This strict discounting makes $\mathcal{T}$ a $\gamma$-contraction with unique fixed point $V^*$, and guarantees that $(I-\gamma P_a)^{-1}=\sum_{t=0}^\infty\gamma^{t}P_a^{t}$ exists so that $\mathcal{V}^\pi$ is well defined. Some bounds in this study carry a factor $1/(1-\gamma)$. They do not degrade as $\gamma\uparrow1$ but become vacuous because $\mathcal{V}^\pi$ diverges. \online{oa: gamma} provides more details in this regard.

\subsection {Linear Program Framework}

The optimal value function $V^*$ can also be computed without explicitly minimizing $\mathcal{V}^\pi$ over all policies. One such method is to formulate $V^*$ as the solution to the \ELP{} given by
\begin{align*}
    \text{ELP}: \max_{V: \mathcal{S} \to \mathbb{R}} & \quad \nu^\top V \\
    s.t.& \quad V(s) \leq c(s,a) + \gamma \sum_{s' \in \mathcal{S}} p(s'|s,a) V(s'), \quad \forall (s,a) \in \mathcal{S}\times\mathcal{A},
\end{align*}
where $\nu$ denotes a normalized vector over the states, commonly referred to as the state-relevance weight vector. If $\nu$ has a full support on $\mathcal{S}$, the optimal solution to \ELP{} coincides with $V^*$ (\citealp{puterman_markov_1994}, Section 6.9). Despite its appealing properties, \ELP{} is computationally expensive due to the large number of constraints and decision variables.

ALP was introduced to improve the scalability of \ELP{} by reducing the number of decision variables. A \emph{basis function} (BF) is a bounded function $\phi:\mathcal{S}\to\mathbb{R}$, and a \emph{BF vector} is an ordered tuple $\Phi=[\phi_1,\phi_2,\dots,\phi_\ell]$ of such functions. For a weight vector $w\in\mathbb{R}^\ell$ we write $\Phi w:=\sum_{i=1}^{\ell}w_i\phi_i$, so that $(\Phi w)(s)=\sum_{i=1}^{\ell}w_i\phi_i(s)$ for all $s\in\mathcal{S}$. When $\mathcal{S}$ is finite, we identify each $\phi_i$ with a vector in $\mathbb{R}^{N}$ and $\Phi$ with the $N\times \ell$ matrix whose $i$-th column is $\phi_i$. We write $\mathbf{1}$ for the function that is identically one on $\mathcal{S}$, and set $\phi_1 = \mathbf{1}$. When the basis vector carries an iteration index we write $\ell_k$ for the number of BFs in $\Phi_k$. The linear combination $\Phi w$ serves as the approximation of the value function, and substituting $V$ with $\Phi w$ in \ELP{} yields ALP, defined as
\begin{align*}
    \text{ALP}(\Phi): \quad \max_{w \in \mathbb{R}^\ell} & \quad \nu^\top\Phi w\\
s.t.&\quad (\Phi w)(s)\leq c(s,a)+\gamma\sum_{s'\in \mathcal{S}}p(s'|s,a)(\Phi w)(s'), \qquad
    \forall (s,a) \in (\mathcal{S}\times\mathcal{A}),
\end{align*}
where $\ell$ denotes the number of BFs in $\Phi$. Comparing ALP with \ELP{} reveals that the number of decision variables decreases from $N$ in \ELP{} to $\ell$ in ALP. Furthermore, it has been shown that maximizing $\nu^\top\Phi w$ in \ALP{\Phi} minimizes $\|V^* - \Phi w\|_{1,\nu}$ (\citealp{de_farias_linear_2003}), where $\|\cdot\|_{1,\nu}$ denotes the $L_1$ norm weighted by $\nu$.

A primary challenge in \ALP{\Phi} is the choice of the BF vector $\Phi$, since the quality of the resulting value estimate is limited by $\|V^*-\Phi\tilde{w}\|_{1,\nu}$, where $\tilde{w}$ denotes an optimal solution to \ALP{\Phi}. Most studies fix $\Phi$ in advance (\citealp{samiedaluie2024approximate, abouee2025platelet, lin_revisiting_2020, adelman_dynamic_2007, de_farias_linear_2003}). Such a choice has two shortcomings: the approximation gap $\|V^*-\Phi\tilde{w}\|_{1,\nu}$ may be large, and because $\Phi$ is fixed once and for all, the formulation provides no mechanism for reducing it. Despite its significance, the construction of $\Phi$ has received limited attention. To the best of our knowledge, \citet{pakiman2025selfguided} is the only study that addresses it explicitly for general stochastic MDPs without relying on application-specific structure. Their method draws BFs at random and consequently requires a prohibitively large number of them to achieve a reasonable weighted distance $\|V^*-\Phi\tilde{w}\|_{1,\nu}$.

In this study we propose a BF learning mechanism that addresses both shortcomings. When the entire state--action pairs can be considered, i.e., when \ALP{\Phi} itself is solvable, the value estimate produced by our method converges to the optimal value function, so that $\|V^*-\Phi\tilde{w}\|_{1,\nu}\to0$. Section \ref{sec: BOBF} presents this mechanism and its theoretical guarantees. When the state and action spaces are too large for \ALP{\Phi} and only a sampled constraint set is retained, our method starts from an initial approximation and improves it over a predefined sample at every iteration. Section \ref{sec: Constraint Sampling} develops this case, in which the intractably large constraint set of \ALP{\Phi} is itself replaced by a sample. The ability to improve a given approximation is the central advantage of the proposed methodology.

Throughout, for any $v \in \mathbb{R}^{N}$, we write
\[
\|v\|_\infty := \max_{s \in \mathcal{S}} \big|v(s)\big|,
\qquad
\|v\|_{1,\nu} := \sum_{s\in\mathcal{S}} \nu(s)\,\big|v(s)\big| ,
\]
$(v)^{+}:=\max\{v,0\}$, and $(v)^{-}:=\max\{-v,0\}$ component-wise, so that
$v=(v)^{+}-(v)^{-}$ and $|v|=(v)^{+}+(v)^{-}$.

\section{Bellman Operator Basis Function (BOBF)} \label{sec: BOBF}
Suppose $\Phi_k$ is predefined vector of BFs, and $\tilde{V}_k = \Phi_k\tilde{w}_k$, where $\tilde{w}_k$ is an optimal solution to \ALP{\Phi_k}. By construction, $\tilde{V}_k$ minimizes $\|V^* - \tilde{V}_k\|_{1,\nu}$ while satisfying the constraint $\tilde{V}_k \leq \mathcal{T}\tilde{V}_k$ component-wise, where $\mathcal{T}$ denotes the Bellman operator as defined in (\ref{eq: Bellman Operator Definition}). We analyze the \textit{net approximation improvement} achieved by adding a new BF $\phi$ to $\Phi_k$, which is defined as \[ \Delta_\phi \nu^\top\tilde{V}_k := \|V^* - \tilde{V}_k\|_{1,\nu} - \|V^* - \tilde{V}_{k+1}\|_{1,\nu}, \]
where $\tilde{V}_{k+1} = \Phi_{k+1}\tilde{w}_{k+1}$, and $\tilde{w}_{k+1}$ is an optimal solution to \ALP{\Phi_{k+1}} with $\Phi_{k+1} = \left[\Phi_k, \phi\right]$. This metric quantifies the reduction in the weighted distance between the two approximated value functions, $\tilde{V}_{k}$ and $\tilde{V}_{k+1}$, to the fixed point $V^*$. We propose $\phi= \mathcal{T}\tilde{V}_k$ as such a BF and refer to $\phi$ as the Bellman operator basis function (BOBF). Before analyzing the net approximation improvement resulting from the addition of a BOBF, we first summarize the key properties of \ALP{\Phi} in Lemma \ref{lem: preliminaries}. Throughout the paper, a tilde ($\tilde{\cdot}$) denotes an \textit{optimal}, while a bar ($\bar{\cdot}$) indicates \textit{a feasible} solution, unless otherwise stated.

\begin{lemma}[Basic properties of ALP($\Phi$)]
    \label{lem: preliminaries}
    Suppose $\bar{w}_k$ is a feasible solution to \ALP{\Phi_k}, and define $\bar{V}_k=\Phi_k\bar{w}_k$. Then, the following properties hold.
    \begin{enumerate}[label=(\roman*)]
    \item $\bar{V}_k(s) \leq V^* (s) $ for all $s\in \mathcal{S}$.

    \item Let $\Phi_{k+1}$ be a BF vector with $\ell_{k+1}$ BFs whose last BF is $\mathcal{T}\bar{V}_k$. Then $\bar{w}_{k+1}=(0,0,\cdots,0,1)^\top\in\mathbb{R}^{\ell_{k+1}}$ is a feasible solution to \ALP{\Phi_{k+1}}, and $\Phi_{k+1}\bar{w}_{k+1}=\mathcal{T}\bar{V}_k$.

    \item $\|V^* - \bar{V}_k\|_{1,\nu} = \nu^\top V^* - \nu^\top \bar{V}_k$.

    \item $\|\mathcal{T}\bar{V}_k - \bar{V}_k\|_{1,\nu} = \nu^\top \mathcal{T}\bar{V}_k - \nu^\top \bar{V}_k$.

    \end{enumerate}

\end{lemma}

The proof is given in \online{oa: proof preliminaries}. All four parts rest on the single chain of inequalities generated by feasibility and the monotonicity of $\mathcal{T}$,
\begin{equation}
\bar{V}_k\leq \mathcal{T}\bar{V}_k\leq \mathcal{T}\mathcal{T}\bar{V}_k\leq\cdots\leq V^*.
\label{eq:fixedpointineq}
\end{equation}

Using case (iii) in Lemma \ref{lem: preliminaries}, we can readily show that the net approximation improvement resulting from adding a new BF is equal to the difference between the two optimal objective values, i.e., $\Delta_\phi \nu^\top\tilde{V}_k = \nu^\top\tilde{V}_{k+1} - \nu^\top\tilde{V}_k$. This identity does not apply to the constraint sampling scheme in Section \ref{sec: Constraint Sampling}, because $\tilde{V}_{k+1}$ need no longer satisfy $\tilde{V}_{k+1}\le V^*$ under sampling. It is important to note that $\Delta_\phi \nu^\top\tilde{V}_k$ is non-negative. This follows from the fact that, given $\Phi_{k+1}=[\Phi_k, \phi]$, $\bar{w}_{k+1} = (\tilde{w}_k,0)^\top$ is a feasible solution to \ALP{\Phi_{k+1}}. Therefore, the optimal objective value $\nu^\top \tilde{V}_{k+1}$ is greater than or equal to $\nu^\top \tilde{V}_k$.

\subsection{A Lower Bound for the Net Approximation Improvement}
In this section, we derive a lower bound for the net approximation improvement achieved by adding a BOBF, i.e., $\Delta_{\phi=\mathcal{T}\tilde{V}_k} \nu^\top\tilde{V}_k$. We then discuss the significance of this improvement. These results are based on the dual formulation of ALP and its associated properties.

Let $u_a \in \mathbb{R}^N$ denote the vector of dual variables corresponding to action $a \in \mathcal{A}$. Specifically, $u_a(s)$ represents the dual variable associated with the state-action pair $(s,a) \in (\mathcal{S}\times\mathcal{A})$. Similarly, let $c_a$ be the cost vector for the action $a \in \mathcal{A}$, where $c_a(s) := c(s,a)$ for all $(s,a) \in (\mathcal{S}\times\mathcal{A})$. Likewise, $P_a$ denotes the transition probability matrix for action $a \in \mathcal{A}$, with entries $[P_a]_{s,s'} := p(s'|s,a) $ for all $(s,a) \in (\mathcal{S}\times\mathcal{A})$ and $s' \in \mathcal{S}$. The dual of \ALP{\Phi} can be then expressed as in Property \ref{prop: dual}.

\begin{property} [Dual of the ALP]
    \label{prop: dual}
    The dual of \ALP{\Phi} is given by
    \begin{align*}
        \textup{Dual($\Phi$)}: \quad \min_{u\in \mathbb{R}_+^{|\mathcal{S}\times \mathcal{A}|}} & \quad z(u) := \sum_{a \in \mathcal{A}} u_a^\top c_a \\
        s.t. & \quad \sum_{a\in \mathcal{A}} u_a^\top \left(I-\gamma P_a \right ) \Phi = \nu^\top \Phi,
    \end{align*}
    where $\mathbb{R}_+$ denotes the set of non-negative real numbers.
\end{property}

\ALP{\Phi} and \DUAL{\Phi} inherit the well-established properties of primal--dual relationships in linear programming, which are formalized in Property \ref{prop: dual properties}.

\begin{property}[Primal--dual properties of ALP]
    \label{prop: dual properties}
    Let $\tilde{V} = \Phi\tilde{w}$, where $\tilde{w}$ is an optimal solution to \ALP{\Phi}, and let $\tilde{u}$ be an optimal solution to \DUAL{\Phi}. Then, the following hold.
    \begin{enumerate}
        \item \textbf{Strong duality}: $z(\tilde{u}) = \sum_{a \in \mathcal{A}} \tilde{u}_a^\top c_a = \nu^\top \tilde{V}$.
        \item \textbf{Complementary slackness}: $\tilde{u}_a(s)\left(c(s,a) -  \left[\left(I-\gamma P_a \right ) \tilde{V}\right]_s \right)= 0$ for all $(s, a) \in (\mathcal{S}\times\mathcal{A})$, where $\left[\left(I-\gamma P_a \right ) \tilde{V}\right]_s $ represents the $s$-th component of the vector $\left(I-\gamma P_a \right ) \tilde{V}$.
    \end{enumerate}
\end{property}

In this paper, $P_\pi$ denotes the transition probability matrix associated with policy $\pi$, i.e., $\left[P_\pi\right]_{s,s'} = p\left(s'|s, \pi(s)\right)$ for all $s, s' \in \mathcal{S}$. Likewise, for a policy $\pi$, we define the cost vector $c_\pi$ by $c_\pi (s) = c(s, \pi(s)) $ and the dual vector $u_\pi$ by $u_\pi(s) = u_{\pi(s)}(s)$ for all $s \in \mathcal{S}$, i.e., $u_\pi$ is a sub-vector of $u$. It is important to note that the optimal dual solution $\tilde{u}$ is typically sparse, i.e., $\tilde{u}_a(s) = 0$ for many state-action pairs $(s, a)$. This sparsity is closely linked to the greedy policy $\tilde{\pi}:= \argmin_{\pi}\{c_\pi+\gamma P_{\pi} \tilde{V}\}$, and it in turn reduces the strong duality and dual feasibility identities of Property \ref{prop: dual properties} to identities indexed by a single policy. Property \ref{prop: dual structure} collects these consequences.

\begin{definition} [Binding primal constraint]
\label{def: binding}
Let $\tilde{V}=\Phi\tilde{w}$ with $\tilde{w}$ an optimal solution to \ALP{\Phi}. A state $s\in\mathcal{S}$ is said to \emph{have a binding primal constraint} in \ALP{\Phi} if the constraint indexed by
$(s,a)$ holds with equality for at least one $a\in\mathcal{A}$, i.e., $\exists a \in \mathcal{A}$ such that $\tilde{V}(s) = c(s,a) + \gamma \sum_{s' \in \mathcal{S}}p(s'|s,a)\tilde{V}(s')$.
\end{definition}

\begin{property}[Dual structure of the ALP under a unique greedy action]
\label{prop: dual structure}
Let $\tilde{V}=\Phi\tilde{w}$, where $\tilde{w}$ is an optimal solution to \ALP{\Phi}, let
$\tilde{u}$ be an optimal solution to \DUAL{\Phi}, and define the greedy policy
$\tilde{\pi}:=\argmin_{\pi}\{c_\pi+\gamma P_{\pi}\tilde{V}\}$. If the greedy action
$\tilde{\pi}(s)$ is unique at every state $s$ with a binding primal constraint (Definition
\ref{def: binding}), then the following properties hold.
\begin{enumerate}
    \item \textbf{Greedy dual sparsity}: $\tilde{u}_a(s)=0$ for all $s\in\mathcal{S}$ and all
    $a\in\mathcal{A}\setminus\{\tilde{\pi}(s)\}$.
    \item \textbf{Strong duality}: $z(\tilde{u})=\sum_{a\in\mathcal{A}}\tilde{u}_a^\top c_a
    =\tilde{u}_{\tilde{\pi}}^\top c_{\tilde{\pi}}=\nu^\top\tilde{V}$.
    \item \textbf{Dual feasibility}: $\sum_{a\in\mathcal{A}}\tilde{u}_a^\top(I-\gamma P_a)\Phi
    =\tilde{u}_{\tilde{\pi}}^\top(I-\gamma P_{\tilde{\pi}})\Phi=\nu^\top\Phi$.
    \item \textbf{Dual-induced successor distribution}: if $\mathbf{1}$ is a column of $\Phi$,
    then $\tilde{u}_{\tilde{\pi}}^\top\mathbf{1}=\dfrac{1}{1-\gamma}$ and
    $
    \mu^\top:=(1-\gamma)\,\tilde{u}_{\tilde{\pi}}^\top P_{\tilde{\pi}}
    $
    is a normalized vector over $\mathcal{S}$.
\end{enumerate}
\end{property}

In Property \ref{prop: dual structure}, part 1 is complementary slackness: a state with a binding constraint can carry dual weight only at its unique greedy action, and a state with no binding constraint carries none. Parts 2 and 3 then follow by deleting the vanishing terms from Property \ref{prop: dual properties}. Part 4 applies the column-wise form of part 3 to $\phi=\mathbf{1}$ and uses $P_{\tilde{\pi}}\mathbf{1}=\mathbf{1}$, which gives $(1-\gamma)\tilde{u}_{\tilde{\pi}}^\top\mathbf{1}=\nu^\top\mathbf{1}=1$. The details are in \online{oa: dual exact}.

Having established the necessary dual properties, we now derive a lower bound for the net approximation improvement in Theorem  \ref{thm: major improvement}.

\begin{theorem} [Net improvement of a BOBF]
    \label{thm: major improvement}
    Let $\tilde{w}_k$ denote an optimal solution to \ALP{\Phi_k} with $\Phi_k = [\mathbf{1}, \phi_2, \dots, \phi_\ell ]$, and let $\tilde{w}_{k+1}$ be an optimal solution to \ALP{\Phi_{k+1}}, where $\Phi_{k+1}=[\Phi_k, \mathcal{T}\Phi_k\tilde{w}_k]$. Suppose $\tilde{u}^{k+1}$ is an optimal solution to \DUAL{\Phi_{k+1}}. Let $\tilde{V}_k=\Phi_k\tilde{w}_k$, $\tilde{V}_{k+1}=\Phi_{k+1}\tilde{w}_{k+1}$, and the greedy policy $\tilde{\pi}_{k+1}:= \argmin_{\pi}\{c_\pi+\gamma P_{\pi} \tilde{V}_{k+1}\} $. If the greedy action $\tilde{\pi}_{k+1}(s)$ is unique at every state $s$ with a binding primal constraint (Definition \ref{def: binding}), we have
    \begin{align}
     \Delta_{\phi=\mathcal{T}\tilde{V}_k} \nu^\top\tilde{V}_k  \geq \|\mathcal{T}\tilde{V}_k-\tilde{V}_k\|_{1,\nu} + \frac{\gamma}{1-\gamma} \|\mathcal{T}\tilde{V}_k - \tilde{V}_k\|_{1,\mu_{k+1}},  \label{eq: major improvement 1}
     \end{align}
     where $\mu_{k+1}^\top = (1-\gamma) \tilde{u}_{\tilde{\pi}_{k+1}}^\top P_{\tilde{\pi}_{k+1}}$ is a normalized vector given by Property \ref{prop: dual structure}(part 4). Here, the notation $\tilde{u}_{\tilde{\pi}_{k+1}}$ refers to $\tilde{u}^{k+1}_{\tilde{\pi}_{k+1}}$.
\end{theorem}

The proof of Theorem \ref{thm: major improvement} is provided in~\nameref{app:A}. The bound \eqref{eq: major improvement 1} certifies that the net approximation improvement by adding a BOBF is at least as large as the $\nu$--weighted Bellman error $\|\mathcal{T}\tilde{V}_k-\tilde{V}_k\|_{1,\nu}$, ensuring a monotone approximation improvement on the support of $\nu$. Moreover, the bound contains a second term, $\frac{\gamma}{1-\gamma}\|\mathcal{T}\tilde{V}_k-\tilde{V}_k\|_{1,\mu_{k+1}}$, which can amplify the improvement because it is scaled by the constant factor $\gamma/(1-\gamma)$ and is evaluated under a dual-indexed vector $\mu_{k+1}$. This vector ties the bound to the optimal dual solution of \DUAL{\Phi_{k+1}} and the greedy dynamics induced by $\tilde{\pi}_{k+1}$, thereby weighting the residual along successor states under the policy that is greedy optimal with respect to $\tilde{V}_{k+1}$. Under the uniqueness of the greedy action at any state with a binding primal constraint (Definition \ref{def: binding}), this suffices to conclude that adding a BOBF yields a strict net approximation improvement whenever $\|\mathcal{T}\tilde{V}_k-\tilde{V}_k\|_{1,\nu}>0$.

\subsection{BF Learning Methods}
\label{sec: BF learning exact}

In this section, we present our BF learning approach and show that it achieves optimality for
a general positive $\nu$ (see Algorithm \ref{alg: BOBBF}). Starting with a vector of BFs
$\Phi_k$, the method computes the value estimate $\tilde{V}_k=\Phi_k\tilde{w}_k$, where
$\tilde{w}_k$ is an optimal solution to \ALP{\Phi_k}. It then evaluates
$\mathcal{T}\tilde{V}_k$ and appends this new BF to the existing vector of BFs, i.e.,
$\Phi_{k+1}=[\Phi_k,\mathcal{T}\tilde{V}_k]$, before solving \ALP{\Phi_{k+1}}. This iterative
process continues until convergence.

Algorithm \ref{alg: BOBBF} offers two options. ``Option 1: BOBF'' retains every BF generated. ``Option 2'', which we call the shrunken Bellman operator basis function (SBOBF) learning method, caps the number of BFs retained in $\Phi$ at $\nmax$ and thereby caps the computational effort required to solve \ALP{\Phi} at each iteration. We retain the most recent BFs, as they are closest to the optimal value function. Although some BFs are discarded under Option 2, our numerical experiments show that its performance is comparable to, and in some cases exceeds, that of Option 1.

\begin{algorithm}
    \caption{BF learning with ``Option 1: BOBF'' and ``Option 2: SBOBF($\nmax$)''.}\label{alg: BOBBF}
    \textbf{Initialization}: Let $\Phi_0=[\textbf{1}]$. Initialize $\nu$ and $\epsilon$ (the optimality tolerance). Solve the \ALP{\Phi_0} problem and let $\tilde{V}_0=\Phi_0 \tilde{w}_0$ represent the approximated value function. Set the iteration counter $k=0$. Select either ``Option 1: BOBF'' or ``Option 2: SBOBF($\nmax$)'', where $\nmax$ is the maximum number of BFs allowed. \\
    \While{$\frac{1}{1-\gamma} \| \mathcal{T}\tilde{V}_k - \tilde{V}_k \|_\infty> \epsilon$ }{
        Compute $\phi = \mathcal{T}\tilde{V}_k.$\\
         Update the BF vector \textbf{$\Phi_{k+1}$} $=$ $\left[\Phi_k, \phi\right]$.\\
        \If{``Option 2: SBOBF($\nmax$)'' is selected}{
            Let $\ell_{k+1}$ be the number of BFs in $\Phi_{k+1}$, i.e., $\Phi_{k+1}=[\phi_1,\phi_2,\cdots,\phi_{\ell_{k+1}}]$.\\
            \If{$\ell_{k+1} > \nmax$}{
                Retain the most recent $\nmax$ BFs: \textbf{$\Phi_{k+1}$} $=$ $[\phi_{\ell_{k+1}-\nmax+1}, \cdots,\phi_{\ell_{k+1}}]$.\\
            }
        }
         Solve \ALP{\Phi_{k+1}} for an optimal solution $\tilde{w}_{k+1}$, and set $\tilde{V}_{k+1}=\Phi_{k+1}\tilde{w}_{k+1} $.\\
        Increment the iteration counter: $k \leftarrow k+1 $.
    }
    \Return{$\tilde{V}_k$}\\
\end{algorithm}

Lemma \ref{lem: epsilon optimal} establishes a connection between an $\epsilon$-optimal value estimate and the Bellman error measured in the maximum norm.

\begin{lemma} [Bellman-residual optimality certificate]
\label{lem: epsilon optimal}
For any value estimate $\bar{V}\in\mathbb{R}^{N}$,
\begin{equation}
\|V^*-\bar{V}\|_{1,\nu}\ \leq\ \|V^*-\bar{V}\|_{\infty}
\ \leq\ \frac{1}{1-\gamma}\,\|\mathcal{T}\bar{V}-\bar{V}\|_\infty .
\label{eq: Pessimistic error bound}
\end{equation}
In particular, if $\frac{1}{1-\gamma}\|\mathcal{T}\bar{V}-\bar{V}\|_\infty\le\epsilon$, then
$\bar{V}$ is $\epsilon$-optimal with respect to $V^*$ and $\|V^*-\bar{V}\|_{1,\nu}\le\epsilon$.
\end{lemma}

The proof of Lemma \ref{lem: epsilon optimal} combines the contraction property of $\mathcal{T}$ with the triangle inequality and is given in \online{app: proofs exact}. Based on this lemma, Theorem \ref{thm: convergence} provides the convergence guarantee of Algorithm \ref{alg: BOBBF} under both options.

\begin{theorem} [Convergence of BF learning]
\label{thm: convergence}
    The following properties hold for Algorithm \ref{alg: BOBBF} under either option.
    \begin{itemize}
        \item[a)] The Bellman error converges to zero in the $\nu$-weighted norm, i.e., $ \left\|\mathcal{T}\tilde{V}_k - \tilde{V}_k\right\|_{1,\nu} \xrightarrow[k\to\infty]{} 0.$

        \item[b)] If $\mathcal{S}$ is finite and $\nu$ has full support on $\mathcal{S}$, then Algorithm \ref{alg: BOBBF} terminates with an $\epsilon$-optimal approximated value function.
    \end{itemize}
\end{theorem}

The proof of Theorem \ref{thm: convergence} is provided in~\nameref{app:A}. Suppose that, at each iteration $k$ of Algorithm \ref{alg: BOBBF} under Option 1, the greedy policy $\tilde{\pi}_k$ is unique at every state $s$ with a binding primal constraint (Definition \ref{def: binding}). Under this assumption, the bound (\ref{eq: major improvement 1}) holds at each iteration $k$, and the factor $\frac{\gamma}{1-\gamma}$ accelerates convergence when the dual-weighted Bellman error is large. Although this statement depends on policy uniqueness at certain states, we argue that substantial improvement per iteration may still occur even when the greedy policy is not unique at binding states. In Section \ref{sec: Numerical Studies}, we provide experimental evidence showing that both options typically converge with only a few iterations. A potential drawback of Option 1 is that the number of BFs in $\Phi$ could, in theory, grow substantially, thereby increasing the size of the \ALP{\Phi} problem at each iteration. However, Option 2 removes this drawback at the price of the amplification certificate.

Part (a) of Theorem \ref{thm: convergence} carries over when $\mathcal{S}$ and $\mathcal{A}$ are continuous and compact, since its proof uses only the feasibility of $(0,\dots,0,1)^\top$, the monotonicity of $\mathcal{T}$, and a telescoping argument. However, Part (b) requires upgrading convergence in $\|\cdot\|_{1,\nu}$ to uniform convergence, which is not automatic on a continuum. Such an upgrade calls for a regularity bound on the model, for instance Lipschitz continuity of the cost and of the transition kernel in the state variable, together with uniform control of the regularity of the iterates. Regularity of this kind is known to propagate through the Bellman operator \citep{hinderer2005lipschitz}. A rigorous treatment is, however, beyond the scope of this study, and we do not pursue it owing to space limitations.

By Lemma \ref{lem: epsilon optimal}, the estimate returned by Algorithm \ref{alg: BOBBF} under either option satisfies the conservative bound $\|V^*-\tilde{V}\|_{1,\nu}\le\epsilon$ upon termination.

\section{Constraint Sampling Scheme}\label{sec: Constraint Sampling}

While our BF learning approaches offer promising theoretical implications, they are hindered by the large number of constraints in the \ALP{\Phi} problem, which must be solved at each iteration. In this section, we rigorously analyze a constraint reduction technique integrated into our BF learning framework and demonstrate its benefits. Previous studies have proposed various strategies to address the extensive constraint set in \ALP{\Phi}, such as constraint sampling (\citealp{de_farias_constraint_2004}), column generation (\citealp{adelman_dynamic_2007}), and constraint violation learning (\citealp{lin_revisiting_2020}). Among these, we adopt the constraint sampling method due to its simplicity and generalizability.

\subsection{Problem Formulation and Properties}

Throughout this section, a \emph{sample} is a finite set $\mathcal{C}\subseteq(\mathcal{S}\times\mathcal{A})$ of state--action pairs. With a slight abuse of notation, we write $\mathcal{C}_{\mathcal{S}}:=\{s\in\mathcal{S}:\ \exists\,a\in\mathcal{A} \text{ with }(s,a)\in\mathcal{C}\}$ for the set of states appearing in $\mathcal{C}$ and, for $s\in\mathcal{C}_{\mathcal{S}}$, $\mathcal{C}_{\mathcal{A}}(s):=\{a\in\mathcal{A}:(s,a)\in\mathcal{C}\}$ for the set of actions sampled at $s$. We set $\mathcal{C}_{\mathcal{A}}(s):=\emptyset$ when $s\in\mathcal{S}\setminus\mathcal{C}_{\mathcal{S}}$. Given a sample $\mathcal{C}$ and a basis vector $\Phi$ with $\ell$ elements, the \emph{reduced linear program} (RLP) retains only the constraints indexed by $\mathcal{C}$:
\begin{align*}
    \text{RLP($\Phi, \mathcal{C}$)}: \quad \max_{w \in \mathbb{R}^\ell} & \quad \nu^\top\Phi w,\\
s.t.&\quad (\Phi w)(s)\leq c(s,a)+\gamma\sum_{s'\in \mathcal{S}}p(s'|s,a)(\Phi w)(s'), \qquad
    \forall (s,a) \in \mathcal{C}.
\end{align*}
Because $\mathcal{C}$ is finite, \RLP{\Phi}{\mathcal{C}} is a finite linear program in $\ell$ variables. Nothing in the definition of \RLP{\Phi}{\mathcal{C}} presupposes how $\mathcal{C}$ was obtained.

Two sample sets play distinct roles in what follows. The first, denoted as $\mathcal{X}$, consists of $M$ independent and identically distributed (i.i.d.) draws from a probability distribution $\psi$ over $(\mathcal{S}\times\mathcal{A})$. \citet{de_farias_constraint_2004} demonstrated that, although many constraints are removed, the optimal solution of \RLP{\Phi}{\mathcal{X}} remains close to that of \ALP{\Phi} when $M$ is sufficiently large and $\psi$ is appropriately chosen. The second, denoted as $\mathcal{Y}$, is a \emph{curated} sample whose role is entirely computational. It is subject to no distributional restriction, and it is constructed once \emph{prior to any realization of $\mathcal{X}$}, so that its construction can draw on no information generated by the sampling experiment. We solve the RLP over the union
\begin{equation}
    \mathcal{Z} := \mathcal{X} \cup \mathcal{Y} \subset (\mathcal{S}\times\mathcal{A}),
    \label{eq: Z}
\end{equation}
so that $\mathcal{Z}_{\mathcal{S}} = \mathcal{X}_{\mathcal{S}} \cup \mathcal{Y}_{\mathcal{S}}$ and
$\mathcal{Z}_{\mathcal{A}}(s) = \mathcal{X}_{\mathcal{A}}(s) \cup \mathcal{Y}_{\mathcal{A}}(s)$ for
every $s\in\mathcal{Z}_{\mathcal{S}}$. Section \ref{sec: sampling architecture} explains why the two designs differ and how $\mathcal{Y}$ is built. The mechanism that expands $\Phi$ is deferred to Section \ref{sec: BF expansion with sampling}, and until then $\Phi$ may be read as a collection of predefined basis functions.

The properties developed in the remainder of this subsection are stated for $\mathcal{Z}$, since that is the sample whose RLP we solve. They are, however, purely linear-programming facts. All of them hold for \RLP{\Phi}{\mathcal{C}} at any finite sample $\mathcal{C}$ and use no property of $\mathcal{X}$, $\mathcal{Y}$, or $\psi$.

For any value estimate $v \in \mathbb{R}^{N}$ and any sample $\mathcal{C}$, we define
the \textit{hybrid greedy policy} $\tilde{\pi}^{\mathcal{C}}_v$ by
\begin{equation}
    \tilde{\pi}^{\mathcal{C}}_v(s) := \begin{cases}
        \argmin_{a\in \mathcal{C}_{\mathcal{A}}(s)} \big\{ c(s,a) + \gamma \sum_{s'\in \mathcal{S}}p(s'\mid s,a)v(s')\big\} \qquad & \text{if } s \in \mathcal{C}_{\mathcal{S}},    \\
        \text{Arbitrary } a\in{\mathcal{A}} & \text{if } s\in \mathcal{S}\setminus  \mathcal{C}_{\mathcal{S}},
    \end{cases} \label{eq: hybrid greedy policy}
\end{equation}
and the \textit{hybrid Bellman operator} $\mathcal{L}_{\mathcal{C}}$ by
\begin{equation}
    (\mathcal{L}_{\mathcal{C}}v)(s) := \begin{cases}
        \min_{a\in \mathcal{C}_{\mathcal{A}}(s)} \big\{ c(s,a) + \gamma \sum_{s'\in \mathcal{S}}p(s'\mid s,a)v(s')\big\} \qquad & \text{if } s \in \mathcal{C}_{\mathcal{S}},    \\
        v(s) & \text{if } s\in \mathcal{S}\setminus  \mathcal{C}_{\mathcal{S}},
    \end{cases} \label{eq: hybrid Bellman operator}
\end{equation}
for all $s\in\mathcal{S}$. Both objects are well defined at $\mathcal{X}$, at $\mathcal{Y}$, and at $\mathcal{Z}$ alike. We allow $\tilde{\pi}^{\mathcal{C}}_v(s)$ to map a state $s\in\mathcal{S}\setminus\mathcal{C}_{\mathcal{S}}$ to an arbitrary action, since all actions at such a state are equivalent from the dual perspective (see Property \ref{prop: rlp duality}). The hybrid greedy policy is employed solely for theoretical analysis. In practice, standard greedy policies can be used to compute policy costs.

Two instances of (\ref{eq: hybrid greedy policy}) and (\ref{eq: hybrid Bellman operator}) recur. The policy $\tilde{\pi}^{\mathcal{Z}}_v$ is the greedy policy associated with \RLP{\Phi}{\mathcal{Z}}, and it is therefore at $\mathcal{Z}$ that the dual properties below are established. The operator $\mathcal{L}_{\mathcal{Y}}$, by contrast, is the computational device that replaces the exact Bellman operator. When the state and action spaces are extremely large, applying $\mathcal{T}$ as in Algorithm \ref{alg: BOBBF} is not possible, since it requires an $\argmin$ over $\mathcal{A}$ at every state of $\mathcal{S}$, whereas $\mathcal{L}_{\mathcal{Y}}$ requires one minimization over $\mathcal{Y}_{\mathcal{A}}(s)$ at each of the finitely many states of $\mathcal{Y}_{\mathcal{S}}$.

\begin{definition} [Binding primal constraint under sampling]
\label{def: binding (sampling)}
Let $\tilde{V}=\Phi\tilde{w}$ with $\tilde{w}$ an optimal solution to \RLP{\Phi}{\mathcal{Z}}. A state $s\in\mathcal{S}$ is said to \emph{have a binding primal constraint} in \RLP{\Phi}{\mathcal{Z}} if the constraint indexed by $(s,a)$ holds with equality for at least one $a\in\mathcal{Z}_\mathcal{A}(s)$, i.e., $\exists\, a \in \mathcal{Z}_\mathcal{A}(s)$ such that $\tilde{V}(s) = c(s,a) + \gamma \sum_{s' \in \mathcal{S}}p(s'|s,a)\tilde{V}(s')$.
\end{definition}

\begin{property}
\label{prop: rlp duality}
The dual of \RLP{\Phi}{\mathcal{Z}}, denoted as \RDUAL{\Phi}{\mathcal{Z}}, is \DUAL{\Phi} with
the additional condition $u_a(s)=0$ for all $(s,a)\in(\mathcal{S}\times\mathcal{A})\setminus\mathcal{Z}$. Strong duality and complementary slackness hold as in Property \ref{prop: dual properties}. Let $\tilde{V} = \Phi \tilde{w}$ with $\tilde{w}$ being an optimal solution to \RLP{\Phi}{\mathcal{Z}}. If the hybrid greedy policy $\tilde{\pi}^{\mathcal{Z}}_{\tilde{V}}(s)$ is unique at every $s\in\mathcal{Z}_{\mathcal{S}}$ with a binding primal constraint (Definition \ref{def: binding (sampling)}), then an optimal dual solution to \RDUAL{\Phi}{\mathcal{Z}}, denoted as $\tilde{u}$, satisfies:
\[
\tilde{u}_{\tilde{\pi}^{\mathcal{Z}}_{\tilde{V}}}^\top c_{\tilde{\pi}^{\mathcal{Z}}_{\tilde{V}}}
=\nu^\top\tilde{V},
\quad
\tilde{u}_{\tilde{\pi}^{\mathcal{Z}}_{\tilde{V}}}^\top
\big(I-\gamma P_{\tilde{\pi}^{\mathcal{Z}}_{\tilde{V}}}\big)\Phi=\nu^\top\Phi,\text{ and}
\quad
\tilde{u}_{\tilde{\pi}^{\mathcal{Z}}_{\tilde{V}}}^\top\mathbf{1}=\frac{1}{1-\gamma},
\]
with the last identity holding whenever $\mathbf{1}$ is a column of $\Phi$.
\end{property}

Property \ref{prop: rlp duality} is the exact analogues of Property \ref{prop: dual structure}, obtained by replacing the greedy policy with its hybrid counterpart. The formal statements and proofs are collected in \online{sec: dual proofs for RLP}. These proofs use no property of $\mathcal{X}$, $\mathcal{Y}$, or $\psi$ beyond finiteness of $\mathcal{Z}$. Since \RLP{\Phi}{\mathcal{Z}} is a finite linear program, no constraint qualification beyond feasibility and boundedness is required. Property \ref{prop: rlp duality} also explains why $\tilde{\pi}^{\mathcal{C}}_v(s)$ may be left arbitrary off $\mathcal{C}_{\mathcal{S}}$. The reason is that all actions at such a state carry zero dual weight.

We record one further elementary fact about $\mathcal{L}_{\mathcal{Y}}$, used repeatedly in the rest of the paper.

\begin{lemma} [Hybrid feasibility dominance]
\label{lem: L_Y monotone}
Let $\mathcal{C}\supseteq\mathcal{Y}$ be a sample and let $\bar{w}$
satisfy the subset of the \RLP{\Phi}{\mathcal{C}} constraints indexed by $\mathcal{Y}$. That is, let $\bar{V}:=\Phi\bar{w}$ satisfy
\[
\bar{V}(s)\ \le\ c(s,a)+\gamma\sum_{s'\in\mathcal{S}}p(s'\mid s,a)\bar{V}(s')
\qquad\text{for all }(s,a)\in\mathcal{Y}.
\]
Then $\mathcal{L}_{\mathcal{Y}}\bar{V}\ge\bar{V}$ component-wise on $\mathcal{S}$. In particular,
the conclusion holds for every feasible solution of
\RLP{\Phi}{\mathcal{C}} whenever $\mathcal{Y}\subseteq\mathcal{C}$.
\end{lemma}

The proof of Lemma \ref{lem: L_Y monotone} restricts the feasibility inequality at each $s\in\mathcal{Y}_{\mathcal{S}}$ to $a\in\mathcal{Y}_{\mathcal{A}}(s)$ and minimizes over that set. Moreover, for states $s \in \mathcal{S}\setminus\mathcal{Y}_{\mathcal{S}}$, $\mathcal{L}_{\mathcal{Y}}\bar{V}=\bar{V}$ by the definition in (\ref{eq: hybrid Bellman operator}). See \online{sec: dual proofs for RLP} for a formal proof.

We begin by formalizing the relationship between the sample size $M$ and the volume of constraint violations in Proposition \ref{prop: probability of violations}.

\begin{proposition} [Scenario bound on constraint violations]
    \label{prop: probability of violations}
    \textup{(Corollary 1 in \citealp{calafiore2006scenario})} Suppose $\mathcal{X}$ consists of $M$
    i.i.d.\ samples drawn from $\psi$. If \RLP{\Phi}{\mathcal{X}} has a unique optimal
    solution $\tilde{w}$, and
    \[ M \geq \left\lceil \frac{2}{\xi} \ln \frac{1}{\delta} + 2\ell + \frac{2\ell}{\xi} \ln \frac{2}{\xi}
    \right\rceil,
\] with $\delta$ and $\xi$ being real numbers in the range $(0,1)$ and $\ell$ representing the number
of BFs in $\Phi$, then
\begin{align}
\psi \left(\left\{ (s,a) \in (\mathcal{S}\times \mathcal{A}): \left(\Phi \tilde{w}\right)(s) >
c(s,a)+\gamma\sum_{s'\in \mathcal{S}}p(s'|s,a)\left(\Phi \tilde{w}\right)(s')\right\}\right)\leq \xi
\label{eq: psi less than xi}
\end{align} holds with a probability of at least $1-\delta$.
\end{proposition}

Given $M$, $\delta$, and a program with $\ell$ basis functions, we write $\xi$ for any level in $(0,1)$ for which the sample-size requirement of Proposition \ref{prop: probability of violations} is met. Every statement below invokes Proposition \ref{prop: probability of violations} at a single program, so $\xi$ always denotes the violation level of the program named in that statement and carries no further index.

Writing the requirement as $ M \;=\; \Theta\!\left(\xi^{-1}\left(\ell\ln\xi^{-1}+\ln\delta^{-1}\right)+\ell\right)$ in Proposition \ref{prop: probability of violations}, the dependence of the sample size $M$ on the confidence level is logarithmic in $1/\delta$, whereas the dependence on the violation level is linear in $1/\xi$ up to a logarithmic factor. What is remarkable is not the rate but the absence of any dependence on $|\mathcal{S}\times\mathcal{A}|$. The sample size $M$ is governed only by the dimension $\ell$ of the approximation architecture and by the violation--confidence pair $(\xi,\delta)$.

\begin{assumption} [Boundedness of the sampled program] \label{ass: rlp-bounded 1}
For the sample $\mathcal{X}$ drawn as in Proposition \ref{prop: probability of violations}, $\mathrm{RLP}(\Phi,\mathcal{X})$ admits a bounded optimal solution.
\end{assumption}

Since $\mathcal{Z}=\mathcal{X}\cup\mathcal{Y}\supseteq\mathcal{X}$, every constraint of
$\mathrm{RLP}(\Phi,\mathcal{X})$ is also a constraint of \RLP{\Phi}{\mathcal{Z}}, so the
latter has the smaller feasible region and an optimal value bounded above by that of the
former. Being a linear program with a bounded objective, \RLP{\Phi}{\mathcal{Z}} attains its
optimum whenever it is feasible, which is what Property \ref{prop: rlp duality} requires.

\subsection{Sampling Design of $\mathcal{X}$, $\mathcal{Y}$, and $\mathcal{Z}$}
\label{sec: sampling architecture}

The two samples are governed by different design requirements. The set $\mathcal{X}$
underlies \RLP{\Phi}{\mathcal{X}} and is the object to which Proposition
\ref{prop: probability of violations} applies. That guarantee is a scenario-optimization
result whose proof bounds the number of support constraints of $\tilde{w}$ by a symmetrization
argument treating all $M$ sampled constraints as exchangeable, which holds only under
independent draws from a single common distribution (\citealp{calafiore2006scenario}). Hence
$\mathcal{X}$ must remain exactly what it was defined to be.

The set $\mathcal{Y}$ plays no role in certifying a probabilistic feasibility guarantee. Its purpose is to make $\mathcal{L}_{\mathcal{Y}}$ computationally tractable, so that it can be used to expand $\Phi$. Since $\mathcal{L}_{\mathcal{Y}}$ replaces the exact minimization over $\mathcal{A}$ with a minimization over $\mathcal{Y}_{\mathcal{A}}(s)$ at each $s\in\mathcal{Y}_{\mathcal{S}}$, the quality of the resulting operator depends on how densely $\mathcal{Y}_{\mathcal{A}}(s)$ covers the action space at each retained state, not on any distributional property of how the states in $\mathcal{Y}_{\mathcal{S}}$ were chosen. What is needed is only that $\mathcal{Y}_{\mathcal{S}}$ be finite and that $\mathcal{Y}_{\mathcal{A}}(s)$ be rich enough at each retained state. The sample $\mathcal{Y}$ should collect the states at which a high-quality value approximation is most desired, with $\mathcal{Y}_{\mathcal{S}}$ as large and $\mathcal{Y}_{\mathcal{A}}(s)$ as action-rich as computational resources permit. In this regard, any approach to construct $\mathcal{Y}$ is admissible. For example, trajectory-based or chain-walk exploration under predefined policies, draws from $\psi$ or any other distribution, expert-driven selection of operationally important pairs, or a combination of different approaches can be applied to construct $\mathcal{Y}$.

The program we solve is accordingly \RLP{\Phi}{\mathcal{Z}} with $\mathcal{Z}=\mathcal{X}\cup\mathcal{Y}$, and the associated hybrid greedy policy is $\tilde{\pi}^{\mathcal{Z}}_v$ of (\ref{eq: hybrid greedy policy}). Because $\mathcal{Y}$ is fixed in advance, the scenario guarantee of Proposition \ref{prop: probability of violations} survives the enlargement of the constraint set intact.

\begin{corollary}[Scenario bound under the enlarged sample]
\label{cor: violations under Z}
Let $\xi,\delta\in(0,1)$ and let $\Phi$ be a BF vector with $\ell$ elements. Let $\mathcal{X}$ consist of $M$ i.i.d.\ state--action pairs drawn from $\psi$ and $\mathcal{Y}$ be the curated sample set, let $\mathcal{Z}=\mathcal{X}\cup\mathcal{Y}$ as in (\ref{eq: Z}), and suppose \RLP{\Phi}{\mathcal{Z}} admits a unique optimal solution $\tilde{w}$. If
\[
M \ \geq\ \left\lceil \frac{2}{\xi}\ln\frac{1}{\delta}+2\ell+\frac{2\ell}{\xi}\ln\frac{2}{\xi}\right\rceil,
\]
then, with probability at least $1-\delta$,
\begin{align}
\psi\bigg(\bigg\{(s,a)\in(\mathcal{S}\times\mathcal{A}):\ \big(\Phi\tilde{w}\big)(s)
> c(s,a)+\gamma\sum_{s'\in\mathcal{S}}p(s'|s,a)\big(\Phi\tilde{w}\big)(s')\bigg\}\bigg)\ \leq\ \xi .
\label{eq: psi less than xi under Z}
\end{align}
\end{corollary}

\textit{Proof of Corollary \ref{cor: violations under Z}.}
The randomness enters only through the $M$ i.i.d.\ pairs of $\mathcal{X}$. Since $\mathcal{Y}$ is fixed before any draw is realized, the constraints it contributes are identical in every realization and may be absorbed into the decision set $\mathscr{W}_{\mathcal{Y}}:=\{w\in\mathbb{R}^\ell:(\Phi w)(s)\le c(s,a)+\gamma\sum_{s'}p(s'\mid s,a)(\Phi w)(s'),\ \forall(s,a)\in\mathcal{Y}\}$, a closed convex set as a finite intersection of half-spaces. Then \RLP{\Phi}{\mathcal{Z}} is exactly the scenario program that maximizes $\nu^\top\Phi w$ over $w\in\mathscr{W}_{\mathcal{Y}}$ subject to the $M$ i.i.d.\ scenarios indexed by $\mathcal{X}$, so Proposition \ref{prop: probability of violations} applies with $\mathscr{W}_{\mathcal{Y}}$ in place of $\mathbb{R}^\ell$ since $\mathscr{W}_{\mathcal{Y}}$ is a closed convex set (see \citealp{calafiore2006scenario}). The constraint functions remain affine in $w$, the number of decision variables remains $\ell$, and the number of random scenarios remains $M$. $\square$

\subsection{The Choice of Sampling Distribution}

A critical factor in the constraint sampling scheme is the choice of the distribution $\psi$. While the existing ALP literature typically restricts this to the discounted occupancy distribution under the optimal policy $\pi^*$ (\citealp{de_farias_constraint_2004} and \citealp{farias2006tetris}), we demonstrate that $\psi$ can be a general sampling distribution, provided it satisfies Assumption \ref{ass: psi vs d^*}. As a baseline we use the discounted occupancy distribution $d_\nu^\pi:=  (1-\gamma)\big(\nu^\top(I-\gamma P_\pi)^{-1}\big)^\top$ of a policy $\pi$ under initial state distribution $\nu$. \online{oa: occupancy} formalizes $d_\nu^\pi$ and proves that it is a distribution, using the Neumann expansion of $(I-\gamma P_\pi)^{-1}$ and $P_\pi\mathbf{1}=\mathbf{1}$.

\begin{assumption} [Localized concentrability]
\label{ass: psi vs d^*}
Let $\Phi$ be a basis vector and let $\mathcal{Y}$ be the fixed curated sample. Denote by $\mathcal{W}(\Phi)$ the set of optimal solutions to
\RLP{\Phi}{\mathcal{X}\cup\mathcal{Y}} obtained as $\mathcal{X}$ ranges over all the
realizations of $M$ i.i.d.\ draws from $\psi$. For $\tilde{w}\in\mathcal{W}(\Phi)$ write
$V_{\tilde{w}} := \Phi \tilde{w}$ and define
\begin{align*}
\mathcal{B}(V_{\tilde{w}}) &:= \Big\{(s,a)\in\mathcal{S}\times\mathcal{A}:\
V_{\tilde{w}}(s) > c(s,a)+\gamma\textstyle\sum_{s'\in\mathcal{S}}p(s'\mid s,a)V_{\tilde{w}}(s')\Big\},\\
\mathcal{B}^{*}(V_{\tilde{w}}) &:= \Big\{s\in\mathcal{S}:\
V_{\tilde{w}}(s) > c(s,\pi^*(s))+\gamma\textstyle\sum_{s'\in\mathcal{S}}p(s'\mid s,\pi^*(s))V_{\tilde{w}}(s')\Big\},
\end{align*}
and let $d_\nu^{\pi^*}$ be the discounted occupancy distribution under the optimal policy $\pi^*$. We assume there exists a
\emph{deterministic} constant $\beta\in\mathbb{R}_+$ such that
\[
\beta\,\psi\big(\mathcal{B}(V_{\tilde{w}})\big)\ \ge\ d_\nu^{\pi^*}\big(\mathcal{B}^{*}(V_{\tilde{w}})\big)
\]
for every $\tilde{w}\in\mathcal{W}(\Phi)$.
\end{assumption}

\begin{lemma} [Sufficient condition for finite concentrability]
\label{lem: beta sufficient}
Suppose $\mathcal{S}$ is finite and $\psi(s,\pi^*(s))>0$ for every $s\in\mathcal{S}$, and set
\[
\beta_{\max}\ :=\ \max_{s\in\mathcal{S}}\ \frac{d_\nu^{\pi^*}(s)}{\psi\big(s,\pi^*(s)\big)}\ <\ \infty .
\]
Then Assumption \ref{ass: psi vs d^*} holds with $\beta=\beta_{\max}$ for all $\tilde{w}\in\mathcal{W}(\Phi)$.
\end{lemma}

The proof of Lemma \ref{lem: beta sufficient} is in \online{oa: proof beta}.  Assumption \ref{ass: psi vs d^*} is a concentrability condition bridging $\psi$ and $d_\nu^{\pi^*}$. It bounds the $d_\nu^{\pi^*}$-mass of the states at which $\pi^*$'s constraint is violated by the $\psi$-mass of the violated pairs, inflated by $\beta$. Quantifying over $\mathcal{W}(\Phi)$ rather than over the single realized solution is what makes $\beta$ deterministic rather than random. The condition is \emph{local}, comparing the two distributions only on the violation sets generated by $\mathcal{W}(\Phi)$, so the full-support requirement of Lemma \ref{lem: beta sufficient} is sufficient but far from necessary. For continuous $\mathcal{S}$ the absolute continuity of $d_\nu^{\pi^*}$ with respect to $\psi$ must be secured either by kernel regularity or by designing $\psi$ as a mixture that includes trajectories of predefined policies (\online{app: concentrability}).

Based on Assumption \ref{ass: psi vs d^*}, we establish in Theorem \ref{thm: ALP vs RLP} a relationship between the approximation errors of the \RLP{\Phi}{\mathcal{Z}} and \ALP{\Phi} solutions.

\begin{theorem} [Approximation error under constraint sampling] \label{thm: ALP vs RLP}
Let $\xi,\delta\in(0,1)$, let $\Phi$ be a BF vector with $\ell$ elements formed independently of $\mathcal{X}$, and let $\mathcal{X}$, $\mathcal{Y}$, $\mathcal{Z}$, and $M$ satisfy the hypotheses of Corollary \ref{cor: violations under Z} at the pair $(\xi,\delta)$ and $\Phi$. Let $\tilde{w}$ denote the unique optimal solution to \RLP{\Phi}{\mathcal{Z}}, and let $\tilde{w}'$ denote an optimal solution to \ALP{\Phi}. Define $\tilde{V} = \Phi \tilde{w}$, $\tilde{V}' = \Phi \tilde{w}'$, and let $\beta$ be as in Assumption \ref{ass: psi vs d^*}. Then, with a probability of at least $1-\delta$,
\begin{align}
\|V^* -  \tilde{V}\|_{1,\nu} \leq  \|V^* -  \tilde{V}'\|_{1,\nu}
+ \frac{2 \beta \xi}{1-\gamma}\left\|(\mathcal{T}_{\pi^*}\tilde{V} - \tilde{V})^-\right\|_\infty,
\label{eq: ALP vs RLP}
\end{align}
where $(v)^- = \max\{-v ,0\}$ (applied component-wise) for any real-valued vector $v$.
\end{theorem}

The proof of Theorem \ref{thm: ALP vs RLP} is provided in~\nameref{app:A}.  Equation (\ref{eq: ALP vs RLP}) indicates that when the violation level $\xi$ is small, the approximation error of the RLP is comparable to that of the ALP with a probability of at least $1-\delta$.

\subsection{Basis Function Expansion with Constraint Sampling}
\label{sec: BF expansion with sampling}

Similar to Section \ref{sec: BOBF}, our objective here is to quantify the net approximation improvement obtained by appending a \emph{hybrid Bellman operator basis function} (HBOBF) to the basis vector. We first isolate a single such expansion, so that the conditions supporting it can be stated and interpreted without reference to any procedure that repeats it.

\begin{definition}[Basis-expansion step]
\label{def: expansion step}
Let $\mathcal{Y}$ be the curated sample. A \emph{basis-expansion step}, labelled by an index $k$, consists of the following. Given a BF vector $\Phi_k$ containing $\ell_k$ elements and a sample $\mathcal{Z}^{k}=\mathcal{X}^{k}\cup\mathcal{Y}$,
let $\tilde{w}_k$ be an optimal solution of \RLP{\Phi_k}{\mathcal{Z}^{k}} and put
$\tilde{V}_k=\Phi_k\tilde{w}_k$. Form
\begin{equation}
\Phi_{k+1}:=\big[\Phi_k,\ \mathcal{L}_{\mathcal{Y}}\tilde{V}_k\big],
\label{eq: expansion}
\end{equation}
\emph{then} draw $\mathcal{X}^{k+1}$, consisting of $M$ i.i.d.\ state--action pairs from $\psi$, and
set $\mathcal{Z}^{k+1}:=\mathcal{X}^{k+1}\cup\mathcal{Y}$. Finally, let $\tilde{w}_{k+1}$ be an
optimal solution of \RLP{\Phi_{k+1}}{\mathcal{Z}^{k+1}} and
$\tilde{V}_{k+1}=\Phi_{k+1}\tilde{w}_{k+1}$.
\end{definition}

The curated sample $\mathcal{Y}$ carries no index, being common to
$\mathcal{Z}^{k}$ and $\mathcal{Z}^{k+1}$ alike. To lighten the notation we abbreviate, for the step of Definition \ref{def: expansion step},
\begin{equation}
\hat{\pi}_{k+1}:=\tilde{\pi}^{\mathcal{Z}^{k+1}}_{\tilde{V}_{k+1}},
\qquad
\hat{u}_{k+1}:=\tilde{u}^{k+1}_{\hat{\pi}_{k+1}},
\label{eq: hat notation}
\end{equation}
where $\tilde{u}^{k+1}$ is an optimal solution of \RDUAL{\Phi_{k+1}}{\mathcal{Z}^{k+1}}.
Sections \ref{sec: one-step conditions} and \ref{sec: net improvement} analyze a single expansion step at a fixed $k$.

\subsubsection{Conditions for a Single Expansion Step}
\label{sec: one-step conditions}

The improvement guarantee of Section \ref{sec: net improvement} rests on two conditions: one controlling the cost of restricting the greedy minimization to $\mathcal{Y}$ and one controlling the effect of the fresh draw $\mathcal{X}^{k+1}$. In this section, we state each for the single expansion step of Definition \ref{def: expansion step} and argue that both are mild. We begin with the ordering of operations in that definition.

The order of operations in Definition \ref{def: expansion step} is a requirement of the scenario certificate. The new column $\phi=\mathcal{L}_{\mathcal{Y}}\tilde{V}_k$ is computed from $\tilde{V}_k$ and hence from $\mathcal{X}^{k}$, so reusing that sample would make the constraint functions of \RLP{\Phi_{k+1}}{\mathcal{Z}^{k+1}} statistically dependent on the very scenarios invoked to certify them. Drawing $\mathcal{X}^{k+1}$ after $\Phi_{k+1}$ has been formed restores the structure Proposition \ref{prop: probability of violations} requires.

\begin{definition}[Action-restriction gap]
\label{def: Y gap}
For a value estimate $\bar{V}\in\mathbb{R}^{N}$ and a sample
$\mathcal{C}\subseteq\mathcal{S}\times\mathcal{A}$ with $\mathcal{C}\supseteq\mathcal{Y}$, define
\begin{align}
\mathscr{E}(\bar{V},\mathcal{C})\ :=\ \max_{s\in\mathcal{Y}_{\mathcal{S}}}
\Big[\big(\mathcal{L}_{\mathcal{Y}}\bar{V}\big)(s)-\big(\mathcal{L}_{\mathcal{C}}\bar{V}\big)(s)\Big]
\label{eq: Y suboptimality gap}
\end{align}
the largest amount, over the curated states, by which restricting the greedy minimization to the actions retained in $\mathcal{Y}$ inflates the Bellman backup relative to minimizing over all actions available in $\mathcal{C}$.
\end{definition}

In (\ref{eq: Y suboptimality gap}), because $\mathcal{C}\supseteq\mathcal{Y}$ implies $\mathcal{Y}_{\mathcal{A}}(s)\subseteq\mathcal{C}_{\mathcal{A}}(s)$, and a minimum over a subset is never smaller than the minimum over the superset, every bracket in (\ref{eq: Y suboptimality gap}) is non-negative. Hence $\mathscr{E}(\bar{V},\mathcal{C})$ always remains non-negative.

\begin{assumption} [Bounded action-restriction gap]
\label{ass: Y suboptimality gap}
For the basis-expansion step of Definition \ref{def: expansion step}, there exists a constant $e\in\mathbb{R}_+$ such that
\begin{equation}
\mathscr{E}\big(\tilde{V}_k,\ \mathcal{Z}^{k+1}\big)\ \le\ e,
\label{eq: e assumption}
\end{equation}
where $\mathscr{E}(\cdot,\cdot)$ is the action-restriction gap of Definition \ref{def: Y gap}, $\tilde{V}_k=\Phi_k\tilde{w}_k$ is the value estimate entering the basis-expansion step, and $\mathcal{Z}^{k+1}=\mathcal{X}^{k+1}\cup\mathcal{Y}$ is the sample leaving it.
\end{assumption}

Assumption \ref{ass: Y suboptimality gap} bounds, uniformly over the curated states, how far the $\mathcal{Y}$-restricted backup of $\tilde{V}_k$ can exceed the backup available over the full sampled action set of \RLP{\Phi_{k+1}}{\mathcal{Z}^{k+1}}. It is mild because the bracket in (\ref{eq: Y suboptimality gap}) is exactly zero at every $s\in\mathcal{Y}_{\mathcal{S}}\setminus\mathcal{X}^{k+1}_{\mathcal{S}}$, and elsewhere is strictly positive only if the independently drawn set $\mathcal{X}^{k+1}_{\mathcal{A}}(s)$ contains an action beating every action retained in the deliberately action-rich set $\mathcal{Y}_{\mathcal{A}}(s)$. In either case, $e$ may be taken close to, and often exactly, zero. \online{app: conditions} gives the detailed argument and explains why the assumption is imposed on the cross pairing $(\tilde{V}_k,\mathcal{Z}^{k+1})$ rather than on a single sample.

We turn to the second condition, which concerns the states contributed by the fresh draw in the basis-expansion step.

\begin{assumption}[Propagated correction dominance]
\label{ass: propagated dominance}
Consider the basis-expansion step of Definition \ref{def: expansion step}, with $\hat{\pi}_{k+1}$ and $\hat{u}_{k+1}$ as in (\ref{eq: hat notation}). We assume
\begin{equation}
\sum_{s\in\mathcal{Z}^{k+1}_{\mathcal{S}}\setminus\mathcal{Y}_{\mathcal{S}}}
\hat{u}_{k+1}(s)
\left[\left(\mathcal{T}_{\hat{\pi}_{k+1}}\mathcal{L}_{\mathcal{Y}}\tilde{V}_k\right)(s)-\tilde{V}_k(s)\right]
\ \geq\ 0 .
\label{eq: propagated dominance}
\end{equation}
\end{assumption}

Assumption \ref{ass: propagated dominance} concerns only the states that the fresh draw
$\mathcal{X}^{k+1}$ contributes outside $\mathcal{Y}_{\mathcal{S}}$. Writing
$r_k:=\mathcal{L}_{\mathcal{Y}}\tilde{V}_k-\tilde{V}_k$, which is non-negative by Lemma
\ref{lem: L_Y monotone} and vanishes off $\mathcal{Y}_{\mathcal{S}}$ by
(\ref{eq: hybrid Bellman operator}), and putting $a:=\hat{\pi}_{k+1}(s)$, each bracket in
(\ref{eq: propagated dominance}) decomposes as
\begin{equation}
\big(\mathcal{T}_{a}\mathcal{L}_{\mathcal{Y}}\tilde{V}_k\big)(s)-\tilde{V}_k(s)
=\underbrace{\Big[\big(\mathcal{T}_{a}\tilde{V}_k\big)(s)-\tilde{V}_k(s)\Big]}_{\text{(I)}}
+\underbrace{\gamma\sum_{s'\in\mathcal{S}}p(s'\mid s,a)\,r_k(s')}_{\text{(II)}} .
\label{eq: part b decomposition}
\end{equation}
Term (II) is always non-negative, and term (I) is non-negative exactly when $\tilde{V}_k$ satisfies the constraint indexed by $(s,a)$. Since $\tilde{V}_k$ is optimal for \RLP{\Phi_k}{\mathcal{Z}^{k}}, it satisfies every constraint indexed by $\mathcal{Z}^{k}$, so (I) can be negative only at a pair drawn afresh into $\mathcal{X}^{k+1}$. The assumption therefore asks that, in dual-weighted aggregate rather than pointwise, the one-step propagation of the correction made by the new BF dominates whatever violations the fresh draw exposes. By Corollary \ref{cor: violations under Z} the states that can subtract carry at most a $\xi$-fraction of the $\psi$-mass with probability at least $1-\delta$. 

\online{app: conditions} provides more details and insights on Assumptions \ref{ass: Y suboptimality gap} and \ref{ass: propagated dominance}.

\subsubsection{Net Approximation Improvement}
\label{sec: net improvement}

With the two conditions of Section \ref{sec: one-step conditions} in place, we can now account for the net approximation improvement in a single basis-expansion step. Theorem \ref{thm: major improvement (with sampling)} shows that a substantial net approximation improvement remains achievable by an HBOBF even when a significant fraction of the state-action space is excluded from the RLP formulation.

\begin{theorem} [Net improvement of an HBOBF]
    \label{thm: major improvement (with sampling)}
Consider the basis-expansion step of Definition \ref{def: expansion step}, with $\hat{\pi}_{k+1}$ and $\hat{u}_{k+1}$ as in (\ref{eq: hat notation}). Suppose that
\begin{enumerate}
    \item[(i)] $\tilde{w}_{k+1}$ is the unique optimal solution of
    \RLP{\Phi_{k+1}}{\mathcal{Z}^{k+1}}, and $\hat{\pi}_{k+1}(s)$ is unique at every
    $s\in\mathcal{Z}^{k+1}_{\mathcal{S}}$ with a binding primal constraint (Definition
    \ref{def: binding (sampling)});
    \item[(ii)] Assumption \ref{ass: Y suboptimality gap} holds with constant $e$;
    \item[(iii)] Assumption \ref{ass: psi vs d^*} holds for $\Phi_{k+1}$ with constant $\beta$; and
    \item[(iv)] $M$ meets the sample-size requirement of Proposition
    \ref{prop: probability of violations} for $\ell_{k+1}$ basis functions and the pair $(\xi,\delta)$, so that Corollary \ref{cor: violations under Z} certifies the violation level $\xi$ for \RLP{\Phi_{k+1}}{\mathcal{Z}^{k+1}}.
\end{enumerate}
Then, with probability at least $1-\delta$,
    \begin{align}
     \Delta_{\phi=\mathcal{L}_{\mathcal{Y}}\tilde{V}_k} \nu^\top\tilde{V}_k
     \ \geq\ \|\mathcal{L}_{\mathcal{Y}}\tilde{V}_k-\tilde{V}_{k}\|_{1,\nu}
     + g\big(\hat{u}_{k+1},\hat{\pi}_{k+1},\tilde{V}_k\big)
     - e\!\!\sum_{s\in \mathcal{Y}_{\mathcal{S}}}\!\!\hat{u}_{k+1}(s)
     - \frac{2 \beta\, \xi}{1-\gamma}
     \left\|(\mathcal{T}_{\pi^*}\tilde{V}_{k+1} - \tilde{V}_{k+1})^-\right\|_\infty ,
     \label{eq: major improvement sampling}
     \end{align}
where
$\Delta_{\phi}\nu^\top\tilde{V}_k:=\|V^*-\tilde{V}_k\|_{1,\nu}-\|V^*-\tilde{V}_{k+1}\|_{1,\nu}$ and
     \begin{align}
         g\big(\hat{u}_{k+1},\hat{\pi}_{k+1},\tilde{V}_k\big)
         &:= \gamma\!\!\sum_{s\in \mathcal{Y}_{\mathcal{S}}}\!\!\hat{u}_{k+1}(s)
         \Big(P_{\hat{\pi}_{k+1}}\big(\mathcal{L}_{\mathcal{Y}}\tilde{V}_k-\tilde{V}_k\big)\Big)(s) \notag \\
         &\quad + \sum_{s\in\mathcal{Z}^{k+1}_{\mathcal{S}} \setminus \mathcal{Y}_{\mathcal{S}}}
         \hat{u}_{k+1}(s)
         \Big(\big(\mathcal{T}_{\hat{\pi}_{k+1}}\mathcal{L}_{\mathcal{Y}}\tilde{V}_k\big)(s)-\tilde{V}_k(s)\Big).
         \label{eq: g}
     \end{align}
\end{theorem}

The violation level $\xi$ in (\ref{eq: major improvement sampling}) is that of the single program \RLP{\Phi_{k+1}}{\mathcal{Z}^{k+1}}. In particular, condition (iv) is a requirement on the sample drawn for a single basis-expansion step. The proofs of Theorem \ref{thm: major improvement (with sampling)} and of Lemma \ref{lem: g nonnegative} below are provided in~\nameref{app:A}.

\begin{lemma} [Non-negativity of the propagation term] \label{lem: g nonnegative}
Under the hypotheses of Theorem \ref{thm: major improvement (with sampling)}, with Assumption \ref{ass: propagated dominance} we have
\[
g\big(\hat{u}_{k+1},\hat{\pi}_{k+1},\tilde{V}_k\big)\ \geq\ 0 .
\]
\end{lemma}

The four terms of (\ref{eq: major improvement sampling}) read as follows. The leading term is the $\nu$-weighted empirical Bellman error, which delivers a monotone improvement whenever it is non-zero. The function $g$ collects a second-order contribution: at the curated states it propagates the correction just made by the new BF one step downstream along the trajectories of $\hat{\pi}_{k+1}$, and this part is non-negative unconditionally, whereas at the states contributed by $\mathcal{X}^{k+1}$ alone the corresponding terms may be negative, since a freshly drawn pair can expose a constraint that $\tilde{V}_k$ violates. Assumption \ref{ass: propagated dominance} is precisely the aggregate condition under which Lemma \ref{lem: g nonnegative} gives $g\ge0$, so that the propagated term can only reinforce the improvement. The third term is the dual-weighted price of computing the hybrid Bellman operator over $\mathcal{Y}_{\mathcal{A}}(s)$ rather than over the full $\mathcal{Z}^{k+1}_{\mathcal{A}}(s)$, and it is generically zero by the design of $\mathcal{Y}$. The fourth is the probabilistic cost of relaxing \ALP{\Phi_{k+1}} to a sampled program, and it shrinks with the violation level $\xi$, hence with a larger sample size $M$. Taken together, with probability at least $1-\delta$, restricting the ALP to a tractable constraint set preserves the net improvement guarantee up to two controllable penalties. A term-by-term discussion on (\ref{eq: major improvement sampling}) is given in \online{oa: reading of bound}.

\subsection{Basis Function Learning Methods}
\label{sec: BF learning with sampling}

In this section, we iterate the basis-expansion step of Definition \ref{def: expansion step} and obtain practical BF learning algorithms with constraint sampling (see Algorithm \ref{alg: HBOBF and SHBOBF}). Within this framework, we also introduce the Shrunken HBOBF (SHBOBF) approach, in which the number of BFs is capped by a parameter $\nmax$.

\begin{algorithm}[h]
    \caption{BF learning with constraint sampling. }\label{alg: HBOBF and SHBOBF}
    \textbf{Initialization}: Let $\Phi_0=[\mathbf{1}]$. Initialize $\nu$ and $\epsilon$ (the optimality tolerance). Construct the action-rich curated sample $\mathcal{Y}$ as described in Section \ref{sec: sampling architecture}. Draw $\mathcal{X}^0$, consisting of $M$ i.i.d.\ state--action pairs from the selected distribution $\psi$, and let $\mathcal{Z}^0 = \mathcal{X}^0\cup \mathcal{Y}$. Solve the \RLP{\Phi_0}{\mathcal{Z}^0} problem, let $\tilde{V}_0=\Phi_0 \tilde{w}_0$ denote the approximated value function, and set the iteration counter $k=0$. Select either ``Option 1: HBOBF'' or ``Option 2: SHBOBF($\nmax$)''. \\

    \While{$\| \mathcal{L}_{\mathcal{Y}}\tilde{V}_k - \tilde{V}_k \|_\infty> \epsilon$ }{
        Compute $\phi = \mathcal{L}_{\mathcal{Y}}\tilde{V}_k.$\\
         Update the BF vector \textbf{$\Phi_{k+1}$} $=$ $[\Phi_k, \phi]$.\\
        \If{``Option 2: SHBOBF($\nmax$)'' is selected}{
            Let $\ell_{k+1}$ be the number of BFs in $\Phi_{k+1}$, i.e., $ \Phi_{k+1}= [\phi_1, \phi_2, \cdots, \phi_{\ell_{k+1}}]$.\\
            \If{$\ell_{k+1} > \nmax$}{
                Retain the most recent $\nmax$ BFs: \textbf{$\Phi_{k+1}$} $=$ $[\phi_{\ell_{k+1}-\nmax+1}, \cdots,\phi_{\ell_{k+1}}]$.\\
            }
        }
        Draw the sample $\mathcal{X}^{k+1}$ which contains $M$ i.i.d. state--action pairs via the distribution $\psi$.\\
        Let $\mathcal{Z}^{k+1} = \mathcal{X}^{k+1}\cup \mathcal{Y}$. \\
        Solve \RLP{\Phi_{k+1}}{\mathcal{Z}^{k+1}} for an optimal solution $\tilde{w}_{k+1}$, and set $\tilde{V}_{k+1}=\Phi_{k+1}\tilde{w}_{k+1}$.\\
        Increment iteration counter: $k \leftarrow k+1 $.
    }
    \Return{$\tilde{V}_k$}\\
\end{algorithm}

As in Algorithm \ref{alg: BOBBF}, Option 2 does not retain the constant BF $\mathbf{1}$. The guarantees invoked below use only the strong duality and dual feasibility identities of Property \ref{prop: rlp duality}, neither of which requires $\mathbf{1}$ to be a column of $\Phi$.

The primary advantage of HBOBF, compared to SHBOBF, lies in its ability to improve approximations within a single iteration, as it retains all BFs. However, increasing the number of BFs degrades the certificate available at fixed $M$ in Proposition \ref{prop: probability of violations}. The smallest violation level $\xi$ admissible at a given confidence $1-\delta$ grows with the number of basis functions in the program being certified. In contrast, SHBOBF caps that number at $\nmax$ and thereby caps $\xi$ uniformly over the run, which constitutes its main advantage.

\begin{assumption}[Uniformly bounded iterates] \label{ass: rlp-bounded}
There exists $\bar{U}<\infty$, independent of the iteration index and of the realization of the samples, such that every program $\mathrm{RLP}(\Phi_k,\mathcal{Z}^{k})$ generated by Algorithm \ref{alg: HBOBF and SHBOBF} admits an optimal solution $\tilde{w}_k$ with $\nu^\top\Phi_k\tilde{w}_k\le\bar{U}$.
\end{assumption}

Assumption \ref{ass: rlp-bounded} is the uniform counterpart of Assumption \ref{ass: rlp-bounded 1}. It asks only that the optimal values generated by Algorithm \ref{alg: HBOBF and SHBOBF} not escape to infinity. An unbounded iterate is an observable event and is remedied by enlarging $\mathcal{Y}$ or increasing $M$.

\begin{theorem} [Finite termination of BF learning with sampling]\label{thm: convergence of HBOBF}
Suppose that
\begin{enumerate}
    \item[(i)] $\mathcal{Y}_{\mathcal{S}}$ is finite and $\nu$ has full support on it, so that
    $\nu_{\min}:=\min_{s\in\mathcal{Y}_{\mathcal{S}}}\nu(s)>0$;
    \item[(ii)] Assumption \ref{ass: Y suboptimality gap} holds with $e=0$ and Assumption
    \ref{ass: propagated dominance} holds, both at every iteration $k\ge0$ of Algorithm \ref{alg: HBOBF and SHBOBF};
    \item[(iii)] at every iteration $i$ of Algorithm \ref{alg: HBOBF and SHBOBF}, $\tilde{u}^{i}$ is an optimal solution of
    \RDUAL{\Phi_i}{\mathcal{Z}^{i}} and $\tilde{\pi}^{\mathcal{Z}^{i}}_{\tilde{V}_{i}}(s)$
    is unique at every $s\in\mathcal{Z}^{i}_{\mathcal{S}}$ with a binding primal constraint
    (Definition \ref{def: binding (sampling)}); and
    \item[(iv)] Assumption \ref{ass: rlp-bounded} holds with constant $\bar{U}$.
\end{enumerate}
Then, for every tolerance $\epsilon>0$, Algorithm \ref{alg: HBOBF and SHBOBF}
terminates after at most
\[
K_{\max}=\left\lceil\frac{\bar{U}-\nu^\top\tilde{V}_0}{\nu_{\min}\,\epsilon}\right\rceil
\]
iterations.
\end{theorem}

The proof of Theorem \ref{thm: convergence of HBOBF} is provided in~\nameref{app:A}. This theorem guarantees that Algorithm \ref{alg: HBOBF and SHBOBF} halts after finitely many iterations, but termination is a statement about the stopping rule, not about solution quality. The algorithm stops as soon as the empirical error $\|\mathcal{L}_{\mathcal{Y}}\tilde{V}_K-\tilde{V}_K\|_\infty$ falls below a prescribed tolerance $\epsilon$, and this residual is by construction measured only on $\mathcal{Y}_{\mathcal{S}}$. A natural question, therefore, remains about what this $\epsilon$-tolerance implies regarding the true Bellman residual $\mathcal{T}V-V$. Theorem \ref{thm: bellman residual at termination} makes this gap explicit. We emphasize that the resulting bound is intentionally loose. It is not designed to certify a tight performance guarantee, but to expose which quantities govern the residual and therefore where the analysis (and the algorithm) should be strengthened.

\begin{theorem}[Bellman residual at a fixed iteration]\label{thm: bellman residual at termination}
Fix an iteration index $k\ge0$ of Algorithm \ref{alg: HBOBF and SHBOBF} and set
$V:=\tilde{V}_k=\Phi_k\tilde{w}_k$. Let $\xi,\delta\in(0,1)$, suppose $\tilde{w}_k$ is the unique optimal solution of \RLP{\Phi_k}{\mathcal{Z}^{k}}, and suppose $M$, $\mathcal{X}^{k}$, $\mathcal{Y}$, and $\Phi_k$ satisfy the hypotheses of Corollary \ref{cor: violations under Z} at $(\xi,\delta)$ with $\ell_k$ BFs. Let $\tilde{\pi}$ be the standard greedy policy with respect to $V$ over $\mathcal{A}$, and define
\begin{equation}
\psi_{\tilde{\pi}}(s):=Z\,\psi(s,\tilde{\pi}(s)),\qquad
Z:=\left(\sum_{s\in\mathcal{S}}\psi(s,\tilde{\pi}(s))\right)^{-1},\qquad s\in\mathcal{S},
\label{eq: psi tilde pi}
\end{equation}
where the scalar $Z$ is a normalizing constant. For any subset $\mathcal{C}\subseteq\mathcal{S}$, let $\psi_{\tilde{\pi}}(\mathcal{C}):=\sum_{s\in \mathcal{C}}\psi_{\tilde{\pi}}(s)$. Then, with probability at least $1-\delta$,
\begin{align}
\|\mathcal{T}V-V\|_{1,\psi_{\tilde{\pi}}}
&\leq
\psi_{\tilde{\pi}}(\mathcal{Y}_{\mathcal{S}}) \, \|\mathcal{L}_{\mathcal{Y}}V-V\|_\infty
+\left(1-\psi_{\tilde{\pi}}(\mathcal{Y}_{\mathcal{S}})\right)
\|(\mathcal{T}V-V)^+\|_\infty
+Z\xi\|(\mathcal{T}V-V)^-\|_\infty .
\label{eq: bellman residual bound}
\end{align}
\end{theorem}

The proof of Theorem \ref{thm: bellman residual at termination} is provided in~\nameref{app:A}. Each term of (\ref{eq: bellman residual bound}) has a distinct source and a distinct degree of controllability. The first is the direct consequence of Algorithm \ref{alg: HBOBF and SHBOBF} and is the only one that shrinks by running the algorithm for more iterations. The second is the price of whatever mass $\psi_{\tilde\pi}$ places outside $\mathcal{Y}_{\mathcal{S}}$, where the hybrid operator imposes no control. It is unaffected by $\epsilon$ and is in general the dominant term. The third is the scenario-optimization contribution, controlled by the violation level $\xi$ and hence by the sample size $M$. When $\psi_{\tilde\pi}$ concentrates on $\mathcal{Y}_{\mathcal{S}}$, and in particular when $\mathcal{S}$ is finite with $\mathcal{Y}_{\mathcal{S}}=\mathcal{S}$, the second term vanishes and the residual is governed by $\epsilon$ and the $\xi$-rare violations alone. The bound is therefore diagnostic rather than tight. It identifies when and why the certificate is strong, without implying a small Bellman error uniformly over $\mathcal{S}$.

\subsection{Discussions}
\label{sec: Discussions}

Theorems \ref{thm: major improvement (with sampling)}, \ref{thm: convergence of HBOBF}, and \ref{thm: bellman residual at termination} together describe what the hybrid sampling scheme can, and cannot, deliver. The arguments behind Theorems \ref{thm: major improvement (with sampling)} and \ref{thm: convergence of HBOBF} use the finiteness of $\mathcal{S}$ only through the finiteness of $\mathcal{Y}_{\mathcal{S}}$ and $\mathcal{Z}$. The framework therefore provides at each basis-expansion step, an improvement certificate that holds with probability at least $1-\delta$ both in the classical finite-state setting and in continuous (see \online{app: continuous} for more details) or otherwise intractably large state spaces, where purely tabular LP-based ADP methods do not apply at all.

Termination and per-iteration improvement do not, on their own, imply that the returned $V$ is uniformly close to $V^*$. By (\ref{eq: bellman residual bound}), the bound collapses toward a uniform guarantee precisely when $\psi_{\tilde\pi}(\mathcal{Y}_{\mathcal{S}})\to1$, and that condition is less elusive than it appears. $\psi_{\tilde\pi}(s)\le Z\,\psi_{\mathcal{S}}(s)$ for \emph{any} greedy policy $\tilde\pi$, where $\psi_{\mathcal{S}}$ is the state marginal of $\psi$, so covering $\operatorname{supp}(\psi_{\mathcal{S}})$ covers $\operatorname{supp}(\psi_{\tilde\pi})$ whichever policy the algorithm returns, and $\mathcal{Y}_{\mathcal{S}}$ may be built accordingly. The genuine limitation is thus an accuracy--tractability tension rather than a structural disconnect. Driving $\psi_{\tilde\pi}(\mathcal{Y}_{\mathcal{S}})$ toward one calls for a dense $\mathcal{Y}_{\mathcal{S}}$, whereas the purpose of $\mathcal{Y}$ is to keep $\mathcal{L}_{\mathcal{Y}}$ cheap. Two routes to relaxing it are sketched in
\online{app: future}.

More broadly, the paper decouples two roles that LP-based ADP has generally conflated: certifying feasibility of the approximate value function and generating basis functions through Bellman backups. One may therefore begin from a modest $\Phi_0$, for which classical suboptimality bounds are available (\citealp{de_farias_linear_2003,de_farias_constraint_2004}), and sharpen the approximation exactly where it is wanted, iteration by iteration, through Theorem \ref{thm: major improvement (with sampling)}. To our knowledge this unconstrained enlargement of a deterministic basis-generating sample has not previously been exploited in the LP-based ADP literature, and we regard it as the central practical advantage of the framework. A comparison with the sample-complexity results known for $Q$-learning and temporal-difference learning is given in \online{oa: discussion}.

\section{Numerical Studies} \label{sec: Numerical Studies}

We apply our methodologies to two application domains: perishable inventory control (PIC) and queueing control (QC) problems. To evaluate performance, we compare our BF learning methods against the Fourier and ReLU random BFs proposed by \cite{pakiman2025selfguided}.

In the Fourier approach, BFs are generated as $\phi(s) = \text{cos}(\theta_0 + \sum_{i=1}^{\mathcal{I}} s_i\theta_i)$, for all $s\in\mathcal{S}$, where $\mathcal{I}$ denotes the dimensionality of $s$, $\theta_i \sim \text{normal}(0,\sigma)$ for $i = 1, \cdots,\mathcal{I}$, and $\theta_0 \sim \text{uniform}(-\pi_{const}, \pi_{const})$. Here, $\pi_{const}$ refers to Archimedes' constant, and we set $\sigma = 1$ in all experiments. In the ReLU approach, BFs are defined as $\phi(s) = \max\left\{0, \theta_0 + \sum_{i=1}^{\mathcal{I}} s_i\theta_i\right\}$, for all $s \in \mathcal{S}$, where $\theta$ is sampled uniformly from $\left(\mathcal{I}+1\right)$-dimensional unit sphere.

We also consider an application-specific ``Fixed BF'' benchmark for each problem, assembled from the expert-recommended bases in the corresponding literature. In all experiments, the state-relevance weight vector $\nu$ is uniform over the state
space, or over $\mathcal{Y}_{\mathcal{S}}$ when sampling schemes are employed. In the sampled experiments, each application supplies a \emph{proposal} $\mathbb{Q}$ on $(\mathcal{S}\times\mathcal{A})$, specified by a state proposal $\mathbb{Q}_{\mathcal{S}}$ and a conditional action proposal $\mathbb{Q}_{\mathcal{A}}(\cdot\mid s)$ (see \online{oa: sampling} for details). A candidate pair $(s,a)\sim\mathbb{Q}$ is accepted if it lies in the admissible set $\mathcal{F}$. Otherwise both coordinates are discarded and the pair is redrawn. Accepted draws are thus i.i.d.\ from the single distribution $\psi=\mathbb{Q}(\cdot\mid\mathcal{F})$, as Proposition~\ref{prop: probability of violations} requires. A fresh $\mathcal{X}^{k}$ is drawn at each iteration, keeping $\Phi_k$ independent of the sample to which Corollary~\ref{cor: violations under Z} is applied. The curated sample $\mathcal{Y}$ is built once beforehand on a dedicated random stream: $\mathcal{Y}_{\mathcal{S}}$ combines chain-walk states under randomized predefined policies with dispersed states from $\mathbb{Q}_{\mathcal{S}}$, and $\mathcal{Y}_{\mathcal{A}}(s)$ attaches a structured set of feasible actions to each retained state. Instance parameters, demand and arrival specifications, the composition of the Fixed BF vectors, and the full sampling design are reported in Sections \ref{oa: implementation} and \ref{oa: sampling} of the e-companion.

\subsection{Perishable Inventory Control (PIC) with Positive Lead Time}

The PIC problem has been extensively studied in the literature, with numerous extensions proposed (\cite{kempf_planning_2011, nahmias_perishable_2011}).  In this application, we focus on the single-product, periodic-review variant of the problem. We assume a positive lead time and that any unfulfilled demand is lost.

The state space of a PIC problem with positive lead time consists of two components. The first component represents the on-hand inventory and its remaining shelf life, denoted by ${x} = (x_m, \cdots, x_2, x_1)$, where newly received items have the maximum remaining shelf life of $m$ periods. The second component corresponds to incoming inventory, denoted by ${z}= (z_l, \cdots, z_2, z_1)$, where $z_i$ indicates the order quantity scheduled to arrive in $i$ periods. In the finite setting, $z \in \mathbb{Z}_+^l$ and $x\in\mathbb{Z}_+^m$. We assume a delivery lead time of $l$ periods for all new orders. Combining these components, the state $s$ is defined as $s = (z_l, \cdots, z_1, x_m, \cdots, x_1)$.

The action $ a \in \mathbb{Z}_{+}$ represents the non-negative order quantity, subject to an upper limit of $\Bar{a}$. Given the state and action, the cost function is defined as
\begin{align*}
    c(s,a) = f\left(1_{a>0}\right)+\mathbb{E}_D\left[h \left(\sum_{i=1}^m x_i - D\right)^+ + \eta \left(D-\sum_{i=1}^m x_i \right)^+ +\theta \left(x_1 - D\right)^+\right],
\end{align*}
where $f$, $h$, $\eta,$ and $\theta$ denote the unit fixed ordering, holding, shortage, and wastage costs. The function $(\cdot)^+$ represents the positive part function, defined as $(\cdot)^+ = \max\{\cdot,0\}$. The parameter $D$ denotes the stochastic demand with a known distribution. In our setting, $D$ follows a negative binomial distribution characterized by parameters $(\iota, q)$ and truncated at $\bar{D}$. We define the state space as \begin{align}
\mathcal{S} = \left\{ s\bigg| \sum_{j=1}^l z_j + \sum_{i=1}^m x_i \leq G \right\}, \label{eq: PIC state space} \end{align}  where $ G$ denotes the smallest threshold such that $ \mathbb{P}(\sum_{t=1}^{l+1} D_t \leq G) \geq \alpha$. The parameter $\alpha$ specifies the service level, i.e., the probability that the demand does not exceed the available supply. Similarly, the action space is defined as \begin{align}
\mathcal{A}_s = \left\{ a\bigg| a + \sum_{j=1}^l z_j + \sum_{i=2}^m x_i \leq G    \right\}. \label{eq: PIC action space} \end{align} In the finite setting, we have $a \in \mathbb{Z}_+$.

\subsubsection{Evaluation Over the Full Set of State--Action Pairs}

\begin{figure}[h]
\FIGURE{
    \includegraphics[width=0.4\textwidth]{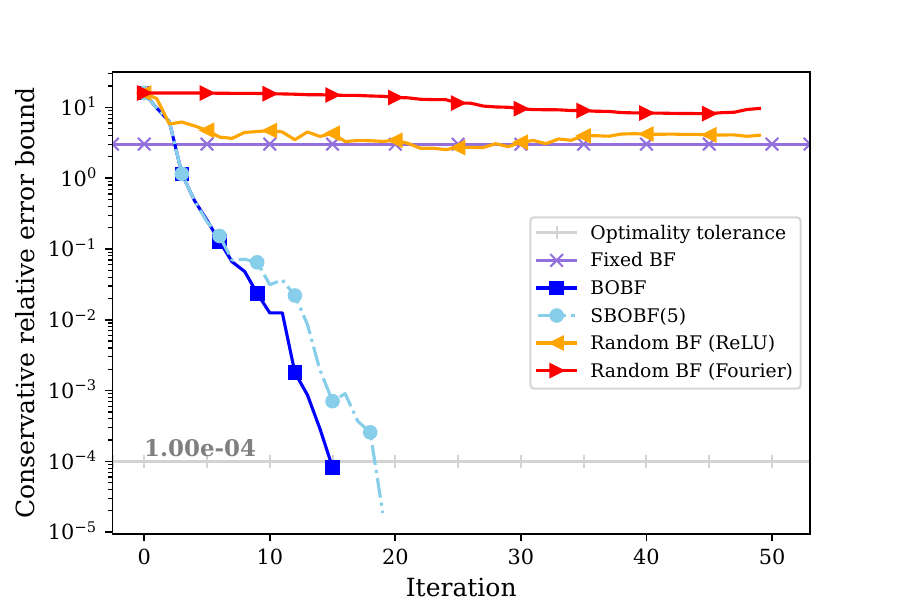}
}{\normalfont Conservative relative error bound $\frac{1}{(1-\gamma) \nu^\top\tilde{V}_k}\parallel \mathcal{T}\tilde{V}_k-\tilde{V}_k\parallel_\infty$ of all approaches on a PIC instance with $(f, h, \eta, \theta) = (100, 1, 20, 5)$, $(l, m) = (1, 4)$, and $\gamma=0.99$. For the random approaches, five BFs are generated per iteration. \label{fig: PIC BOBF}}{}
\end{figure}

Figure \ref{fig: PIC BOBF} illustrates the conservative relative optimality gap for all solution methods for an instance characterized by $(f, h, \eta, \theta) = (100, 1, 20, 5)$, $(l, m) = (1, 4)$, and $\gamma=0.99$. Both BOBF and SBOBF with $n_{max} = 5$ (denoted as SBOBF(5)) converge within a few iterations, whereas all other benchmark methods fail to achieve convergence. The Random BF (Fourier) approach yields only a slight improvement in the optimality gap, while the Random BF (ReLU) approach demonstrates a more noticeable reduction. It decreases the gap during the initial iterations, but the gap remains relatively high thereafter. Interestingly, the Fixed BF approach performs comparably to the Random BF (ReLU) method while using significantly fewer BFs, 29 versus a maximum of 251. Even under a conservative error bound, BOBF consistently outperforms SBOBF, since incorporating more BFs enables a larger improvement of the approximated value function per iteration and hence a tighter worst-case bound. The effect of $n_{max}$ on SBOBF, including a comparison of SBOBF(1) with value iteration, is reported in \online{oa: nmax}.

The first block of Table \ref{tab: summary} reports the conservative relative error bounds for all solution methods evaluated over 20 iterations, averaged within each instance group. The BOBF approach, followed by SBOBF(5), consistently outperforms the baseline methods across all scenarios, underscoring the effectiveness of our BF learning strategies. The instance-level results, given in \online{oa: full results}, additionally reveal that the Fixed BF approach achieves superior performance bounds compared to Random BF generation in nearly all cases, highlighting the advantage of expert-defined BFs over those generated randomly.

\subsubsection{Evaluation Over a Sampled Set of State--Action Pairs}

In this section, we address large-scale continuous PIC instances using constraint sampling with $M=50{,}000$. Here $\mathbb{Q}_{\mathcal{S}}$ is a two-component mixture over $(z_l,\cdots,z_1,x_m,\cdots,x_1)$, demand-centered with weight $0.8$ and dispersed uniform with weight $0.2$, and $\mathbb{Q}_{\mathcal{A}}(\cdot\mid s)$ is a mixture with the same weights whose shaped component targets a jittered order-up-to level (see \online{oa: sampling} for more information). Lastly, $\mathcal{F}$ is given by (\ref{eq: PIC state space}) and (\ref{eq: PIC action space}) at $\alpha=0.99$. The chain walks generating $\mathcal{Y}_{\mathcal{S}}$ run under randomized no-order, base-stock, and $(s,S)$ policies, and $\mathcal{Y}_{\mathcal{A}}(s)$ is an evenly spaced grid over $[0,a_{\max}(s)]$, where $a_{\max}(s)$ denotes  the largest order feasible at $s$. The same instances under continuously distributed demand are reported in Section~\ref{oa: PIC continuous demand}.

\begin{figure}[htbp!]
\FIGURE{
    \includegraphics[width=0.4\textwidth]{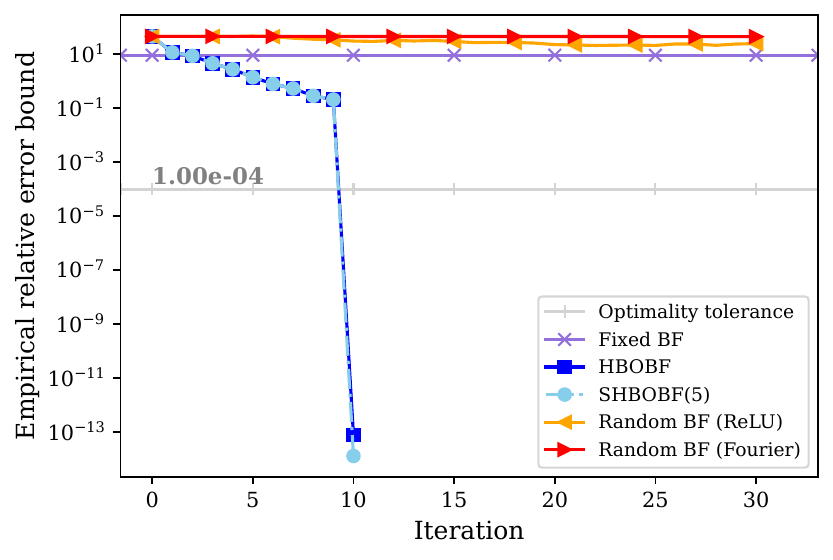}
}{\normalfont Empirical relative error bound $\frac{1}{(1-\gamma) \nu^\top\tilde{V}_k}\parallel \mathcal{L}_{\mathcal{Y}}\tilde{V}_k - \tilde{V}_k \parallel_\infty$ for all solution methods executed over 30 iterations in a continuous PIC instance with parameter settings $(f, h, \eta, \theta) = (100, 1, 20, 5)$, $(l, m) = (6, 10)$, and $\gamma=0.99$. For the random approaches, five BFs are generated per iteration. \label{fig: PIC HBOBF}}{}
\end{figure}

Figure \ref{fig: PIC HBOBF} illustrates the empirical error bound for a PIC instance with $(l, m) = (6, 10)$. Our HBOBF approach achieves the optimality tolerance in ten iterations, as does SHBOBF(5), whereas other benchmark methods exhibit substantially higher error bounds. In particular, the random BF expansion methods fail to produce any significant improvement over their initial error bounds even after 30 iterations. The second block of Table \ref{tab: summary} reports the error bounds across a range of instances, where our methods consistently outperform all benchmark approaches.

\subsection{Queueing Control (QC)}

In this section, we implement the QC problem as formulated by \cite{samiedaluie2024approximate}. The problem considers an $M/M/c/c$ queueing system that serves $J$ customer classes, each arriving according to a Poisson process. Because the system's capacity equals the number of servers, no queue can form; any arriving customer is rejected when all servers are occupied. Upon completing service for a customer of class $j$, a profit of $p_j$ is realized. The parameters $\lambda_j'$ and $\omega_j'$ denote the arrival and service rates for customer class $j$, respectively. Let $\kappa$ represent both the number of servers and the maximum number of customers allowed in the system. Given this capacity constraint, the objective is to maximize total profit by deciding whether to accept or reject customers from multiple classes.

The problem is formulated as a continuous-time, infinite-horizon discounted MDP, which we convert into a discrete-time problem. To achieve this, we apply uniformization to the arrival and service rates, ensuring a consistent decision epoch structure. Decisions are made upon customer arrivals. The state $s$ denotes the number of customers in the system, categorized by class, i.e., $s=(s_1,s_2, \cdots, s_J)$. Consequently, the state space is defined as \begin{align}
 \mathcal{S} = \left\{ s \bigg| s\in \mathbb{Z}_+^J, \sum_{j=1}^J s_j \leq \kappa \right\}. \label{eq: QC state space}\end{align} The action $a = (a_1, a_2, \cdots, a_J)$ specifies whether each customer class is accepted upon arrival, where $a_j\in \{0,1\}$ for all $j \in {1,\cdots, J}$. Accordingly, the action space is defined as \begin{align}
 \mathcal{A}(s) = \left\{a \bigg| a\in \{0,1\}^J, a_i + \sum_{j=1}^J s_j \leq \kappa \quad \forall i\in\{1,\cdots, J\}\right\}. \label{eq: QC action space}\end{align} The optimality equations for the corresponding discrete-time dynamic programming problem can be expressed as \[V(s) = \max_{a\in\mathcal{A}(s)}\left\{\sum_{j=1}^J \lambda_j \left(p_ja_j + V(s+a_j{e}_j)\right)+\sum_{j=1}^Js_j\omega_jV(s-{e}_j) + \left( \vartheta - \sum_{j=1}^J \lambda_j - \sum_{j=1}^J s_j\omega_j \right)V(s)\right\}, \] where $\lambda_j = \frac{\lambda_j'}{\zeta + \vartheta '}$, $\omega_j = \frac{\omega_j'}{\zeta + \vartheta '}$, $\vartheta = \frac{\vartheta'}{\zeta + \vartheta '}$, $\zeta$ denotes the continuous-time discount rate, and $\vartheta' = \sum_{j=1}^J \lambda_j ' + \kappa \max_{j \in \{1,\cdots,J\}}\{\omega_j'\}$ (\cite{samiedaluie2024approximate}). The vector ${e}_j$ is an all-zero vector except for its $j$-th element, which equals 1. The time is scaled such that $\zeta+\vartheta' = 1$. For the fixed BF approach we employ the affine basis $\Phi(s)=[1,s_1,\cdots,s_J]$ (see \cite{samiedaluie2024approximate}).

\subsubsection{Evaluation Over the Full Set of State--Action Pairs} 
\begin{figure}[htbp!]
  \caption{\normalfont Conservative relative error bound $\frac{1}{(1-\gamma) \nu^\top\tilde{V}_k}\parallel \mathcal{T}\tilde{V}_k-\tilde{V}_k\parallel_\infty$ for all solution methods executed over 100 iterations in a QC instance with parameter settings  $(J, \kappa) = (3, 15)$. The service load is set to (a) $\rho = 0.8$ and (b) $\rho = 10$. For the random approaches, five BFs are generated per iteration.}
  \centering
  \begin{subfigure}[t]{0.4\textwidth}
    \centering
    \includegraphics[width=\textwidth]{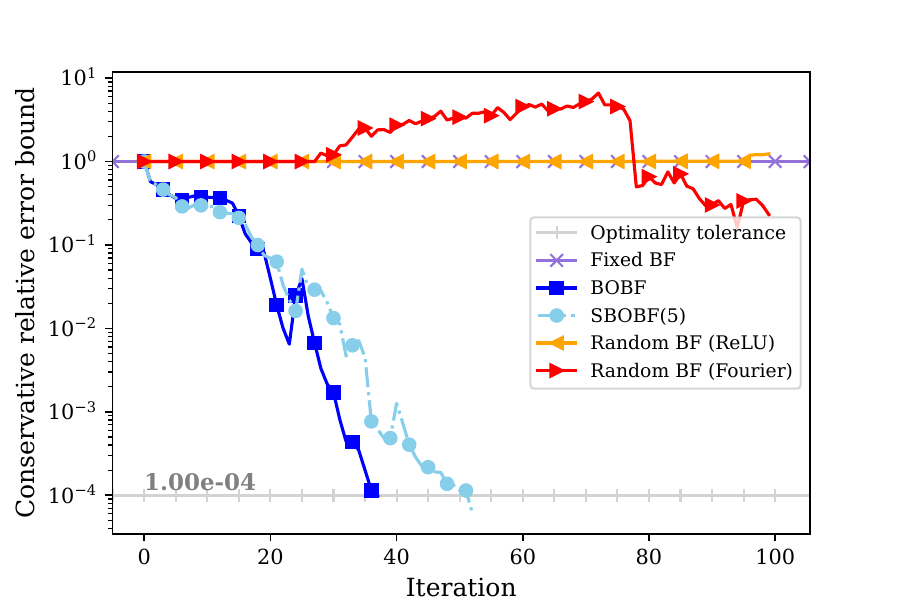}
    \caption{\small $\rho = 0.8$.}
    \label{fig:plot1}
  \end{subfigure}
  \hfill
  \begin{subfigure}[t]{0.4\textwidth}
    \centering
    \includegraphics[width=\textwidth]{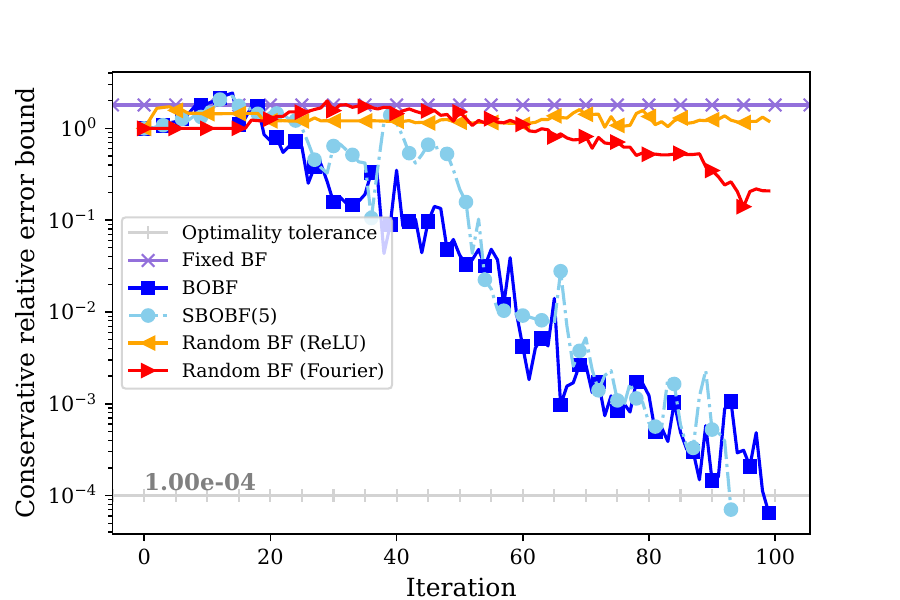}
    \caption{ \small $\rho = 10$.}
    \label{fig:plot2}
  \end{subfigure}
  \label{fig: QC Benchmark}
\end{figure}

Figure \ref{fig: QC Benchmark} illustrates the performance of our solution methods for an instance with $(J, \kappa) = (3, 15)$ under service loads of (a) 0.8 and (b) 10. The BOBF and SBOBF(5) learning approaches demonstrate strong performance, converging within at most 100 iterations. In contrast, the random and fixed BFs yield high error bounds.

Further insights can be drawn from Figure \ref{fig: QC Benchmark}. Under a low service load (part (a)), both BOBF and SBOBF approaches steadily reduce the optimality gap until convergence, with BOBF outperforming in most iterations. However, under a high service load (part (b)), convergence becomes more challenging for both methods. Notably, SBOBF(5) achieves convergence slightly earlier than BOBF, despite using only five BFs compared to nearly one hundred BFs employed by BOBF in the final iteration. This observation underscores the efficiency of SBOBF(5) in high-load scenarios.

Another noteworthy observation from Figure \ref{fig: QC Benchmark} is the sharp increase in the conservative relative error bound during certain iterations. This rise may be attributed to an increase in $\parallel \mathcal{T}\tilde{V}_k - \tilde{V}_k \parallel_\infty$. However, this is not the sole contributing factor. Specifically, the increase can also be partially explained by a decrease in the denominator of the conservative relative error bound, i.e., $\nu^\top \tilde{V}_k = \nu^\top \Phi_k\tilde{w}_k$, since the objective of the ALP in the QC problem is minimization.

The third block of Table \ref{tab: summary} summarizes the performance of the solution methods for the QC problem across a broad range of scenarios, reporting the best conservative relative error bound achieved within 100 iterations. Consistent with the results for the PIC problem, both BOBF and SBOBF learning approaches significantly outperform the alternative methods. Our experiments further reveal that higher service loads generally lead to larger optimality bounds, indicating that BOBF and SBOBF require more iterations to converge. An additional noteworthy finding is that, although BOBF outperforms SBOBF(5) in most individual experiments, the average performance of SBOBF(5) is slightly superior, and in particular under high service loads SBOBF(5) achieves a better performance bound. This result is significant because it demonstrates that the convergence of SBOBF(5) is comparable to that of BOBF, where the net approximation improvement per iteration can be substantial (see Theorem \ref{thm: major improvement}).

\subsubsection{Evaluation Over a Sampled Set of State--Action Pairs}

For the large-scale QC instances, $\mathbb{Q}_{\mathcal{S}}$ constructs a state $s=(s_1,\cdots,s_J)$ by sequentially allocating the capacity of $\kappa$ servers across
classes, $\mathbb{Q}_{\mathcal{A}}$ draws each $a_j$ independently and uniformly on
$\{0,1\}$, and $\mathcal{F}$ is given by (\ref{eq: QC action space}); we again set
$M=50{,}000$. The chain walks generating $\mathcal{Y}_{\mathcal{S}}$ run under randomized
static priority-threshold policies, and $\mathcal{Y}_{\mathcal{A}}(s)$ collects the
reject-all and admit-all actions, the priority-threshold family, and uniform random fills,
restricted to those admissible at $s$. The construction is detailed in \online{oa: sampling}.
\begin{figure}[htbp!]
  \caption{\normalfont Empirical relative error bound $\frac{1}{(1-\gamma) \nu^\top\tilde{V}_k}\parallel \mathcal{L}_{\mathcal{Y}}\tilde{V}_k - \tilde{V}_k \parallel_\infty$ for all solution methods executed over 50 iterations in a QC instance with parameter settings $(J, \kappa) = (8, 80)$. The service load is set to (a) $\rho = 3$ and (b) $\rho = 10$. For the random approaches, five BFs are generated per iteration.}
  \centering
  \begin{subfigure}[t]{0.4\textwidth}
    \centering
    \includegraphics[width=\textwidth]{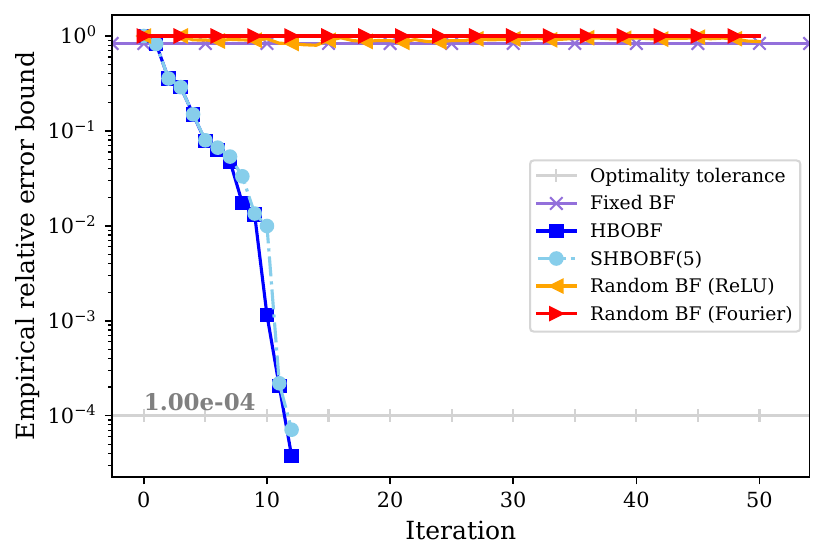}
    \caption{\small $\rho = 3$.}
    \label{fig:plot1-oa}
  \end{subfigure}
  \hfill
  \begin{subfigure}[t]{0.4\textwidth}
    \centering
    \includegraphics[width=\textwidth]{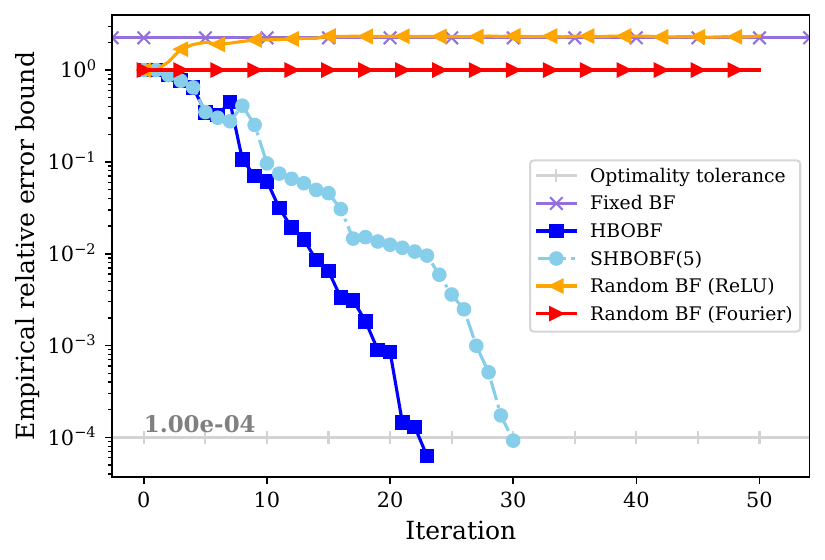}
    \caption{ \small $\rho = 10$.}
    \label{fig:plot2-oa}
  \end{subfigure}
  \label{fig: QC Benchmark (sampling)}
\end{figure}

\begin{table}[htbp!]
\centering
\caption{\normalfont Relative error bound of all solution methods, averaged within each instance group. The first and third blocks report the conservative bound $\frac{1}{(1-\gamma) \nu^\top\tilde{V}_k}\parallel \mathcal{T}\tilde{V}_k-\tilde{V}_k\parallel_\infty$ over the full set of state--action pairs; the second and fourth report the empirical bound $\frac{1}{(1-\gamma) \nu^\top\tilde{V}_k}\parallel \mathcal{L}_{\mathcal{Y}}\tilde{V}_k-\tilde{V}_k\parallel_\infty$ under constraint sampling.  For the random approaches, five BFs are generated per iteration. Instance-level results are reported in \online{oa: full results}.} 
\label{tab: summary}
\scriptsize
\begin{tabular}{|ll|ccccc|}
\hline
\multicolumn{2}{|c|}{Setting and instance group} & \begin{tabular}[c]{@{}c@{}}BOBF /\\ HBOBF\end{tabular} & \begin{tabular}[c]{@{}c@{}}SBOBF(5) /\\ SHBOBF(5)\end{tabular} & Fixed & \begin{tabular}[c]{@{}c@{}}Random\\ (ReLU)\end{tabular} & \begin{tabular}[c]{@{}c@{}}Random\\ (Fourier)\end{tabular} \\ \hline
\multirow{3}{*}{\begin{tabular}[c]{@{}l@{}}PIC,\\ full\end{tabular}} & $(l,m)=(2,2)$ & \textbf{1.31e-07} & 1.43e-05 & 4.06e+00 & 6.78e+00 & 3.20e+01 \\
 & $(l,m)=(1,4)$ & \textbf{3.58e-07} & 1.73e-04 & 3.09e+00 & 6.11e+00 & 2.89e+01 \\
 & Average & \textbf{2.45e-07} & 9.36e-05 & 3.57e+00 & 6.44e+00 & 3.04e+01 \\ \hline
\multirow{3}{*}{\begin{tabular}[c]{@{}l@{}}PIC,\\ sampled\end{tabular}} & $(l,m)=(5,6)$ & \textbf{3.07e-05} & \textbf{3.07e-05} & 3.06e+01 & 5.56e+01 & 1.23e+02 \\
 & $(l,m)=(6,10)$ & \textbf{2.04e-05} & \textbf{2.04e-05} & 1.94e+01 & 6.12e+01 & 9.99e+01 \\
 & Average & \textbf{2.56e-05} & \textbf{2.56e-05} & 2.50e+01 & 5.84e+01 & 1.11e+02 \\ \hline
\multirow{3}{*}{\begin{tabular}[c]{@{}l@{}}QC,\\ full\end{tabular}} & $(J,\kappa)=(3,15)$ & \textbf{1.48e-05} & 2.90e-05 & 1.15e+00 & 2.02e-01 & 8.58e-01 \\
 & $(J,\kappa)=(4,25)$ & 1.97e-03 & \textbf{1.13e-03} & 1.53e+00 & 1.00e+00 & 1.00e+00 \\
 & Average & 9.94e-04 & \textbf{5.78e-04} & 1.34e+00 & 6.01e-01 & 9.29e-01 \\ \hline
\multirow{3}{*}{\begin{tabular}[c]{@{}l@{}}QC,\\ sampled\end{tabular}} & $(J,\kappa)=(8,80)$ & \textbf{2.76e-05} & 3.28e-05 & 1.37e+00 & 9.60e-01 & 1.00e+00 \\
 & $(J,\kappa)=(10,100)$ & \textbf{2.02e-05} & 4.83e-05 & 1.29e+00 & 9.23e-01 & 1.00e+00 \\
 & Average & \textbf{2.39e-05} & 4.05e-05 & 1.33e+00 & 9.42e-01 & 1.00e+00 \\ \hline
\end{tabular}
\end{table}
On an instance with $(J,\kappa)=(8,80)$, both HBOBF and SHBOBF(5) achieve the optimality tolerance in 12 to 32 iterations at service loads $\rho=3$ and $\rho=10$, while the other benchmark methods exhibit substantially higher error levels (Figure \ref{fig: QC Benchmark (sampling)}). Similar patterns are observed across multiple scenarios, as reported in the fourth block of Table \ref{tab: summary}.

\section{Conclusion} \label{sec: Conclusion}

In this paper, we introduced a BF learning approach for ALP and showed that it converges to the optimal value function $V^*$, which is a novel contribution to the literature. Our methodology uses the Bellman operator as the foundation for two learning mechanisms, BOBF and SBOBF. We proved that appending a BOBF improves the approximation by at least the weighted Bellman error, together with a dual-weighted term scaled by $\gamma/(1-\gamma)$, so that the improvement is strict whenever the Bellman error is non-zero. Iterating the construction returns an $\epsilon$-optimal value estimate. The SBOBF approach caps the number of BFs retained and therefore the computational effort per iteration, while achieving convergence with performance comparable to BOBF.

We then incorporated constraint sampling and developed the HBOBF and SHBOBF learning approaches, which substantially enhance scalability. Within this scheme we established, with probability at least $1-\delta$, a bound on the net approximation improvement obtained from a single HBOBF. We showed that the procedure terminates after finitely many iterations, and we bounded the Bellman residual at termination. That last bound is deliberately diagnostic rather than tight. It makes explicit that termination alone does not certify a small Bellman error uniformly over the state space. We discussed the conditions that can remedy this limitation in Section \ref{sec: Discussions} and \online{app: future}.

We compared our BF learning approaches against existing methods that rely on fixed or randomly generated BFs. In both applications considered, the PIC and QC problems, the gap of our methods decreases substantially within a few iterations for both ALP and RLP, whereas the benchmarks exhibit persistently high gaps.

Two limitations point to future work. The first is a tension between accuracy and tractability. Tightening the residual bound calls for a dense curated sample, whereas the purpose of that sample is to keep the hybrid Bellman operator cheap. Extrapolating the values obtained on the curated states to unsampled ones, or learning a statistical surrogate for the operator off the curated set, would relax this tension, and working out the corresponding approximation-error and sample-complexity trade-offs is a substantial undertaking in its own right. The second is the constraint reduction technique itself. Alternatives such as constraint violation learning (\cite{lin_revisiting_2020}) may prove complementary. 

\section*{Acknowledgements}
This research was supported by the Natural Sciences and Engineering Research Council of Canada (NSERC) through Discovery Grant RGPIN-2026-04777. This research was enabled in part by support provided by Compute Ontario (computeontario.ca) and the Digital Research Alliance of Canada (alliancecan.ca) through Resource Allocation Competition award RRG 2026 \#3903.

Computations were performed on the Trillium supercomputer at the SciNet HPC Consortium (\cite{ponce2019deploying,loken2010scinet}). SciNet is funded by Innovation, Science and Economic Development Canada; the Digital Research Alliance of Canada; the Ontario Research Fund: Research Excellence; and the University of Toronto.

The authors used generative AI tools to improve the language of the manuscript and to assist with coding the experiments. All outputs were reviewed and verified by the authors, who take full responsibility for the content of this work.

\newpage
\putbib[references]

\newpage

\section*{\label{app:A} Appendix A}

This appendix presents the remaining lemmas and proofs referenced in our study.

\textit{Proof of Theorem \ref{thm: major improvement}.}
Throughout this proof we abbreviate $\tilde{\pi}:=\tilde{\pi}_{k+1}$ and
$\tilde{u}:=\tilde{u}^{k+1}_{\tilde{\pi}_{k+1}}$.

\emph{Step 1 (the dual identity at the new BF).} By hypothesis, $\tilde{\pi}(s)$ is unique at every state with a binding primal constraint (Definition \ref{def: binding}), so Property \ref{prop: dual structure} applies to \DUAL{\Phi_{k+1}} and gives
\begin{align}
\tilde{u}^\top c_{\tilde{\pi}} &= \nu^\top\tilde{V}_{k+1}, \label{eq: strong duality ALP}\\
\tilde{u}^\top\left(I-\gamma P_{\tilde{\pi}}\right)\Phi_{k+1} &= \nu^\top\Phi_{k+1}.
\label{eq: dual feasibility ALP}
\end{align}
The identity (\ref{eq: dual feasibility ALP}) holds BF by BF, i.e., for every $\phi \in \Phi_{k+1}$. Applying it to the last BF of $\Phi_{k+1}$, namely $\phi=\mathcal{T}\tilde{V}_k$, and rearranging gives
\begin{equation}
\tilde{u}^\top\mathcal{T}\tilde{V}_k
= \nu^\top\mathcal{T}\tilde{V}_k
+\gamma\,\tilde{u}^\top P_{\tilde{\pi}}\mathcal{T}\tilde{V}_k .
\label{eq: feasibility 2 (revised)}
\end{equation}

\emph{Step 2 (bounding the same quantity through the greedy policy).} Since $\mathcal{T}\tilde{V}_k=\min_{\pi}\{c_\pi+\gamma P_\pi\tilde{V}_k\} \le c_{\tilde{\pi}}+\gamma P_{\tilde{\pi}}\tilde{V}_k$ component-wise and $\tilde{u}\ge0$,
\begin{equation}
\tilde{u}^\top\mathcal{T}\tilde{V}_k
\ \le\ \tilde{u}^\top c_{\tilde{\pi}}+\gamma\,\tilde{u}^\top P_{\tilde{\pi}}\tilde{V}_k
\ =\ \nu^\top\tilde{V}_{k+1}+\gamma\,\tilde{u}^\top P_{\tilde{\pi}}\tilde{V}_k ,
\label{eq: eq02}
\end{equation}
the equality being the strong duality (\ref{eq: strong duality ALP}).

\emph{Step 3 (matching the two expressions).} Substituting (\ref{eq: feasibility 2 (revised)}) into the left-hand side of (\ref{eq: eq02}) and subtracting $\nu^\top\tilde{V}_k$ from both sides,
\begin{align}
\nu^\top\tilde{V}_{k+1}-\nu^\top\tilde{V}_{k}
&\ \geq\ \nu^\top\big(\mathcal{T}\tilde{V}_k-\tilde{V}_k\big)
+\gamma\,\tilde{u}^\top P_{\tilde{\pi}}\big(\mathcal{T}\tilde{V}_k-\tilde{V}_k\big) \notag\\
&\ =\ \nu^\top\big(\mathcal{T}\tilde{V}_k-\tilde{V}_k\big)
+\frac{\gamma}{1-\gamma}\,\mu_{k+1}^\top\big(\mathcal{T}\tilde{V}_k-\tilde{V}_k\big),
\notag
\end{align}
where $\mu_{k+1}^\top:=(1-\gamma)\tilde{u}^\top P_{\tilde{\pi}}$ is a normalized vector by
Property \ref{prop: dual structure} (part 4).

\emph{Step 4 (from objective values to approximation errors).} By (\ref{eq:fixedpointineq}) we have $\mathcal{T}\tilde{V}_k-\tilde{V}_k\ge0$ component-wise, so the two inner products on the right-hand side are the norms $\|\mathcal{T}\tilde{V}_k-\tilde{V}_k\|_{1,\nu}$ and $\|\mathcal{T}\tilde{V}_k-\tilde{V}_k\|_{1,\mu_{k+1}}$, while the left-hand side equals $\Delta_{\phi=\mathcal{T}\tilde{V}_k}\nu^\top\tilde{V}_k$ by Lemma \ref{lem: preliminaries} case (iii), which results in (\ref{eq: major improvement 1}). $\square$

\textit{Proof of Theorem \ref{thm: convergence}.} In each iteration $k$ of Algorithm \ref{alg: BOBBF}, a new BF $\phi=\mathcal{T}\tilde{V}_k$ is appended to the existing vector of BFs, forming $[\Phi_k,\mathcal{T}\tilde{V}_k]$. Under Option 2, all but the most recent $\nmax$ columns of that vector are then discarded. In either case the resulting $\Phi_{k+1}$ has $\mathcal{T}\tilde{V}_k$ as its last column, so by Lemma \ref{lem: preliminaries} case (ii) the vector $\bar{w}_{k+1}=(0,0,\cdots,0,1)^\top$ is a feasible solution to \ALP{\Phi_{k+1}}. Hence,
\begin{align}
    \nu^\top\tilde{V}_{k+1}&\geq\nu^\top \Phi_{k+1} \bar{w}_{k+1} = \nu^\top\mathcal{T}\tilde{V}_k \notag \\
    \implies \nu^\top\tilde{V}_{k+1} - \nu^\top\tilde{V}_k &\geq \nu^\top\left(\mathcal{T}\tilde{V}_k - \tilde{V}_k\right) = \left\|\mathcal{T}\tilde{V}_k - \tilde{V}_k\right\|_{1,\nu}\notag\\
    \implies \sum_{k=0}^{K-1}  \left(\nu^\top\tilde{V}_{k+1} - \nu^\top\tilde{V}_k \right) & \geq \sum_{k=0}^{K-1} \left\|\mathcal{T}\tilde{V}_k - \tilde{V}_k\right\|_{1,\nu} \notag \\
    \implies \nu^\top\tilde{V}_{K} - \nu^\top\tilde{V}_{0} &\geq \sum_{k=0}^{K-1}\left\|\mathcal{T}\tilde{V}_k - \tilde{V}_k\right\|_{1,\nu}\notag \\
    \implies \left\|V^* -\tilde{V}_{K}\right\|_{1,\nu} & \leq \left\|V^* - \tilde{V}_{0}\right\|_{1,\nu} -   \sum_{k=0}^{K-1}\left\|\mathcal{T}\tilde{V}_k - \tilde{V}_k\right\|_{1,\nu}, \label{eq: telescopic sum}
\end{align}
where the last line is concluded by Lemma \ref{lem: preliminaries} case (iii). Furthermore, the inequality (\ref{eq:fixedpointineq}) implies that $\tilde{V}_{K}\leq\mathcal{T}\tilde{V}_{K}\leq V^*$. Therefore, by Lemma \ref{lem: preliminaries} case (iii) and (iv), we conclude that \[ \left\| \mathcal{T}\tilde{V}_{K} -  \tilde{V}_{K}  \right\|_{1,\nu} \leq \left\| V^* -  \tilde{V}_{K}\right\|_{1,\nu}. \] Hence the inequality (\ref{eq: telescopic sum}) can be reformulated as
\begin{align}
       \left\| \mathcal{T}\tilde{V}_{K} -  \tilde{V}_{K}  \right\|_{1,\nu} \leq  \left\| V^* - \tilde{V}_{0}\right\|_{1,\nu} - \sum_{k=0}^{K-1}\left\|\mathcal{T}\tilde{V}_k - \tilde{V}_k\right\|_{1,\nu}, \notag
\end{align}
or equivalently,
\begin{align}
       \sum_{k=0}^{K}\left\|\mathcal{T}\tilde{V}_k - \tilde{V}_k\right\|_{1,\nu} \leq  \left\| V^* - \tilde{V}_{0}\right\|_{1,\nu} . \label{eq: Bellman error bound}
\end{align}
Based on the bound (\ref{eq: Bellman error bound}), the series converges, implying that its terms vanish asymptotically, i.e., $\left\|\mathcal{T}\tilde{V}_K - \tilde{V}_K\right\|_{1,\nu}\xrightarrow[K\to\infty]{} 0$, which concludes part a. Since $\nu$ has full support on $\mathcal{S}$ (where $\mathcal{S}$ is finite), norms are equivalent, and $\left\|\mathcal{T}\tilde{V}_K - \tilde{V}_K\right\|_{\infty}\xrightarrow[K\to\infty]{} 0$. This uniform decay guarantees that the stopping condition is met. Let $K'$ be the first iteration at which Algorithm \ref{alg: BOBBF} terminates. Then
\[
    \frac{1}{1-\gamma}\left\| \mathcal{T}\tilde{V}_{K'} - \tilde{V}_{K'}\right\|_\infty \leq \epsilon,
\] and by Lemma \ref{lem: epsilon optimal} we conclude that $\tilde{V}_{K'}$ is $\epsilon$-optimal (part b). $\square$

\begin{lemma} [Sampling correction to the $\nu$-weighted error]
\label{lem: psi to xi}
Let $\Phi$, $\mathcal{X}$, $\mathcal{Y}$, $\mathcal{Z}$, $\xi$, $\delta$, and the unique optimal
solution $\tilde{w}$ of \RLP{\Phi}{\mathcal{Z}} satisfy the hypotheses of Corollary
\ref{cor: violations under Z}. Set $\tilde{V}=\Phi\tilde{w}$ and let $\beta$ be as in Assumption
\ref{ass: psi vs d^*}. Then, with probability at least $1-\delta$,
\begin{align}
    \nu^\top (V^* - \tilde{V}) \ \geq\ \|V^* - \tilde{V}\|_{1,\nu}
    - \frac{2 \beta\, \xi}{1-\gamma}
    \left\|(\mathcal{T}_{\pi^*}\tilde{V} - \tilde{V})^-\right\|_\infty ,
    \label{eq: psi to xi}
\end{align}
where $\pi^*$ is the optimal policy.
\end{lemma}

\textit{Proof of Lemma \ref{lem: psi to xi}}. Throughout this proof write
$\Lambda:=(I-\gamma P_{\pi^*})^{-1}\ge 0$ and
$r':=\mathcal{T}_{\pi^*}\tilde{V}-\tilde{V}=c_{\pi^*}-(I-\gamma P_{\pi^*})\tilde{V}$, so that
$\Lambda r'=V^*-\tilde{V}$ because $V^*=\Lambda c_{\pi^*}$. Since $\Lambda\ge0$ component-wise and
$|r'|=r'+2(r')^-$ (the last paragraph in Section \ref{sec: MDP formulation}), we obtain
\begin{equation}
\|V^* - \tilde{V}\|_{1,\nu}=\nu^\top|\Lambda r'|\ \le\ \nu^\top \Lambda|r'|
\ =\ \nu^\top (V^* - \tilde{V})+2\,\nu^\top \Lambda (r')^-,
\label{eq: equation 13 in RLP 2004}
\end{equation}
which is Equation (13) in \citet{de_farias_constraint_2004}. Moreover, $(r')^-$ vanishes off
$\mathcal{B}^{*}(\tilde{V})$ and $(1-\gamma)\nu^\top \Lambda=(d^{\pi^*}_\nu)^\top$, hence
\begin{equation}
\nu^\top \Lambda (r')^- \le \|(r')^-\|_\infty\!\!\sum_{s\in\mathcal{B}^{*}(\tilde{V})}\!\!\big[\nu^\top \Lambda\big]_s
  = \frac{\|(r')^-\|_\infty}{1-\gamma}\,d^{\pi^*}_\nu\big(\mathcal{B}^{*}(\tilde{V})\big) 
 \le \frac{\beta\,\|(r')^-\|_\infty}{1-\gamma}\,\psi\big(\mathcal{B}(\tilde{V})\big)
 \le \frac{\beta\,\xi}{1-\gamma}\|(r')^-\|_\infty
\label{eq: intermediate bound}
\end{equation}
with probability at least $1-\delta$, where $\mathcal{B}$ and $\mathcal{B}^{*}$ are as in Assumption
\ref{ass: psi vs d^*}, the third inequality is that assumption applied to $\Phi$, and the fourth is
Corollary \ref{cor: violations under Z} applied to \RLP{\Phi}{\mathcal{Z}}. Combining
(\ref{eq: equation 13 in RLP 2004}) and (\ref{eq: intermediate bound}) concludes the proof. $\square$

\textit{Proof of Theorem \ref{thm: ALP vs RLP}.} The constraints indexed by $\mathcal{Z}$ form a subset of those of \ALP{\Phi}, so \RLP{\Phi}{\mathcal{Z}} is a relaxation and
$\nu^\top \tilde{V} \geq \nu^\top \tilde{V}'$. This leads to
\[
\nu^\top V^*-\nu^\top\tilde{V} \leq \nu^\top V^*-\nu^\top \tilde{V}' = \| V^* - \tilde{V}'\|_{1,\nu}
\]
by Lemma \ref{lem: preliminaries} case (iii). Therefore, by Lemma \ref{lem: psi to xi} applied to
\RLP{\Phi}{\mathcal{Z}}, we can conclude the proof. $\square$

\begin{lemma}[Action-restriction bound]\label{lem: e step}
For the basis-expansion step of Definition \ref{def: expansion step} under
Assumption \ref{ass: Y suboptimality gap},
\begin{equation}
\big(\mathcal{L}_{\mathcal{Y}}\tilde{V}_k\big)(s)\ \le\
\big(\mathcal{T}_{\hat{\pi}_{k+1}}\tilde{V}_k\big)(s)+e
\qquad\text{for every }s\in\mathcal{Y}_{\mathcal{S}}.
\label{eq: e step}
\end{equation}
\end{lemma}

\textit{Proof.} Fix $s\in\mathcal{Y}_{\mathcal{S}}\subseteq\mathcal{Z}^{k+1}_{\mathcal{S}}$.
By (\ref{eq: hybrid greedy policy}) and (\ref{eq: hat notation}), $\hat{\pi}_{k+1}(s)\in\mathcal{Z}^{k+1}_{\mathcal{A}}(s)$, so the minimum defining $(\mathcal{L}_{\mathcal{Z}^{k+1}}\tilde{V}_k)(s)$ in
(\ref{eq: hybrid Bellman operator}) runs over a set containing $\hat{\pi}_{k+1}(s)$, giving
$(\mathcal{L}_{\mathcal{Z}^{k+1}}\tilde{V}_k)(s)\le(\mathcal{T}_{\hat{\pi}_{k+1}}\tilde{V}_k)(s)$.
Adding $(\mathcal{L}_{\mathcal{Y}}\tilde{V}_k)(s)-(\mathcal{L}_{\mathcal{Z}^{k+1}}\tilde{V}_k)(s)
\le\mathscr{E}(\tilde{V}_k,\mathcal{Z}^{k+1})\le e$, which is Definition \ref{def: Y gap} with
Assumption \ref{ass: Y suboptimality gap}, yields (\ref{eq: e step}). $\square$

\textit{Proof of Theorem \ref{thm: major improvement (with sampling)}.}
Throughout this proof, we write $\tilde{\pi}:=\hat{\pi}_{k+1}$ and $\tilde{u}:=\hat{u}_{k+1}$
for the abbreviations introduced in (\ref{eq: hat notation}), set
\[
\mathcal{R}^{k+1}:=\mathcal{Z}^{k+1}_{\mathcal{S}}\setminus\mathcal{Y}_{\mathcal{S}}
=\mathcal{X}^{k+1}_{\mathcal{S}}\setminus\mathcal{Y}_{\mathcal{S}},
\]
and use repeatedly that $\mathcal{Y}\subseteq\mathcal{Z}^{k+1}$ implies $\mathcal{Z}^{k+1}_{\mathcal{S}}\cap\mathcal{Y}_{\mathcal{S}}=\mathcal{Y}_{\mathcal{S}}$, so that $\{\mathcal{Y}_{\mathcal{S}},\mathcal{R}^{k+1}\}$ partitions $\mathcal{Z}^{k+1}_{\mathcal{S}}$.

\emph{Step 0 (the dual identities, one per basis function).} By hypothesis $\tilde{w}_{k+1}$ is the unique optimal solution of \RLP{\Phi_{k+1}}{\mathcal{Z}^{k+1}} and $\tilde{\pi}(s)$ is unique at every $s\in\mathcal{Z}^{k+1}_{\mathcal{S}}$ with a binding primal constraint (Definition \ref{def: binding (sampling)}), so Property \ref{prop: rlp duality} applies to \RDUAL{\Phi_{k+1}}{\mathcal{Z}^{k+1}} and yields
\begin{align}
\tilde{u}^\top c_{\tilde{\pi}} &= \nu^\top\tilde{V}_{k+1}, \label{eq: strong duality RLP}\\
\tilde{u}^\top\left(I-\gamma P_{\tilde{\pi}}\right)\Phi_{k+1} &= \nu^\top\Phi_{k+1}.
\label{eq: dual feasibility RLP}
\end{align}

Both sides of (\ref{eq: dual feasibility RLP}) are row vectors of length $\ell_{k+1}$, obtained by pairing the linear functionals $\tilde{u}^\top(I-\gamma P_{\tilde{\pi}})$ and $\nu^\top$ with the BFs of $\Phi_{k+1}$ one at a time. The identity therefore holds BF by BF. For every $\phi$ in $\Phi_{k+1}$, we have
\begin{equation}
\tilde{u}^\top\left(I-\gamma P_{\tilde{\pi}}\right)\phi\ =\ \nu^\top\phi.
\label{eq: per BF identity}
\end{equation}
Let $\phi=\mathcal{L}_{\mathcal{Y}}\tilde{V}_k$, which is by construction the last BF in $\Phi_{k+1}$:
\begin{equation}
\tilde{u}^\top\mathcal{L}_{\mathcal{Y}}\tilde{V}_k
=\nu^\top\mathcal{L}_{\mathcal{Y}}\tilde{V}_k
+\gamma\,\tilde{u}^\top P_{\tilde{\pi}}\mathcal{L}_{\mathcal{Y}}\tilde{V}_k .
\label{eq: feasibility 2 RLP (revised)}
\end{equation}

\emph{Step 1 (decomposition).} By the greedy dual sparsity of Property \ref{prop: rlp duality}, $\tilde{u}(s)=0$ for every
$s\in\mathcal{S}\setminus\mathcal{Z}^{k+1}_{\mathcal{S}}$, and
$(\mathcal{L}_{\mathcal{Y}}\tilde{V}_k)(s)=\tilde{V}_k(s)$ for $s\notin\mathcal{Y}_{\mathcal{S}}$ by
(\ref{eq: hybrid Bellman operator}). Hence
\begin{equation}
\tilde{u}^\top\mathcal{L}_{\mathcal{Y}}\tilde{V}_k
=\sum_{s\in\mathcal{Y}_{\mathcal{S}}}\tilde{u}(s)\big(\mathcal{L}_{\mathcal{Y}}\tilde{V}_k\big)(s)
+\sum_{s\in\mathcal{R}^{k+1}}\tilde{u}(s)\,\tilde{V}_k(s).
\label{eq: step1}
\end{equation}

\emph{Step 2 (paying the action-restriction gap).} Applying Lemma \ref{lem: e step} to each term of
the first sum in (\ref{eq: step1}) and using $\tilde{u}\ge0$,
\begin{align}
\tilde{u}^\top\mathcal{L}_{\mathcal{Y}}\tilde{V}_k
&\le \sum_{s\in\mathcal{Y}_{\mathcal{S}}}\tilde{u}(s)
\Big(c_{\tilde{\pi}}(s)+\gamma\big(P_{\tilde{\pi}}\tilde{V}_k\big)(s)+e\Big)
+\sum_{s\in\mathcal{R}^{k+1}}\tilde{u}(s)\,\tilde{V}_k(s) \notag\\
&= \sum_{s\in\mathcal{Z}^{k+1}_{\mathcal{S}}}\tilde{u}(s)c_{\tilde{\pi}}(s)
+\gamma\!\!\sum_{s\in\mathcal{Y}_{\mathcal{S}}}\!\!\tilde{u}(s)\big(P_{\tilde{\pi}}\tilde{V}_k\big)(s)
+e\!\!\sum_{s\in\mathcal{Y}_{\mathcal{S}}}\!\!\tilde{u}(s)
+\!\!\sum_{s\in\mathcal{R}^{k+1}}\!\!\tilde{u}(s)\big(\tilde{V}_k(s)-c_{\tilde{\pi}}(s)\big) \notag\\
&= \nu^\top\tilde{V}_{k+1}
+\gamma\!\!\sum_{s\in\mathcal{Y}_{\mathcal{S}}}\!\!\tilde{u}(s)\big(P_{\tilde{\pi}}\tilde{V}_k\big)(s)
+e\!\!\sum_{s\in\mathcal{Y}_{\mathcal{S}}}\!\!\tilde{u}(s)
+\!\!\sum_{s\in\mathcal{R}^{k+1}}\!\!\tilde{u}(s)\big(\tilde{V}_k(s)-c_{\tilde{\pi}}(s)\big),
\label{eq: eq02 RLP}
\end{align}
where the second line adds and subtracts
$\sum_{s\in\mathcal{R}^{k+1}}\tilde{u}(s)c_{\tilde{\pi}}(s)$ so as to complete the cost sum over all
of $\mathcal{Z}^{k+1}_{\mathcal{S}}$, and the third applies strong duality
(\ref{eq: strong duality RLP}).

\emph{Step 3 (matching the two expressions).} Splitting
$\gamma\tilde{u}^\top P_{\tilde{\pi}}\mathcal{L}_{\mathcal{Y}}\tilde{V}_k$ over the partition
$\{\mathcal{Y}_{\mathcal{S}},\mathcal{R}^{k+1}\}$ and substituting
(\ref{eq: feasibility 2 RLP (revised)}) into the left-hand side of (\ref{eq: eq02 RLP}), then
rearranging to isolate $\nu^\top\tilde{V}_{k+1}$ and writing the step cost and transition terms on
$\mathcal{R}^{k+1}$ by means of the policy Bellman operator $\mathcal{T}_{\tilde{\pi}}$, we obtain
\begin{align}
\nu^\top\tilde{V}_{k+1}
&\ \geq\ \nu^\top\mathcal{L}_{\mathcal{Y}}\tilde{V}_k
+\gamma\!\!\sum_{s\in\mathcal{Y}_{\mathcal{S}}}\!\!\tilde{u}(s)
\Big(P_{\tilde{\pi}}\big(\mathcal{L}_{\mathcal{Y}}\tilde{V}_k-\tilde{V}_k\big)\Big)(s)
+\!\!\sum_{s\in\mathcal{R}^{k+1}}\!\!\tilde{u}(s)
\Big(\big(\mathcal{T}_{\tilde{\pi}}\mathcal{L}_{\mathcal{Y}}\tilde{V}_k\big)(s)-\tilde{V}_k(s)\Big)
-e\!\!\sum_{s\in\mathcal{Y}_{\mathcal{S}}}\!\!\tilde{u}(s) \notag\\
&\ =\ \nu^\top\mathcal{L}_{\mathcal{Y}}\tilde{V}_k
+g\big(\hat{u}_{k+1},\hat{\pi}_{k+1},\tilde{V}_k\big)
-e\!\!\sum_{s\in\mathcal{Y}_{\mathcal{S}}}\!\!\tilde{u}(s),
\label{eq: eq03 RLP}
\end{align}
with $g$ as in (\ref{eq: g}).

\emph{Step 4 (from objective values to approximation errors).} Subtracting $\nu^\top\tilde{V}_k$
from both sides of (\ref{eq: eq03 RLP}) and using
$\mathcal{L}_{\mathcal{Y}}\tilde{V}_k\ge\tilde{V}_k$ (Lemma \ref{lem: L_Y monotone} applied to
$\tilde{w}_k$, which is feasible for \RLP{\Phi_k}{\mathcal{Z}^{k}} and hence satisfies in
particular the constraints indexed by $\mathcal{Y}\subseteq\mathcal{Z}^{k}$), which turns
$\nu^\top(\mathcal{L}_{\mathcal{Y}}\tilde{V}_k-\tilde{V}_k)$ into
$\|\mathcal{L}_{\mathcal{Y}}\tilde{V}_k-\tilde{V}_k\|_{1,\nu}$, gives
\begin{equation}
\nu^\top(V^*-\tilde{V}_{k})-\nu^\top(V^*-\tilde{V}_{k+1})
\ \geq\ \big\|\mathcal{L}_{\mathcal{Y}}\tilde{V}_k-\tilde{V}_{k}\big\|_{1,\nu}
+g\big(\hat{u}_{k+1},\hat{\pi}_{k+1},\tilde{V}_k\big)
-e\!\!\sum_{s\in\mathcal{Y}_{\mathcal{S}}}\!\!\tilde{u}(s).
\label{eq: eq04 RLP}
\end{equation}
Finally \[\nu^\top(V^*-\tilde{V}_k)\le\|V^*-\tilde{V}_k\|_{1,\nu},\] while Lemma
\ref{lem: psi to xi}, applied to \RLP{\Phi_{k+1}}{\mathcal{Z}^{k+1}} and with $\Phi_{k+1}$ formed before $\mathcal{X}^{k+1}$ was drawn (Definition
\ref{def: expansion step}), gives
\[
\nu^\top(V^*-\tilde{V}_{k+1})\ \geq\ \|V^*-\tilde{V}_{k+1}\|_{1,\nu}
-\frac{2\beta\,\xi}{1-\gamma}\big\|(\mathcal{T}_{\pi^*}\tilde{V}_{k+1}-\tilde{V}_{k+1})^{-}\big\|_\infty
\]
with probability at least $1-\delta$. Combining the last two inequalities with (\ref{eq: eq04 RLP}) yields (\ref{eq: major improvement sampling}). $\square$

\textit{Proof of Lemma \ref{lem: g nonnegative}.} Write $\tilde{\pi}:=\hat{\pi}_{k+1}$ and
$\tilde{u}:=\hat{u}_{k+1}$. For the first sum of $g$ in (\ref{eq: g}), each term is
$\gamma\,\tilde{u}(s)\sum_{s'\in\mathcal{S}}p(s'\mid s,\tilde{\pi}(s))
\big(\mathcal{L}_{\mathcal{Y}}\tilde{V}_k(s')-\tilde{V}_k(s')\big)$, in which every factor is
non-negative: $\tilde{u}\ge0$ because \textup{R-Dual} inherits the sign constraints of
\DUAL{\Phi_{k+1}} (Property \ref{prop: rlp duality}), $\gamma\in(0,1)$ and $p(\cdot\mid s,a)\ge0$,
and $\mathcal{L}_{\mathcal{Y}}\tilde{V}_k\ge\tilde{V}_k$ component-wise by Lemma
\ref{lem: L_Y monotone} applied to $\tilde{w}_k$, which satisfies the constraints indexed by
$\mathcal{Y}$. The second sum, over
$s\in\mathcal{Z}^{k+1}_{\mathcal{S}}\setminus\mathcal{Y}_{\mathcal{S}}$, is exactly the
left-hand side of (\ref{eq: propagated dominance}) and is non-negative by Assumption
\ref{ass: propagated dominance}. Summing the two yields $g\ge0$. $\square$

\textit{Proof of Theorem \ref{thm: convergence of HBOBF}}. For any iteration counter $k\ge0$, write $r_k:=\mathcal{L}_{\mathcal{Y}}\tilde{V}_k-\tilde{V}_k$, which is non-negative by Lemma \ref{lem: L_Y monotone} and vanishing off $\mathcal{Y}_{\mathcal{S}}$ by (\ref{eq: hybrid Bellman operator}). Under the conditions (ii) and (iii) the deterministic part of the proof of Theorem \ref{thm: major improvement (with sampling)}, namely (\ref{eq: eq04 RLP}), applies at iteration $k+1$, and with $e=0$ and $g\ge0$ (Lemma \ref{lem: g nonnegative}), it gives
\begin{equation}
\nu^\top\tilde{V}_{k+1}-\nu^\top\tilde{V}_{k}\ \geq\ \|r_k\|_{1,\nu}.
\label{eq: e1}
\end{equation}
Note that (\ref{eq: eq04 RLP}) does not invoke Proposition \ref{prop: probability of violations} or Corollary \ref{cor: violations under Z}. In particular it does not require $\tilde{w}_{k+1}$ to be unique, and the quantities $\nu^\top\tilde{V}_{k}$ appearing in (\ref{eq: e1}) are optimal values of their respective programs.

Suppose the algorithm does not terminate, i.e.\ $\|r_k\|_\infty>\epsilon$ for every $k\ge0$. Since $r_k\ge0$ is supported on $\mathcal{Y}_{\mathcal{S}}$ and $\nu$ has full support there,
\[
\|r_k\|_{1,\nu}=\sum_{s\in\mathcal{Y}_{\mathcal{S}}}\nu(s)\,r_k(s)
\ \geq\ \nu_{\min}\max_{s\in\mathcal{Y}_{\mathcal{S}}}r_k(s)=\nu_{\min}\|r_k\|_\infty>\nu_{\min}\epsilon,
\]
so each increment in (\ref{eq: e1}) exceeds $\nu_{\min}\epsilon>0$. Telescoping (\ref{eq: e1}) over
$k=0,\dots,K$ and invoking Assumption \ref{ass: rlp-bounded},
\[
\bar{U}-\nu^\top\tilde{V}_0\ \geq\ \nu^\top\tilde{V}_{K+1}-\nu^\top\tilde{V}_0
\ \geq\ (K+1)\,\nu_{\min}\epsilon
\]
for every $K$, which fails once $K+1>(\bar{U}-\nu^\top\tilde{V}_0)/(\nu_{\min}\epsilon)$. Hence the stopping
condition is met at some $K\le K_{\max}$. $\square$

\textit{Proof of Theorem \ref{thm: bellman residual at termination}.} Throughout this proof write
$\epsilon':=\|\mathcal{L}_{\mathcal{Y}}V-V\|_\infty$. By the definition of the weighted $L_1$ norm,
\begin{align}
\|\mathcal{T}V-V\|_{1,\psi_{\tilde{\pi}}}
&=
\sum_{s\in\mathcal{S}}\psi_{\tilde{\pi}}(s)(\mathcal{T}V(s)-V(s))^+
+
\sum_{s\in\mathcal{S}}\psi_{\tilde{\pi}}(s)(\mathcal{T}V(s)-V(s))^- .
\label{eq: decomposition}
\end{align}
For $s\in\mathcal{Y}_{\mathcal{S}}$, since $\mathcal{Y}_{\mathcal{A}}(s)\subseteq\mathcal{A}$,
\begin{align}
(\mathcal{T}V)(s)
& =
\min_{a\in\mathcal{A}}\left\{c(s,a) + \gamma \sum_{s' \in \mathcal{S}} p(s'|s,a) V(s')\right\} \notag \\
& \leq
\min_{a\in\mathcal{Y}_{\mathcal{A}}(s)}\left\{c(s,a) + \gamma \sum_{s' \in \mathcal{S}} p(s'|s,a) V(s')\right\}
=
(\mathcal{L}_{\mathcal{Y}}V)(s)
\ \leq\ V(s)+\epsilon',
\end{align}
where the last inequality holds because $|(\mathcal{L}_{\mathcal{Y}}V)(s)-V(s)|\le\epsilon'$ by the
definition of $\epsilon'$. Hence,
\begin{equation}
(\mathcal{T}V(s)-V(s))^+\leq\epsilon',
\qquad s\in\mathcal{Y}_{\mathcal{S}} .
\label{eq: positive Ys}
\end{equation}
Splitting the positive part of (\ref{eq: decomposition}) according to $\mathcal{Y}_{\mathcal{S}}$,
\begin{align}
\sum_{s\in\mathcal{S}}\psi_{\tilde{\pi}}(s)(\mathcal{T}V(s)-V(s))^+
&=
\sum_{s\in\mathcal{Y}_{\mathcal{S}}}\psi_{\tilde{\pi}}(s)(\mathcal{T}V(s)-V(s))^+ +
\sum_{s\in\mathcal{S}\setminus\mathcal{Y}_{\mathcal{S}}}\psi_{\tilde{\pi}}(s)(\mathcal{T}V(s)-V(s))^+
\nonumber\\
&\leq
\epsilon'\,\psi_{\tilde{\pi}}(\mathcal{Y}_{\mathcal{S}})
+
(1-\psi_{\tilde{\pi}}(\mathcal{Y}_{\mathcal{S}}))
\|(\mathcal{T}V-V)^+\|_\infty ,
\label{eq: positive bound}
\end{align}
where the second line is implied by (\ref{eq: positive Ys}). For the negative part,
\begin{align}
\sum_{s\in\mathcal{S}}\psi_{\tilde{\pi}}(s)&(\mathcal{T}V(s)-V(s))^-
\leq
\|(\mathcal{T}V-V)^-\|_\infty
\sum_{\{s\in\mathcal{S}:(\mathcal{T}V)(s)<V(s)\}}\psi_{\tilde{\pi}}(s)
\notag\\
&=
Z\|(\mathcal{T}V-V)^-\|_\infty\,
\psi\Big(\big\{(s,\tilde{\pi}(s)):\ s \in\mathcal{S},\ (\mathcal{T}V)(s)<V(s)\big\}\Big)
\notag\\
&\leq
Z\|(\mathcal{T}V-V)^-\|_\infty\,
\psi\left(\left\{(s,a)\in(\mathcal{S}\times\mathcal{A}):V(s)>c(s,a)+\gamma\sum_{s'}p(s'|s,a)V(s')\right\}\right)
\notag\\
&\leq
Z\xi\|(\mathcal{T}V-V)^-\|_\infty ,
\label{eq: negative bound}
\end{align}
with a probability of at least $1-\delta$, where the second line uses the definition
(\ref{eq: psi tilde pi}) of $\psi_{\tilde{\pi}}$ together with the fact that distinct states index
distinct pairs, the third holds because $(\mathcal{T}V)(s)<V(s)$ implies
$V(s)>(\mathcal{T}_{\tilde{\pi}(s)}V)(s)$ so that the pair $(s,\tilde{\pi}(s))$ belongs to the
violation set, and the fourth is (\ref{eq: psi less than xi under Z}) of Corollary
\ref{cor: violations under Z}, applied to \RLP{\Phi_k}{\mathcal{Z}^{k}} with unique optimal solution
$\tilde{w}_k$ and $V=\Phi_k\tilde{w}_k$. Combining (\ref{eq: decomposition}),
(\ref{eq: positive bound}), and (\ref{eq: negative bound}) gives (\ref{eq: bellman residual bound}).
$\square$

\end{bibunit}

\newpage


\begin{bibunit}[informs2014] 
\ECSwitch          
\ECHead{E-Companion}

This e-companion accompanies the main paper. It collects the conditions under which the results of the paper extend to continuous state and action spaces, the proofs omitted from the main body, and supplementary discussion of the sampling design.

\section{Continuous State and Action Spaces}\label{app: continuous}

Throughout the paper, all results are established for finite $\mathcal{S}$ and finite $\mathcal{A}$. This section states what is required for each of them when $\mathcal{S}$ and $\mathcal{A}$ are continuous and compact, which of them extend unchanged, and where an extension is unavailable. One obstruction is that the concentrability constant of Assumption \ref{ass: psi vs d^*} is finite under mild support conditions when $\mathcal{S}$ is finite, whereas in the continuous case it additionally requires the discounted occupancy measure of the optimal policy to be absolutely continuous with respect to the sampling distribution.

\subsection{Standing Assumptions}\label{app: continuous assumptions}

The following conditions replace finiteness of $\mathcal{S}$ and $\mathcal{A}$ and are
invoked individually below.
\begin{enumerate}[label=(C\arabic*)]
\item \textbf{Model regularity.} $\mathcal{S}$ and $\mathcal{A}$ are compact, the cost
function $c$ is continuous, and the transition kernel $p(\cdot\mid s,a)$ is weakly
continuous. These conditions guarantee that the minimum in
(\ref{eq: Bellman Operator Definition}) is attained and that $\mathcal{T}$ maps bounded
continuous functions to bounded continuous functions.
\item \textbf{Constraint qualification.} \ALP{\Phi} admits at least one strictly feasible
solution $\bar{w}$, i.e.\ one at which all constraints hold strictly. The feasible region
then has non-empty interior and Slater's condition delivers the strong duality used in
Property \ref{prop: dual properties}.

\item \textbf{Absolute continuity.} The condition of Section \ref{app: concentrability},
invoked wherever Assumption \ref{ass: psi vs d^*} is used.
\end{enumerate}

If $\mathcal{S}$ and $\mathcal{A}$ are continuous, we assume that each $\phi_i$ is continuous
for all $i=1, \cdots, \ell$ in Sections \ref{sec: MDP formulation} and \ref{sec: BOBF} of the
main paper.

\subsection{Which Results Extend}\label{app: continuous results}

Under (C1) and (C2), Lemma \ref{lem: preliminaries}, Property \ref{prop: dual},
Property \ref{prop: dual properties}, Lemma \ref{lem: sparsity of u},
Lemma \ref{lem: simplified dual properties}, Lemma \ref{lem: mu normalized}, and Theorem
\ref{thm: major improvement} extend verbatim, with sums over $\mathcal{S}$ replaced by
integrals against the transition kernel and $\Phi w$ read as
$\sum_{i}w_i\phi_i$ in the space of bounded measurable functions. Part (a) of Theorem \ref{thm: convergence} likewise extends.

\subsection{Concentrability in Continuous Spaces}\label{app: concentrability}

Assumption \ref{ass: psi vs d^*} establishes a concentrability condition that bridges the sampling distribution $\psi$ and the discounted occupancy distribution $d_\nu^{\pi^*}$ induced by the optimal policy. This assumption bounds the probability of encountering a Bellman violation along the optimal trajectory by the probability of sampling a violation under $\psi$, scaled by $\beta$. The theoretical conditions necessary to guarantee the existence of the structural constant $\beta$ depend heavily on the topology of the state-action space.

When the state space $\mathcal{S}$ is finite, establishing a finite density ratio is mathematically straightforward (Lemma \ref{lem: beta sufficient}). The fundamental condition requires that the sampling distribution $\psi$ maintains adequate support over $\mathcal{S} \times \mathcal{A}$. A sufficient global condition to guarantee a finite $\beta$ is for $\psi$ to possess full support. For example, via a uniform sampling distribution where $\psi(s,a) > 0$ everywhere. This trivially bounds the worst-case global density ratio.

Extending concentrability conditions to continuous settings necessitates stricter measure-theoretic foundations \citep{munos_performance_2007}. To prevent the density ratio from diverging, the optimal occupancy distribution $d_\nu^{\pi^*}$ must be absolutely continuous with respect to the sampling distribution $\psi$, meaning $d_\nu^{\pi^*}$ assigns a probability of zero wherever $\psi$ does. Furthermore, to prevent $d_\nu^{\pi^*}$ from collapsing into infinite-density singularities, the transition kernel is traditionally required to be absolutely continuous with respect to the Lebesgue measure, meaning it possesses a valid probability density function, to ensure the occupancy distribution remains smooth and bounded. The role of transition kernel regularity in ensuring the stability of concentrability coefficients is well-established in \citet{farahmand2010error} and \citet{rust_using_1997}.

Assumption \ref{ass: psi vs d^*} is nonetheless weaker than a global density-ratio bound, because it is formulated locally. It does not compare $d_\nu^{\pi^*}$ with $\psi$ over the whole state--action space, but only over the violation sets generated by the solutions in $\mathcal{W}(\Phi)$. When $\mathcal{S}$ is finite, this removes the need for global full support. It suffices that $\psi$ place positive mass on the pairs $(s,\pi^*(s))$ that can appear in $\mathcal{B}^{*}(V_{\tilde{w}})$ for some $\tilde{w}\in\mathcal{W}(\Phi)$, and Lemma \ref{lem: beta sufficient} is the cruder sufficient condition obtained by ignoring this localization.

\subsection{The Undiscounted Limit}\label{app: gamma to one}

\begin{remark}[On the discount factor and the validity of ALP]
\label{rem: gamma-to-one}
The ALP formulation in this study is stated for the discounted setting and relies on $\gamma<1$ in an essential way, not merely for tightness of the bounds. For fixed action $a$, invertibility of $(I-\gamma P_a)$ follows from the Neumann series
\begin{align*}
    (I-\gamma P_a)^{-1} = \sum_{t=0}^{\infty} \gamma^t P_a^t,
\end{align*}
which converges only because the spectral radius of $\gamma P_a$ is strictly less than 1. Since $P_a$ is row-stochastic, $P_a \mathbf{1} = \mathbf{1}$, so at $\gamma = 1$ the matrix $(I-P_a)$ is singular with $\mathbf{1}$ in its null space. Three consequences follow. First, the Bellman operator $\mathcal{T}$ ceases to be a contraction, so the fixed point $V^*$ is no longer unique (only the differential value function, defined up to an additive constant, remains well posed). Second, because $\sum_{s'} p(s'\mid s,a) = 1$, adding any constant to $\Phi w$ preserves feasibility of every constraint in $\mathrm{ALP}(\Phi)$ at $\gamma=1$, if a constant basis function lies in $\Phi$. This is a recession direction of the feasible polyhedron, and since $\nu^\top \mathbf{1} = 1$, the objective $\nu^\top \Phi w$ becomes unbounded above, i.e., $\mathrm{ALP}(\Phi)$ admits no optimal solution. Third, the approximation guarantee of \citet{de_farias_linear_2003} carries an explicit $\frac{1}{1-\gamma}$ factor that diverges as $\gamma \to 1$. Hence $\mathrm{ALP}(\Phi)$ is not simply looser as $\gamma \to 1$. It is structurally undefined, and the proofs in this paper are not intended to extend to that limit. For the average-cost regime the appropriate object is a dedicated average-cost ALP built on the differential value function (\citealp{de_farias_average_2006}), not a limit of the discounted formulation.
\end{remark}

\section{Omitted Proofs for the Exact-ALP Setting}\label{app: proofs exact}

This section collects the proofs omitted from Section \ref{sec: BOBF} of the main paper.

\subsection{The Role of Strict Discounting}
\label{oa: gamma}

The main text fixes $\gamma\in(0,1)$ once and does not restate the consequences. They are
the following. First, $\mathcal{T}$ is a $\gamma$-contraction in $\|\cdot\|_\infty$ with
unique fixed point $V^*$, which is what makes the residual $\|\mathcal{T}\bar{V}-\bar{V}\|_\infty$
an optimality certificate (Lemma \ref{lem: epsilon optimal}). Second,
$(I-\gamma P_a)^{-1}=\sum_{t\ge0}\gamma^{t}P_a^{t}$ exists and is non-negative, so that
$\mathcal{V}^\pi$ is finite and the discounted occupancy measures of
\online{oa: occupancy} are well defined. Third, \ALP{\Phi} attains its optimum, since every
feasible $\Phi w$ satisfies $\nu^\top\Phi w\le\nu^\top V^*<\infty$ by Lemma
\ref{lem: preliminaries}(i).

Several bounds in the paper carry a factor $1/(1-\gamma)$: The amplification term of Theorem
\ref{thm: major improvement}, the certificate of Lemma \ref{lem: epsilon optimal}, the
sampling penalty of Theorem \ref{thm: ALP vs RLP}, and the normalization
$\tilde{u}_{\tilde{\pi}}^\top\mathbf{1}=1/(1-\gamma)$ of Property \ref{prop: dual structure}(4).
These do not degrade gracefully as $\gamma\uparrow1$ but become vacuous, and at $\gamma=1$ the
formulation itself breaks down (see \online{app: gamma to one}).

\subsection{Proof of Lemma \ref{lem: preliminaries}}
\label{oa: proof preliminaries}

\textit{Proof of Lemma \ref{lem: preliminaries}.} The constraints in \ALP{\Phi_k} imply
$\bar{V}_k\leq\mathcal{T}\bar{V}_k$. By the monotonicity of the Bellman operator, we obtain
(\ref{eq:fixedpointineq}), i.e.
$\bar{V}_k\leq \mathcal{T}\bar{V}_k\leq \mathcal{T}\mathcal{T}\bar{V}_k\leq\cdots\leq V^*$,
which establishes case (i). Note that $V^*$ is the unique fixed point of $\mathcal{T}$. For case (ii), let $\bar{w}_{k+1}=(0,0,\cdots,0,1)^\top\in\mathbb{R}^{\ell_{k+1}}$. Its first
$\ell_{k+1}-1$ components are zero, so the first $\ell_{k+1}-1$ BFs of $\Phi_{k+1}$ enter the linear
combination with zero weight and
$\bar{V}_{k+1}=\Phi_{k+1}\bar{w}_{k+1}=\mathcal{T}\bar{V}_{k}$. From
(\ref{eq:fixedpointineq}) we know $\mathcal{T}\bar{V}_{k}\leq\mathcal{T}\mathcal{T}\bar{V}_{k}$.
Therefore, we have $\bar{V}_{k+1}=\mathcal{T}\bar{V}_{k}\leq\mathcal{T}\mathcal{T}\bar{V}_{k}
=\mathcal{T}\bar{V}_{k+1}$, which completes case (ii). For case (iii), by the definition of
the weighted $L_1$ norm,
\[
\|V^* - \bar{V}_k\|_{1,\nu}=\sum_{s \in \mathcal{S}} \nu(s)\big|V^*(s) - \bar{V}_k(s)\big|.
\]
From case (i), $V^*(s)\geq\bar{V}_k(s)$ for all $s\in\mathcal{S}$, and therefore
\begin{align*}
    \|V^* - \bar{V}_k\|_{1,\nu}
    & = \sum_{s \in \mathcal{S}} \nu(s) \left(V^*(s) - \bar{V}_k(s)\right)
      = \nu^\top V^* - \nu^\top \bar{V}_k,
\end{align*}
which proves case (iii). With a similar argument and inequality
(\ref{eq:fixedpointineq}), case (iv) follows. $\square$

\subsection{Dual Structure of ALP($\Phi$): Statements and Proofs}
\label{oa: dual exact}

Property \ref{prop: dual structure} of the main text collects three statements, which we
restate separately here in the form in which they are proved.

\begin{lemma}[Greedy dual sparsity] \label{lem: sparsity of u}
    Let $\tilde{V} = \Phi \tilde{w}$, where $\tilde{w}$ denotes an optimal solution to \ALP{\Phi}, and let $\tilde{u}$ represent an optimal solution to \DUAL{\Phi}. Define the greedy policy $\tilde{\pi}:= \argmin_{\pi}\{c_\pi+\gamma P_{\pi} \tilde{V}\} $. If the greedy action $\tilde{\pi}(s)$ is unique for any state $s$ with a binding primal constraint (Definition \ref{def: binding}), then for all $s \in \mathcal{S}$,
    \begin{align*}
        \tilde{u}_a(s) = 0 \qquad \forall a\in\mathcal{A}\setminus\{\tilde{\pi}(s)\}.
    \end{align*}
\end{lemma}

\textit{Proof of Lemma \ref{lem: sparsity of u}.}  The result follows directly from the complementary slackness condition in Property \ref{prop: dual properties}. We consider two cases for the state $s$. First, if the primal constraints at $s$ are non-binding for all actions, then $\tilde{u}_a(s) = 0$ for all $a \in \mathcal{A}$, and the result holds trivially. Second, consider any state $s$ where the primal constraint is binding, which means $c(s,\tilde{\pi}(s)) + \gamma \sum_{s'\in \mathcal{S}} p(s'|s,\tilde{\pi}(s)) \tilde{V}(s') = \tilde{V}(s)$. Since $\tilde{\pi}(s)$ is the unique greedy action with respect to $\tilde{V}$, for any alternative action $a \in \mathcal{A} \setminus \{ \tilde{\pi}(s) \}$, we have
    \begin{align*}
        c(s,a) + \gamma \sum_{s'\in\mathcal{S}} p(s'|s,a) \tilde{V}(s')
        > c(s,\tilde{\pi}(s)) + \gamma \sum_{s'\in \mathcal{S}} p(s'|s,\tilde{\pi}(s)) \tilde{V}(s') = \tilde{V}(s).
    \end{align*}
    Consequently, the primal constraint for pair $(s,a)$ is strictly satisfied. By complementary slackness, we have
    \begin{align*}
        \tilde{u}_a(s) = 0, \quad \forall a \in \mathcal{A} \setminus \{ \tilde{\pi}(s) \}. \quad \square
    \end{align*}

\begin{lemma} [Policy-reduced dual properties] \label{lem: simplified dual properties}
    Under the hypotheses of Lemma \ref{lem: sparsity of u}, the following hold.
    \begin{enumerate}
        \item \textbf{Strong duality}: $z(\tilde{u}) = \sum_{a \in \mathcal{A}} \tilde{u}_a^\top c_a = \tilde{u}_{\tilde{\pi}}^\top c_{\tilde{\pi}} = \nu^\top \tilde{V}$.
        \item \textbf{Dual feasibility}: $\sum_{a\in \mathcal{A}} \tilde u_a^\top \left(I-\gamma P_a \right ) \Phi = \tilde u_{\tilde{\pi}}^\top \left(I-\gamma P_{\tilde{\pi}} \right ) \Phi = \nu^\top \Phi$.
    \end{enumerate}
\end{lemma}

\textit{Proof of Lemma \ref{lem: simplified dual properties}.} Under the hypothesis of the lemma, Lemma \ref{lem: sparsity of u} gives $\tilde{u}_a(s)=0$ for every $s\in\mathcal{S}$ and every $a\in\mathcal{A}\setminus\{\tilde{\pi}(s)\}$, so that only the greedy action contributes at each state:
\begin{equation}
\sum_{a\in\mathcal{A}}\tilde{u}_a^\top c_a
=\sum_{s\in\mathcal{S}}\sum_{a\in\mathcal{A}}\tilde{u}_a(s)\,c_a(s)
=\sum_{s\in\mathcal{S}}\tilde{u}_{\tilde{\pi}(s)}(s)\,c_{\tilde{\pi}(s)}(s)
=\tilde{u}_{\tilde{\pi}}^\top c_{\tilde{\pi}},
\label{eq: sparsity collapse}
\end{equation}
which proves part 1. For part 2, fix a BF $\phi$ in $\Phi$. Similarly, since the $s$-th component of $(I-\gamma P_a)\phi$ depends on $a$ only through the $s$-th row of $P_a$, we can write by Lemma \ref{lem: sparsity of u} that $\sum_{a\in\mathcal{A}}\tilde{u}_a^\top(I-\gamma P_a)\phi =\tilde{u}_{\tilde{\pi}}^\top(I-\gamma P_{\tilde{\pi}})\phi=\nu^\top\phi$, with the last equality being imposed by the dual feasibility in Property \ref{prop: dual}. As $\phi$ ranges over the BFs of $\Phi$, these identities are precisely the entries of part 2. $\square$

\begin{lemma} [Dual-induced successor distribution]
\label{lem: mu normalized}
Under the hypotheses of Lemma \ref{lem: sparsity of u}, and if in addition $\mathbf{1}$ is a
column of $\Phi$, then $\tilde{u}_{\tilde{\pi}}^\top\mathbf{1}=\dfrac{1}{1-\gamma}$ and
\[
\mu^\top:=(1-\gamma) \tilde{u}_{\tilde{\pi}}^\top P_{\tilde{\pi}}
\]
is a normalized vector over $\mathcal{S}$.
\end{lemma}

\textit{Proof of Lemma \ref{lem: mu normalized}.}
Dual feasibility in Lemma \ref{lem: simplified dual properties} holds column-wise and hence for every basis function
$\phi$ in $\Phi$:
$\tilde{u}_{\tilde{\pi}}^\top(I-\gamma P_{\tilde{\pi}})\phi=\nu^\top\phi$.
Taking $\phi=\mathbf{1}$ and using $P_{\tilde{\pi}}\mathbf{1}=\mathbf{1}$ gives
$(1-\gamma)\,\tilde{u}_{\tilde{\pi}}^\top\mathbf{1}=\nu^\top\mathbf{1}=1$.
Non-negativity of $\mu$ follows from $\tilde{u}\ge0$ and $P_{\tilde{\pi}}\ge0$, and
$\mu^\top\mathbf{1}=(1-\gamma)\tilde{u}_{\tilde{\pi}}^\top P_{\tilde{\pi}}\mathbf{1}
=(1-\gamma)\tilde{u}_{\tilde{\pi}}^\top\mathbf{1}=1$. $\square$

\subsection{Other Proofs for ALP}\label{oa: other proofs for ALP}

\textit{Proof of Lemma \ref{lem: epsilon optimal}.} It is widely known that the operator $\mathcal{T}$ is a contraction mapping, i.e.,
\begin{equation}
    \|\mathcal{T}V^* - \mathcal{T}\bar{V} \|_\infty \leq \gamma \|V^* - \bar{V} \|_\infty, \label{eq: contraction mapping}
\end{equation}
if $0<\gamma<1$. Moreover, by the triangle inequality, one can also conclude
\begin{align}
    \|V^* - \bar{V} \|_\infty &\leq \|V^* - \mathcal{T}\bar{V} \|_\infty + \|\mathcal{T}\bar{V}-\bar{V} \|_\infty. \label{eq: triangle}
\end{align}
Therefore, by (\ref{eq: contraction mapping}) and due to the fact that $V^* = \mathcal{T}V^*$, the inequality (\ref{eq: triangle}) can be rewritten as
\begin{align}
            \|V^* - \bar{V} \|_\infty  & \leq \|\mathcal{T}V^* - \mathcal{T}\bar{V} \|_\infty + \|\mathcal{T}\bar{V}-\bar{V} \|_\infty \notag \\
                &\leq \gamma \|V^* - \bar{V} \|_\infty + \|\mathcal{T}\bar{V}-\bar{V} \|_\infty \notag \\
 \implies \|V^* - \bar{V} \|_\infty & \leq \frac{1}{1-\gamma} \|\mathcal{T}\bar{V}-\bar{V} \|_\infty. \label{eq: bound}
\end{align}
Therefore, $\bar{V}$ is $\epsilon$-optimal if $\frac{1}{1-\gamma} \|\mathcal{T}\bar{V}-\bar{V} \|_\infty \leq \epsilon$. Moreover, by the definition of norms, we know that $\|V^* - \bar{V} \|_{1,\nu} \leq  \|V^* - \bar{V} \|_\infty$. Combining this with (\ref{eq: bound}) yields inequality (\ref{eq: Pessimistic error bound}), which completes the proof.  $\square$

\section{Duality under Constraint Sampling} \label{sec: dual proofs for RLP}

This section collects the dual objects associated with \RLP{\Phi}{\mathcal{Z}}, summarized in
Property \ref{prop: rlp duality} of the main text, together with the proofs of the
elementary facts used there and in the appendix of the main paper. Every statement holds for
\RLP{\Phi}{\mathcal{C}} at any finite sample $\mathcal{C}$ and uses no property of
$\mathcal{X}$, $\mathcal{Y}$, or $\psi$.

\begin{property}[Dual of the RLP]
    \label{prop: dual of RLP}
    The dual of \RLP{\Phi}{\mathcal{Z}}, denoted as \RDUAL{\Phi}{\mathcal{Z}},
    corresponds to the dual of \ALP{\Phi} (see \textup{Property \ref{prop: dual}}) with an
    additional condition that $u_a(s) = 0$ for all $(s,a) \in (\mathcal{S}\times \mathcal{A})
    \setminus \mathcal{Z}$.
\end{property}

\textit{Proof of Property \ref{prop: dual of RLP}.}
Assign a dual variable $u_a(s)\ge0$ to the primal constraint indexed by $(s,a)\in\mathcal{Z}$ and write the constraint as $\big[(I-\gamma P_a)\Phi w\big]_s\le c_a(s)$. The Lagrangian of \RLP{\Phi}{\mathcal{Z}} is
\[
\mathscr{L}(w,u)=\nu^\top\Phi w
+\sum_{(s,a)\in\mathcal{Z}}u_a(s)\Big(c_a(s)-\big[(I-\gamma P_a)\Phi w\big]_s\Big).
\]
Because $w$ is unrestricted in sign, $\sup_{w\in\mathbb{R}^\ell}\mathscr{L}(w,u)$ is finite only if the coefficient of $w$ vanishes, that is, only if $\sum_{(s,a)\in\mathcal{Z}}u_a(s)\big[(I-\gamma P_a)\Phi\big]_s=\nu^\top\Phi$, in which case $\sup_{w}\mathscr{L}(w,u)=\sum_{(s,a)\in\mathcal{Z}}u_a(s)c_a(s)$. Extending $u$ to all of $\mathcal{S}\times\mathcal{A}$ by $u_a(s):=0$ for $(s,a)\notin\mathcal{Z}$ leaves both expressions unchanged and rewrites them as $\sum_{a\in\mathcal{A}}u_a^\top(I-\gamma P_a)\Phi=\nu^\top\Phi$ and $\sum_{a\in\mathcal{A}}u_a^\top c_a$, respectively. The dual problem $\min_{u\ge0}\sup_{w}\mathscr{L}(w,u)$ is therefore \DUAL{\Phi} augmented with the restriction $u_a(s)=0$ for all $(s,a)\in(\mathcal{S}\times\mathcal{A})\setminus\mathcal{Z}$. $\square$

\begin{property} [Primal--dual properties of the RLP]
    \label{prop: r-dual properties}
    Let $\tilde{V}=\Phi\tilde{w}$, where $\tilde{w}$ is an optimal solution to \RLP{\Phi}{\mathcal{Z}}, and let $\tilde{u}$ denote an optimal solution to \RDUAL{\Phi}{\mathcal{Z}}. Then, the strong duality and complementary slackness of \RDUAL{\Phi}{\mathcal{Z}} are identical to those stated in \textup{Property \ref{prop: dual properties}} with the additional requirement that $\tilde{u}_a(s) = 0$ for all $(s,a) \in (\mathcal{S}\times \mathcal{A})\setminus \mathcal{Z}$.
\end{property}

\textit{Proof of Property \ref{prop: r-dual properties}.} \RLP{\Phi}{\mathcal{Z}} is a linear program in $\ell$ variables with $|\mathcal{Z}|<\infty$ inequality constraints. It is feasible, since $c\ge0$ implies that $w=0$ satisfies every constraint, and its optimal value is bounded above by Assumption \ref{ass: rlp-bounded 1}. By the strong duality theorem of linear programming, \RLP{\Phi}{\mathcal{Z}} and \RDUAL{\Phi}{\mathcal{Z}} both attain their optima, their optimal values coincide, and every pair of optimal solutions satisfies complementary slackness on the constraints indexed by
$\mathcal{Z}$:
\[
\tilde{u}_a(s)\Big(c(s,a)-\big[(I-\gamma P_a)\tilde{V}\big]_s\Big)=0,
\qquad (s,a)\in\mathcal{Z}.
\]
For $(s,a)\notin\mathcal{Z}$ the same identity holds trivially, since $\tilde{u}_a(s)=0$ by Property \ref{prop: dual of RLP}. The two conditions are therefore exactly those of Property \ref{prop: dual properties}, subject to the additional requirement that $\tilde{u}$ vanish off $\mathcal{Z}$. $\square$

\begin{lemma}[Hybrid dual sparsity]
\label{lem: sparsity of u RLP}
Let $\tilde{V}=\Phi\tilde{w}$, where $\tilde{w}$ is an optimal solution to \RLP{\Phi}{\mathcal{Z}}, and let $\tilde{u}$ be an optimal solution to \RDUAL{\Phi}{\mathcal{Z}}. Let $\tilde{\pi}^{\mathcal{Z}}_{\tilde{V}}$ be the hybrid greedy policy defined in (\ref{eq: hybrid greedy policy}). If $\tilde{\pi}^{\mathcal{Z}}_{\tilde{V}}(s)$ is unique at every state $s\in\mathcal{Z}_{\mathcal{S}}$ with a binding primal constraint (Definition \ref{def: binding (sampling)}), then
\[
\tilde{u}_a(s)=0
\qquad\text{for every }(s,a)\in(\mathcal{S}\times\mathcal{A})\setminus
\big\{\big(s,\tilde{\pi}^{\mathcal{Z}}_{\tilde{V}}(s)\big):s\in\mathcal{Z}_{\mathcal{S}}\big\}.
\]
In particular, $\tilde{u}_a(s)=0$ for all $a\in\mathcal{A}$ whenever $s\in\mathcal{S}\setminus\mathcal{Z}_{\mathcal{S}}$.
\end{lemma}
\textit{Proof of Lemma \ref{lem: sparsity of u RLP}.}
Write $\tilde{\pi}:=\tilde{\pi}^{\mathcal{Z}}_{\tilde{V}}$ and fix a pair $(s,a)$ outside
$\{(s,\tilde{\pi}(s)):s\in\mathcal{Z}_{\mathcal{S}}\}$. We show $\tilde{u}_a(s)=0$ in each of
three exhaustive cases.

\emph{Case 1: $(s,a)\notin\mathcal{Z}$.} Then $\tilde{u}_a(s)=0$ by Property
\ref{prop: dual of RLP}. This case contains every pair with $s\notin\mathcal{Z}_{\mathcal{S}}$,
which proves the final claim.

\emph{Case 2: $(s,a)\in\mathcal{Z}$ and $s$ has no binding primal constraint.} The constraint
indexed by $(s,a)$ is slack, so complementary slackness (Property \ref{prop: r-dual properties})
gives $\tilde{u}_a(s)=0$.

\emph{Case 3: $(s,a)\in\mathcal{Z}$ and $s$ has a binding primal constraint.} We first show that
the binding actions at $s$ are exactly the minimizers defining $\tilde{\pi}(s)$. Feasibility of
$\tilde{w}$ gives $\tilde{V}(s)\le(\mathcal{T}_{a'}\tilde{V})(s)$ for every
$a'\in\mathcal{Z}_{\mathcal{A}}(s)$, hence $\tilde{V}(s)\le(\mathcal{L}_{\mathcal{Z}}\tilde{V})(s)$.
If the constraint binds at some $a_0\in\mathcal{Z}_{\mathcal{A}}(s)$, then
$\tilde{V}(s)=(\mathcal{T}_{a_0}\tilde{V})(s)\ge(\mathcal{L}_{\mathcal{Z}}\tilde{V})(s)$.
Therefore $\tilde{V}(s)=(\mathcal{L}_{\mathcal{Z}}\tilde{V})(s)$, and an action binds at $s$ if and
only if it attains the minimum in (\ref{eq: hybrid Bellman operator}). By the assumed uniqueness,
that minimizer is $\tilde{\pi}(s)$. Since $a\neq\tilde{\pi}(s)$, we obtain
$(\mathcal{T}_{a}\tilde{V})(s)>(\mathcal{L}_{\mathcal{Z}}\tilde{V})(s)=\tilde{V}(s)$. The
constraint indexed by $(s,a)$ is therefore slack, and complementary slackness gives
$\tilde{u}_a(s)=0$. $\square$

\begin{lemma} [Policy-reduced dual properties under sampling] \label{lem: simplified dual properties RLP}
    Let $\tilde{V} = \Phi\tilde{w}$, where $\tilde{w}$ is an optimal solution to \RLP{\Phi}{\mathcal{Z}}, and let $\tilde{u}$ be an optimal solution to \RDUAL{\Phi}{\mathcal{Z}}. If the hybrid greedy policy $\tilde{\pi}^\mathcal{Z}_{\tilde{V}}(s)$ is unique for any state $s\in\mathcal{Z}_{\mathcal{S}}$ with a binding primal constraint (Definition \ref{def: binding (sampling)}), the following properties hold.
    \begin{enumerate}
        \item \textbf{Strong duality}: $z(\tilde{u}) = \sum_{a \in \mathcal{A}} \tilde{u}_a^\top c_a
        = \tilde{u}_{\tilde{\pi}^\mathcal{Z}_{\tilde{V}}}^\top c_{\tilde{\pi}^\mathcal{Z}_{\tilde{V}}}
        = \nu^\top \tilde{V}$.
        \item \textbf{Dual feasibility}: $\sum_{a\in \mathcal{A}} \tilde u_a^\top \left(I-\gamma P_a
        \right ) \Phi = \tilde u_{\tilde{\pi}^\mathcal{Z}_{\tilde{V}}}^\top \left(I-\gamma
        P_{\tilde{\pi}^\mathcal{Z}_{\tilde{V}}} \right ) \Phi = \nu^\top \Phi$.
    \end{enumerate}
\end{lemma}

\textit{Proof of Lemma \ref{lem: simplified dual properties RLP}.}
Write $\tilde{\pi}:=\tilde{\pi}^{\mathcal{Z}}_{\tilde{V}}$ for brevity. By Lemma \ref{lem: sparsity of u RLP}, the only pairs carrying positive dual weight are
$\{(s,\tilde{\pi}(s)):s\in\mathcal{Z}_{\mathcal{S}}\}$, so
\[
\sum_{a\in\mathcal{A}}\tilde{u}_a^\top c_a
=\sum_{s\in\mathcal{Z}_{\mathcal{S}}}\tilde{u}_{\tilde{\pi}(s)}(s)\,c_{\tilde{\pi}(s)}(s)
=\sum_{s\in\mathcal{S}}\tilde{u}_{\tilde{\pi}}(s)\,c_{\tilde{\pi}}(s)
=\tilde{u}_{\tilde{\pi}}^\top c_{\tilde{\pi}},
\]
where the middle equality holds because $\tilde{u}_{\tilde{\pi}}(s)=\tilde{u}_{\tilde{\pi}(s)}(s)=0$
for $s\notin\mathcal{Z}_{\mathcal{S}}$, whatever action the hybrid greedy policy assigns there. Invoking strong duality in Property \ref{prop: r-dual properties} gives part 1.
Taking $(I-\gamma P_a)\phi$ for a BF $\phi$ in $\Phi$, and using that the $s$-th component of
$(I-\gamma P_a)\phi$ depends on $a$ only through the $s$-th row of $P_a$, gives
$\sum_{a\in\mathcal{A}}\tilde{u}_a^\top(I-\gamma P_a)\phi
=\tilde{u}_{\tilde{\pi}}^\top(I-\gamma P_{\tilde{\pi}})\phi=\nu^\top\phi$, where the last equality is
dual feasibility in Property \ref{prop: r-dual properties}. Ranging over the BFs of $\Phi$ yields
part 2. $\square$

\begin{lemma} [Dual-induced successor distribution under sampling]
\label{lem: mu normalized (sampling)}
Let $\tilde{u}$ be an optimal solution to \RDUAL{\Phi}{\mathcal{Z}} with $\Phi = [\mathbf{1}, \phi_2, \dots, \phi_\ell]$, and let $\tilde{\pi}^\mathcal{Z}_{\tilde{V}}$ be the hybrid greedy policy with respect to the value estimate $\tilde{V}=\Phi\tilde{w}$, where $\tilde{w}$ is an optimal solution to \RLP{\Phi}{\mathcal{Z}}. Suppose $\tilde{\pi}^\mathcal{Z}_{\tilde{V}}(s)$ is unique at every state $s$ with a binding primal constraint (Definition \ref{def: binding (sampling)}). Then
$\tilde{u}_{\tilde{\pi}^\mathcal{Z}_{\tilde{V}}}^\top\mathbf{1}=\frac{1}{1-\gamma}$, and
\[\mu^\top:=(1-\gamma)\,\tilde{u}_{\tilde{\pi}^\mathcal{Z}_{\tilde{V}}}^\top
P_{\tilde{\pi}^\mathcal{Z}_{\tilde{V}}}
\]
is a normalized vector over $\mathcal{S}$.
\end{lemma}

\textit{Proof of Lemma \ref{lem: mu normalized (sampling)}.}
Dual feasibility in Lemma \ref{lem: simplified dual properties RLP} holds column-wise and hence for
every basis function $\phi$ in $\Phi$:
$\tilde{u}_{\tilde{\pi}^\mathcal{Z}_{\tilde{V}}}^\top(I-\gamma P_{\tilde{\pi}^\mathcal{Z}_{\tilde{V}}})\phi
=\nu^\top\phi$. Taking $\phi=\mathbf{1}$ and using $P_{\tilde{\pi}^\mathcal{Z}_{\tilde{V}}}\mathbf{1}
=\mathbf{1}$ gives $(1-\gamma)\,\tilde{u}_{\tilde{\pi}^\mathcal{Z}_{\tilde{V}}}^\top\mathbf{1}
=\nu^\top\mathbf{1}=1$. Non-negativity of $\mu$ follows from $\tilde{u}\ge0$ and
$P_{\tilde{\pi}^\mathcal{Z}_{\tilde{V}}}\ge0$, and
$\mu^\top\mathbf{1}=(1-\gamma)\tilde{u}_{\tilde{\pi}^\mathcal{Z}_{\tilde{V}}}^\top
P_{\tilde{\pi}^\mathcal{Z}_{\tilde{V}}}\mathbf{1}
=(1-\gamma)\tilde{u}_{\tilde{\pi}^\mathcal{Z}_{\tilde{V}}}^\top\mathbf{1}=1$. $\square$

\textit{Proof of Lemma \ref{lem: L_Y monotone}.}
If $s\notin\mathcal{Y}_{\mathcal{S}}$ then $(\mathcal{L}_{\mathcal{Y}}\bar{V})(s)=\bar{V}(s)$ by
(\ref{eq: hybrid Bellman operator}) and the claim holds with equality. If
$s\in\mathcal{Y}_{\mathcal{S}}$ then every pair $(s,a)$ with $a\in\mathcal{Y}_{\mathcal{A}}(s)$ lies
in $\mathcal{Y}$, so the hypothesis gives
$\bar{V}(s)\le c(s,a)+\gamma\sum_{s'}p(s'\mid s,a)\bar{V}(s')$ for every such $a$. Minimizing the
right-hand side over $a\in\mathcal{Y}_{\mathcal{A}}(s)$ preserves the inequality and yields $\bar{V}(s)\le(\mathcal{L}_{\mathcal{Y}}\bar{V})(s)$. $\square$

\section{Supplementary Discussion of the Sampling Design}\label{app: conditions}

\subsection{Remarks on the Bounded Action-Restriction Gap}\label{app: e remarks}

This section expands the discussion of Assumption \ref{ass: Y suboptimality gap} given in the main text.

Assumption \ref{ass: Y suboptimality gap} bounds how much the value backup computed by the $\mathcal{Y}$-restricted greedy minimization, with respect to $\tilde{V}_k$, can exceed the backup available over the full sampled action set used in \RLP{\Phi_{k+1}}{\mathcal{Z}^{k+1}}, uniformly over the curated states. It is easily satisfied in practice, for two independent reasons.

First, because $\mathcal{Z}^{k+1}=\mathcal{X}^{k+1}\cup\mathcal{Y}$, we have $\mathcal{Z}^{k+1}_{\mathcal{A}}(s)\supseteq\mathcal{Y}_{\mathcal{A}}(s)$ for every $s\in\mathcal{Y}_{\mathcal{S}}$, with equality whenever $s\notin\mathcal{X}^{k+1}_{\mathcal{S}}$. At such a state, the two minimizations in (\ref{eq: Y suboptimality gap}) run over the same action set and the term in the bracket is exactly zero, i.e., $\big(\mathcal{L}_{\mathcal{Y}}\tilde{V}_k\big)(s)-\big(\mathcal{L}_{\mathcal{Z}^{k+1}}\tilde{V}_k\big)(s) = 0$ if $\mathcal{Z}^{k+1}_{\mathcal{A}}(s)=\mathcal{Y}_{\mathcal{A}}(s)$ for some $s$. Since the states in $\mathcal{Y}_{\mathcal{S}}$ need not belong to $\mathcal{X}^{k+1}_{\mathcal{S}}$, and in large or continuous state spaces generically will not, the exact equality $\mathcal{Z}^{k+1}_{\mathcal{A}}(s)=\mathcal{Y}_{\mathcal{A}}(s)$ is the common case.

Second, and independently of overlap, the bracket at $s\in\mathcal{Y}_{\mathcal{S}}$ can be strictly positive only if $\mathcal{X}^{k+1}_{\mathcal{A}}(s)$ contains an action achieving a strictly lower $\tilde{V}_k$-backup than every action already retained in $\mathcal{Y}_{\mathcal{A}}(s)$. Because many actions are retained per state when constructing $\mathcal{Y}$, unlike $\mathcal{X}$, the set $\mathcal{Y}_{\mathcal{A}}(s)$ can be made action-rich at every curated state, which makes it increasingly unlikely that the comparatively small, independently drawn set $\mathcal{X}^{k+1}_{\mathcal{A}}(s)$ improves upon it. In either case, Assumption \ref{ass: Y suboptimality gap} holds with $e$, which is typically small and often exactly zero.

\subsection{Remarks on Propagated Correction Dominance}\label{app: dominance remarks}

This section expands the discussion of Assumption \ref{ass: propagated dominance} and of the decomposition (\ref{eq: part b decomposition}) given in the main text. Term (II) is always non-negative. However, term (I) is non-negative exactly when $\tilde{V}_k$ satisfies the constraint indexed by $(s,a)$. Since $\tilde{V}_k$ is optimal for \RLP{\Phi_k}{\mathcal{Z}^{k}}, it satisfies every constraint indexed by $\mathcal{Z}^{k}$, so term (I) can be negative only at a pair drawn afresh into $\mathcal{X}^{k+1}$ and absent from $\mathcal{Z}^{k}$. Assumption \ref{ass: propagated dominance} therefore asks that, in dual-weighted aggregate, the one-step propagation of the correction just made by the new BF dominate whatever violations the fresh draw exposes.

Assumption \ref{ass: propagated dominance} could be mild for three reasons. First, it is an aggregate condition and not a point-wise one. Every fresh state whose constraint $\tilde{V}_k$ already satisfies contributes a non-negative term to (\ref{eq: propagated dominance}), and these are the large majority. By Corollary \ref{cor: violations under Z} applied to \RLP{\Phi_k}{\mathcal{Z}^{k}} with a unique optimal solution, at most a $\xi$-fraction of the $\psi$-mass is violated by $\tilde{V}_k$ with a probability of at least $1-\delta$. Only violated states can subtract, and they must subtract more than all the remaining states add. Moreover, (\ref{eq: propagated dominance}) holds trivially when no fresh state outside $\mathcal{Y}_{\mathcal{S}}$ carries dual weight.

Second, only states carrying dual weight matter. If $\tilde{u}^{k+1}$ is chosen as a basic optimal solution of \RDUAL{\Phi_{k+1}}{\mathcal{Z}^{k+1}}, it has at most $\ell_{k+1}$ positive components, because \RDUAL{\Phi_{k+1}}{\mathcal{Z}^{k+1}} has $\ell_{k+1}$ equality constraints and non-negative variables. The sum in (\ref{eq: propagated dominance}) thus ranges over at most $\ell_{k+1}$ states, whereas $\mathcal{X}^{k+1}$ contributes $M\gg \ell_{k+1}$ of them; the constraints that matter are few, even though the sample is large.

Third, and most usefully, the condition is governed by a design lever we already control. Writing $\tilde{\pi}:=\tilde{\pi}^{\mathcal{Z}^{k+1}}_{\tilde{V}_{k+1}}$ and $r_k:=\mathcal{L}_{\mathcal{Y}}\tilde{V}_k-\tilde{V}_k$, a sufficient condition for (\ref{eq: propagated dominance}) is that, at every state $s\in\mathcal{Z}^{k+1}_{\mathcal{S}}\setminus\mathcal{Y}_{\mathcal{S}}$ carrying dual weight,
\begin{equation}
\gamma\sum_{s'\in\mathcal{Y}_{\mathcal{S}}}p\big(s'\mid s,\tilde{\pi}(s)\big)\,r_k(s')
\ \geq\
\Big(\tilde{V}_k(s)-\big(\mathcal{T}_{\tilde{\pi}(s)}\tilde{V}_k\big)(s)\Big)^{+} ,
\label{eq: pointwise dominance}
\end{equation} i.e.\ that a fresh state whose constraint cuts transitions, in one step, into the region where the new BF has just made corrections. Since Section \ref{sec: sampling architecture} places no distributional restriction on how $\mathcal{Y}_{\mathcal{S}}$ is built, requiring only that it be fixed in advance, the analyst can promote (\ref{eq: pointwise dominance}) directly by constructing $\mathcal{Y}_{\mathcal{S}}$ through trajectory or chain-walk exploration started from $\nu$ and propagated under the predefined policies of that section: such a set is close to being \emph{closed} under the resulting dynamics, which is exactly what makes $p(\mathcal{Y}_{\mathcal{S}}\mid s,a)$ large at the states the returned policy actually reaches.

\subsection{Directions for Tightening the Residual Bound}\label{app: future}

We see at least two natural ways to mitigate the accuracy--tractability tension discussed in Section \ref{sec: Discussions} of the main paper, both of which we view as promising directions rather than contributions of the present paper. The first is to extrapolate the value information obtained on $\mathcal{Y}_{\mathcal{S}}$ to unsampled states, for instance via local interpolation, kernel-weighted averaging, or a Gaussian-process posterior fit to the values $\{\Phi_K\tilde w_K(s):s\in\mathcal{Y}_{\mathcal{S}}\}$ (\citealp{rasmussen2003gaussian}). Such an extrapolated surface would let one replace the pointwise definition $(\mathcal{L}_{\mathcal{Y}}v)(s)=v(s)$ for $s\in\mathcal{S} \setminus\mathcal{Y}_{\mathcal{S}}$ in (\ref{eq: hybrid Bellman operator}) with a data-driven estimate, effectively enlarging the practical reach of $\mathcal{Y}_{\mathcal{S}}$ without enlarging the RLP itself. The second is to directly learn a statistical approximation of the hybrid Bellman operator on the unsampled complement $\mathcal{S}\setminus\mathcal{Y}_{\mathcal{S}}$, in the spirit of fitted value iteration with generalization guarantees (\citealp{munos2008finite}), so that the identity branch of (\ref{eq: hybrid Bellman operator}) is replaced by a regression estimate of $\min_{a}\{c(s,a)+\gamma\sum_{s'}p(s'\mid s,a)v(s')\}$ trained on the states in $\mathcal{Y}_{\mathcal{S}}$ and generalized elsewhere. Either route would let $\psi_{\tilde\pi}(\mathcal{Y}_{\mathcal{S}})$ in Theorem \ref{thm: bellman residual at termination} be effectively replaced by coverage of an extended, generalized $\mathcal{Y}_{\mathcal{S}}$, tightening the bound toward uniform convergence without a proportional increase in the number of exact backups computed. Working out the corresponding approximation-error and sample-complexity trade-offs, however, is a substantial undertaking on its own and falls outside the scope of the present paper. Thus, we leave it to future work.

\section{Omitted Proofs for the Constraint-Sampling Setting}\label{sec: proofs sampling}

\subsection{Discounted Occupancy Distribution}
\label{oa: occupancy}

\begin{property} [Discounted occupancy distribution] \label{prop: occupancy}
Let $\nu$ be the initial state distribution and $\pi$ be a policy. Then, the following statements hold:
\begin{enumerate}
    \item[(a)] The vector $ d_\nu^\pi =  (1-\gamma)\big(\nu^\top(I-\gamma P_\pi)^{-1}\big)^\top$ is a probability distribution over $\mathcal{S}$.
    \item[(b)] The product $\dfrac{1}{|\mathcal{A}|}d_\nu^\pi$ is a probability distribution over $\mathcal{S} \times \mathcal{A}$.
\end{enumerate}
\end{property}

\textit{Proof of Property \ref{prop: occupancy}.} By the Neumann expansion,
$(I-\gamma P_\pi)^{-1}=\sum_{t\ge0}(\gamma P_\pi)^t$, which converges because $\gamma\in(0,1)$
and $P_\pi$ is a stochastic matrix, and which is non-negative entry-wise. Hence
$d_\nu^\pi\ge0$. For the total mass, $P_\pi\mathbf{1}=\mathbf{1}$ gives
$(I-\gamma P_\pi)^{-1}\mathbf{1}=\sum_{t\ge0}\gamma^{t}\mathbf{1}=\frac{1}{1-\gamma}\mathbf{1}$,
so that
\[
(d_\nu^\pi)^\top\mathbf{1}
=(1-\gamma)\,\nu^\top(I-\gamma P_\pi)^{-1}\mathbf{1}
=(1-\gamma)\cdot\frac{1}{1-\gamma}\,\nu^\top\mathbf{1}=1,
\]
which is (a). Statement (b) follows by assigning mass $\frac{1}{|\mathcal{A}|}d_\nu^\pi(s)$ to
each pair $(s,a)$: the entries are non-negative and sum to
$\sum_{s\in\mathcal{S}}\sum_{a\in\mathcal{A}}\frac{1}{|\mathcal{A}|}d_\nu^\pi(s)
=\sum_{s\in\mathcal{S}}d_\nu^\pi(s)=1$. $\square$

\subsection{Proof of Lemma \ref{lem: beta sufficient}}
\label{oa: proof beta}

\textit{Proof of Lemma \ref{lem: beta sufficient}.}
Fix $\tilde{w}\in\mathcal{W}(\Phi)$. If $s\in\mathcal{B}^{*}(V_{\tilde{w}})$ then, by definition, the
pair $(s,\pi^*(s))$ lies in $\mathcal{B}(V_{\tilde{w}})$. Hence
$\{(s,\pi^*(s)):s\in\mathcal{B}^{*}(V_{\tilde{w}})\}\subseteq\mathcal{B}(V_{\tilde{w}})$, and these
pairs are distinct. Therefore
\[
d_\nu^{\pi^*}\big(\mathcal{B}^{*}(V_{\tilde{w}})\big)=\sum_{s\in\mathcal{B}^{*}(V_{\tilde{w}})}d_\nu^{\pi^*}(s)
\ \le\ \beta_{\max}\!\!\sum_{s\in\mathcal{B}^{*}(V_{\tilde{w}})}\!\!\psi\big(s,\pi^*(s)\big)
\ \le\ \beta_{\max}\,\psi\big(\mathcal{B}(V_{\tilde{w}})\big).
\]
The bound holds for all $\tilde{w}\in\mathcal{W}(\Phi)$, which completes the proof. $\square$

\subsection{Term-by-Term Reading of the Improvement Bound}
\label{oa: reading of bound}

The bound (\ref{eq: major improvement sampling}) decomposes the guaranteed net improvement of
an HBOBF into four terms, each admitting a distinct interpretation.

The leading term $\|\mathcal{L}_{\mathcal{Y}}\tilde{V}_k-\tilde{V}_{k}\|_{1,\nu}$ is the
$\nu$-weighted magnitude of the empirical Bellman error, and it contributes a monotone
improvement to the approximation whenever the error is non-zero.

The function $g$ collects a second-order contribution that the leading term does not capture,
arising through two distinct mechanisms (see (\ref{eq: g})). At the curated states
$s\in\mathcal{Y}_{\mathcal{S}}$, the correction just made by the new BF is propagated one step
downstream along the trajectories of the greedy policy $\hat{\pi}_{k+1}$. This part of $g$ is
non-negative unconditionally, as shown in the proof of Lemma \ref{lem: g nonnegative}. At the
states contributed by $\mathcal{X}^{k+1}$ alone, the corresponding terms may be negative, since
a freshly drawn pair can expose a constraint that $\tilde{V}_k$ violates. The function $g$ is
therefore not non-negative by construction. The discussion of Assumption
\ref{ass: propagated dominance} indicates why negative values are unlikely rather than
impossible: the quantity is an aggregate over the fresh states, not a pointwise requirement,
and the states whose constraints $\tilde{V}_k$ violates carry at most a $\xi$-fraction of the
$\psi$-mass with probability at least $1-\delta$. Assumption \ref{ass: propagated dominance} is
precisely the condition under which the aggregate is non-negative, and Lemma
\ref{lem: g nonnegative} then gives $g\ge0$, so that the propagated term can only reinforce the
net approximation improvement.

The term $-\,e\sum_{s\in\mathcal{Y}_{\mathcal{S}}}\hat{u}_{k+1}(s)$ is the dual-weighted price
of computing the hybrid Bellman operator over $\mathcal{Y}_{\mathcal{A}}(s)$ rather than over
the full $\mathcal{Z}^{k+1}_{\mathcal{A}}(s)$, governed by the action-restriction gap $e$ of
Assumption \ref{ass: Y suboptimality gap}. It vanishes whenever
$\mathcal{Y}_{\mathcal{S}}$ and $\mathcal{X}^{k+1}_{\mathcal{S}}$ do not intersect, or whenever
$\mathcal{Y}_{\mathcal{A}}(s)$ is populated densely enough to already contain a best action
available in $\mathcal{X}^{k+1}_{\mathcal{A}}(s)$.

The final term is inherited from the constraint-sampling guarantee of Theorem
\ref{thm: ALP vs RLP}, applied to \RLP{\Phi_{k+1}}{\mathcal{Z}^{k+1}}. It is the probabilistic
cost of relaxing \ALP{\Phi_{k+1}} to a sampled program, which shrinks with the violation level
$\xi$, hence with a larger sample size $M$.

\subsection{Comparison with Sample-Complexity Results for Iterative RL Methods}
\label{oa: discussion}

The certification sample of the framework obeys
$M=\Theta\big(\xi^{-1}(\ell\ln\xi^{-1}+\ln\delta^{-1})+\ell\big)$ and does not depend on
$|\mathcal{S}\times\mathcal{A}|$. This contrasts with the polynomial dependence on
$N|\mathcal{A}|$ and on the effective horizon $1/(1-\gamma)$ established for $Q$-learning and
temporal-difference learning (\citealp{bhandari2021finite,li2020sample}). The generating
sample $\mathcal{Y}$, meanwhile, carries no distributional requirement and may be enlarged at
will at no cost to the feasibility guarantee, since by Corollary \ref{cor: violations under Z}
the scenario count is $|\mathcal{X}|$ and not $|\mathcal{Z}|$. The two samples are thus
governed by unrelated budgets: one by the statistical requirement of the certificate, the
other by the computational cost of a Bellman backup.

\section{Implementation Notes for the Numerical Studies}
\label{oa: implementation}

\subsubsection{Perishable Inventory Control}

Our experimental setup is inspired by two recent studies (\cite{abouee2025platelet,
lin_revisiting_2020}). In all experiments, we set the demand parameters $(\iota, q)$ to
$(11, 0.6145)$ and the truncation point $\bar{D}$ to $20$. The remaining problem parameters vary across
scenarios to ensure a diverse set of test cases (see Table \ref{tab: PIC benchmark}). The
service level $\alpha$ is fixed at $99\%$ throughout the experiments.

To incorporate expert recommendations on the choice of BFs for the PIC problem, we adopt the BFs suggested by \cite{lin_revisiting_2020} and by \cite{abouee2025platelet}. These recommendations can be aggregated into $\Phi_{fixed} (s) = \left [1 \right ] \cup [(x_1-D)^+, (x_1+x_2-2D)^+, (x_1+x_2+z_1-3D)^+, (2D - x_1-x_2-z_1)^+, (D-x_2-z_1)^+ | D \in \{\mathbb{E}[D],D^{0.25},D^{0.5}\}   ] \cup [x_1, x_2, \cdots , x_{m-1},x_1^2, x_2^2, \cdots , x_{m-1}^2, x_1^3, x_2^3, \cdots , x_{m-1}^3, z_1, z_2, \cdots , z_{l},z_1^2, z_2^2, \cdots , z_{l}^2, z_1^3, z_2^3, \cdots , z_{l}^3, \phi_{empirical} ]$ for all $s \in \mathcal{S}$, where $D^{0.25}$ and $D^{0.5}$ denote the $25^{\text{th}}$ and $50^{\text{th}}$ quantiles of the demand distribution, and $\phi_{empirical}$ represents the optimal value function of the corresponding nonperishable inventory problem. We refer to $\Phi_{fixed}$ as the ``Fixed BF'' approach in our experiments.

\subsubsection{Queueing Control}

In our experiments, the continuous-time discount rate $\zeta$ is selected such that the
normalized discount factor $\vartheta$ equals 0.99. We set $p_j=j$ for all
$j\in\{1,\cdots,J\}$, and define the service rates $\omega'_j$ as equally spaced values within
the interval $[1,2]$. The arrival rates $\lambda_j'$ are randomly generated to ensure that the
service load $\rho=\sum_{j=1}^J\frac{\lambda_j'}{\kappa\omega_j'}$ matches a specific value. In
our experiments, $\rho$ is varied across a wide range of values in $\{0.8, 1, 3, 5, 10\}$. The
parameters $J$ and $\kappa$ are also adjusted to create multiple scenarios (see Table
\ref{tab: QC Benchmark}). For the fixed BF approach, we employ affine BFs defined as
$\Phi(s)=[1, s_1,\cdots, s_J]$ for all $s\in\mathcal{S}$ (see
\cite{samiedaluie2024approximate} for more details).

\subsection{Effect of $\nmax$ on the Shrunken Method}
\label{oa: nmax}

\begin{figure}[htbp!]
\FIGURE{
    \includegraphics[width=0.5\textwidth]{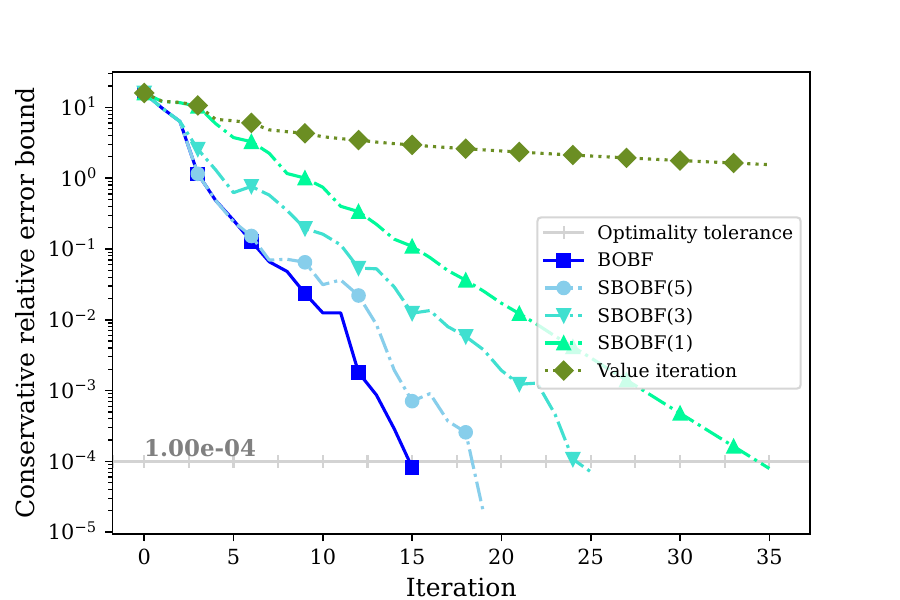}
}{\normalfont Conservative relative error bound $\frac{1}{(1-\gamma) \nu^\top\tilde{V}_k}\parallel \mathcal{T}\tilde{V}_k-\tilde{V}_k\parallel_\infty$ of BOBF and SBOBF approaches on a PIC instance with $(f, h, \eta, \theta) = (100, 1, 20, 5)$, $(l, m) = (1, 4)$, and $\gamma=0.99$. \label{fig: PIC SBOBF}}{}
\end{figure}

Reducing the parameter $\nmax$ lowers the computational cost per iteration for the SBOBF
approach, as fewer BFs are retained in $\Phi$. However, our experiments show that when
$\nmax$ is too small, SBOBF requires significantly more iterations to converge. As illustrated
in Figure \ref{fig: PIC SBOBF}, SBOBF(5) performs comparably to BOBF, whereas SBOBF(1)
converges much more slowly. This highlights a trade-off between per-iteration and overall
computational cost: increasing the number of BFs raises the computational burden per iteration
but reduces the total number of iterations needed for convergence, thereby lowering overall
cost. Consequently, the optimal choice of $\nmax$ depends on the specific application and the
available computational resources.

Despite the similarities between SBOBF(1) and the value iteration algorithm, our experiments
demonstrate the advantage of SBOBF(1) over value iteration. In value iteration, the update
follows $\tilde{V}_k\rightarrow\tilde{V}_{k+1}=\mathcal{T}\tilde{V}_k$; whereas in SBOBF(1),
the process is
$\tilde{V}_k\rightarrow\Phi_{k+1}=[\mathcal{T}\tilde{V}_k]\rightarrow\textup{ALP(}\Phi_{k+1}\textup{)}\rightarrow\tilde{V}_{k+1}$,
so that the single retained BF is rescaled by the optimal weight rather than accepted as is.
Although SBOBF(1) solves a more complex problem in each iteration, Figure
\ref{fig: PIC SBOBF} shows that it converges within 35 iterations, whereas value iteration
requires more than 900 iterations to achieve the optimality tolerance. Consequently, the
overall computational effort for SBOBF(1) may be lower than that of value iteration due to the
substantial improvement achieved in each iteration.

Figures \ref{fig: PIC BOBF} and \ref{fig: PIC SBOBF} indicate that, even under a conservative
error bound, BOBF consistently outperforms SBOBF. In essence, incorporating more BFs enables
the BOBF approach to substantially improve the approximated value function, thereby achieving
a tighter worst-case error bound compared to the improvements offered by SBOBF.

\section{Sampling Design for the Numerical Studies}
\label{oa: sampling}

This section specifies the sampling mechanisms used in the sampled experiments of
Section~\ref{sec: Numerical Studies}, and instantiates the two-sample architecture of
Section~\ref{sec: sampling architecture} for each application. Instance parameters, the demand
and arrival specifications, and the composition of the Fixed BF vectors are reported separately
in \online{oa: implementation}.

Recall the two roles the samples play. The set $\mathcal{X}^{k}$ is drawn i.i.d.\ from a single
distribution $\psi$ over $(\mathcal{S}\times\mathcal{A})$ and is the object to which
Proposition~\ref{prop: probability of violations} and Corollary~\ref{cor: violations under Z}
apply. The curated set $\mathcal{Y}$  exists to make the hybrid Bellman operator $\mathcal{L}_{\mathcal{Y}}$ computable,
and it is subject to no distributional restriction beyond being fixed before any realization of
$\mathcal{X}^{0}$. The program solved at iteration $k$ is
$\RLP{\Phi_k}{\mathcal{Z}^{k}}$ with $\mathcal{Z}^{k}=\mathcal{X}^{k}\cup\mathcal{Y}$.

Throughout, $\mathcal{F}\subseteq(\mathcal{S}\times\mathcal{A})$ denotes the admissible set of the
application at hand, $M=|\mathcal{X}^{k}|$ the number of pairs drawn at each iteration $k$ of
Algorithm~\ref{alg: HBOBF and SHBOBF}, $n_{\mathcal{S}}=|\mathcal{Y}_{\mathcal{S}}|$ the number of
curated states, and $n_{\mathcal{A}}$ the number of candidate actions attached to each curated
state. Numerical values are collected in Table~\ref{tab: sampling parameters}.

\subsection{The Sampling Distribution \texorpdfstring{$\psi$}{psi} and Feasibility Rejection}
\label{oa: psi}

In both applications, $\psi$ is obtained by accept--reject from a proposal that is specified
hierarchically. A \emph{joint proposal} $\mathbb{Q}$ on $(\mathcal{S}\times\mathcal{A})$ consists
of a state proposal $\mathbb{Q}_{\mathcal{S}}$ and a conditional action proposal
$\mathbb{Q}_{\mathcal{A}}(\cdot\mid s)$, and assigns to a pair $(s,a)$ the mass
\begin{align}
\mathbb{Q}(s,a)\;=\;\mathbb{Q}_{\mathcal{S}}(s)\,\mathbb{Q}_{\mathcal{A}}(a\mid s).
\label{eq: Q hierarchical}
\end{align}
Neither $\mathbb{Q}_{\mathcal{S}}$ nor $\mathbb{Q}_{\mathcal{A}}(\cdot\mid s)$ is required to
respect admissibility. Both of them are chosen for ease of simulation. Admissibility is imposed by
rejection: a candidate pair $(s,a)\sim\mathbb{Q}$ is accepted if $(s,a)\in\mathcal{F}$, and is
otherwise discarded \emph{in both coordinates} and redrawn. Accepted draws are i.i.d.\
realizations of the single distribution
\begin{align}
\psi(\cdot)\;=\;\mathbb{Q}\big(\,\cdot\,\big|\,\mathcal{F}\big),
\label{eq: psi definition}
\end{align}
supported on $\mathcal{F}$. Algorithm~\ref{alg: X sampling} states the mechanism; the two
applications differ only in the triple $(\mathbb{Q}_{\mathcal{S}},\mathbb{Q}_{\mathcal{A}},
\mathcal{F})$ supplied to it, given in Sections~\ref{oa: sampling PIC} and \ref{oa: sampling QC}.

\begin{algorithm}[htbp!]
\caption{Drawing the i.i.d.\ sample $\mathcal{X}^{k}$ from $\psi=\mathbb{Q}(\cdot\mid\mathcal{F})$.}
\label{alg: X sampling}
\textbf{Input}: state proposal $\mathbb{Q}_{\mathcal{S}}$, conditional action proposal
$\mathbb{Q}_{\mathcal{A}}(\cdot\mid s)$, admissible set $\mathcal{F}$, sample size $M$.\\
\textbf{Output}: $\mathcal{X}^{k}$, a collection of $M$ i.i.d.\ draws from $\psi$.\\
Initialize $\mathcal{X}^{k}\leftarrow\emptyset$.\\
\While{$|\mathcal{X}^{k}|<M$}{
    Draw a candidate state $s\sim\mathbb{Q}_{\mathcal{S}}$.\\
    Draw a candidate action $a\sim\mathbb{Q}_{\mathcal{A}}(\cdot\mid s)$.\\
    \uIf{$(s,a)\in\mathcal{F}$}{
        $\mathcal{X}^{k}\leftarrow\mathcal{X}^{k}\cup\{(s,a)\}$\\
    }
    \Else{
        Discard $s$ and $a$ together.\\
    }
}
\Return{$\mathcal{X}^{k}$}\\
\end{algorithm}

\paragraph{Redrawing $\mathcal{X}^{k}$ at every iteration.} A fresh sample is drawn at each
iteration $k$, on the same random stream and independently of all earlier draws. This is not
merely a variance-reduction device. The basis $\Phi_k$ produced by
Algorithm~\ref{alg: HBOBF and SHBOBF} is a function of the iterates
$\tilde{V}_0,\dots,\tilde{V}_{k-1}$ and hence of $\mathcal{X}^{0},\dots,\mathcal{X}^{k-1}$.
Corollary~\ref{cor: violations under Z} requires the basis to be formed independently of the
sample against which the violation bound is asserted, and redrawing restores exactly that
independence at every iteration. Reusing a single frozen $\mathcal{X}$ across iterations would
forfeit the guarantee from the second iteration onward.

\paragraph{Vectorized implementation.} The implementation executes the loop of
Algorithm~\ref{alg: X sampling} in blocks. It proposes a batch of candidate pairs, retains the
admissible ones, and repeats until $M$ have been collected, sizing each batch from the number
still outstanding. Because candidates are proposed i.i.d.\ and are screened one pair at a time,
this leaves the law of the accepted draws unchanged.

\subsection{Perishable Inventory Control}
\label{oa: sampling PIC}

\subsubsection{States, actions, and the admissible set.}
A state is the concatenation of the outstanding pipeline and the on-hand inventory graded by
remaining shelf life,
\[
s=(z_l,\dots,z_1,x_m,\dots,x_1)\in\mathbb{R}_+^{\mathcal{I}},
\qquad \mathcal{I}=l+m,
\]
where $z_j$ is the quantity ordered $l-j+1$ periods ago and scheduled to arrive in $j$ periods,
and $x_i$ is the quantity of on-hand stock with $i$ periods of remaining shelf life. Therefore, $x_1$ perishes at the end of the current period and $z_1$ arrives at the start of the next
one. Coordinates are indexed in this order throughout, so that coordinate $1$ of $s$ is $z_l$ and
coordinate $\mathcal{I}$ is $x_1$. An action $a\in\mathbb{R}_+$ is the order quantity placed in
the current period, which enters the pipeline in position $z_l$. The sampled instances of
Section~\ref{sec: Numerical Studies} are the continuous-state instances, so states and order
quantities are real-valued.

Combining the state space (\ref{eq: PIC state space}) with the action space
(\ref{eq: PIC action space}) gives the admissible set
\begin{align}
\mathcal{F}^{\mathrm{PIC}}=\Big\{(s,a):\ \textstyle\sum_{j=1}^{l}z_j+\sum_{i=1}^{m}x_i\le G,\quad
a+\sum_{j=1}^{l}z_j+\sum_{i=2}^{m}x_i\le G,\quad 0\le a\le\bar{a}\Big\},
\label{eq: PIC F}
\end{align}
the second inequality omitting $x_1$ because that stock leaves the system before the order
arrives. It is convenient to write the largest admissible order at $s$ as
\begin{align}
a_{\max}(s)\;=\;\max\Big\{G-\textstyle\sum_{j=1}^{l}z_j-\sum_{i=2}^{m}x_i,\ 0\Big\},
\label{eq: PIC amax}
\end{align}
so that the admissible action set at an admissible state is the interval $[0,a_{\max}(s)]$.

\subsubsection{Primitives entering the proposal.}
Demand is negative binomial with parameters $(\iota,q)=(11,\,0.6145)$, truncated at $\bar{D}=20$
with the residual tail mass carried by the atom at $\bar{D}$; the resulting pmf is denoted $p_D$
and has $\mu_D=\mathbb{E}[D]=6.898$ and $\sigma_D^2=\mathrm{Var}(D)=11.167$. The state truncation
level of (\ref{eq: PIC state space}) is set to the conservative value $\bar{a}=G=l\bar{D}$, which
gives $G=100$ at $(l,m)=(5,6)$ and $G=120$ at $(l,m)=(6,10)$.

\subsubsection{The proposal $\mathbb{Q}^{\mathrm{PIC}}$.}
The state proposal is a two-component mixture, drawn coordinatewise conditional on a single
mixture indicator per state,
\begin{align}
\mathbb{Q}^{\mathrm{PIC}}_{\mathcal{S}}
=0.8\bigotimes_{i=1}^{\mathcal{I}}\mathbb{Q}^{\mathrm{sh}}
+0.2\bigotimes_{i=1}^{\mathcal{I}}\mathbb{Q}^{\mathrm{dis}},
\qquad
\mathbb{Q}^{\mathrm{sh}}=\mathrm{normal}(\mu_D,2^2)\big|_{[0,\bar{D}]},
\qquad
\mathbb{Q}^{\mathrm{dis}}=\mathrm{uniform}[0,\bar{D}],
\label{eq: PIC state proposal}
\end{align}
where $\cdot|_{[0,\bar{D}]}$ denotes clipping to $[0,\bar{D}]$. The demand-centered component
concentrates the sample where the system operates; the dispersed component retains full support
on the truncated state space. Drawing the indicator once per state rather than once per
coordinate keeps the two components recognizable as operating regimes: a state is either
demand-like in every bucket or dispersed in every bucket.

\begin{algorithm}[H]
\caption{One candidate draw from the PIC proposal $\mathbb{Q}^{\mathrm{PIC}}$.}
\label{alg: PIC proposal}
\textbf{Input}: dimensions $(l,m)$ and $\mathcal{I}=l+m$; demand mean $\mu_D$ and truncation
$\bar{D}$; state truncation $G$ and order limit $\bar{a}=G$.\\
\textbf{Output}: a candidate pair $(s,a)\sim\mathbb{Q}^{\mathrm{PIC}}$.\\
\emph{State.} Draw $\varrho\sim\mathrm{uniform}(0,1)$.\\
\uIf{$\varrho\le 0.8$}{
    Draw $\tilde{s}_i\sim\mathrm{normal}(\mu_D,2^2)$ independently for $i=1,\dots,\mathcal{I}$.\\
}
\Else{
    Draw $\tilde{s}_i\sim\mathrm{uniform}[0,\bar{D}]$ independently for $i=1,\dots,\mathcal{I}$.\\
}
Set $s\leftarrow\big(\mathrm{clip}(\tilde{s}_1,0,\bar{D}),\dots,
\mathrm{clip}(\tilde{s}_{\mathcal{I}},0,\bar{D})\big)$, read as $(z_l,\dots,z_1,x_m,\dots,x_1)$.\\
\emph{Action.} Draw $\varrho'\sim\mathrm{uniform}(0,1)$, independently of $\varrho$.\\
\uIf{$\varrho'\le 0.8$}{
    Draw $\tau\leftarrow\mu_D(l+1)\,\mathrm{uniform}(0.8,1.2)$ and
    $\epsilon\sim\mathrm{normal}\big(0,(\bar{a}/10)^2\big)$.\\
    Set $a\leftarrow\mathrm{clip}\big(\tau-\sum_{i=1}^{\mathcal{I}}s_i+\epsilon,\,0,\,\bar{a}\big)$.\\
}
\Else{
    Draw $a\sim\mathrm{uniform}[0,\bar{a}]$.\\
}
\Return{$(s,a)$}\\
\end{algorithm}

The action proposal is a mixture with the same weights, and it is genuinely conditional on $s$:
its shaped component targets a jittered order-up-to level covering expected demand over the lead
time plus one review period,
\begin{align}
\mathbb{Q}^{\mathrm{PIC}}_{\mathcal{A}}(\cdot\mid s)
=0.8\,\mathcal{U}\Big(\mathrm{clip}\big(\tau-\textstyle\sum_{i}s_i+\epsilon,\,0,\,\bar{a}\big)\Big)
+0.2\,\mathrm{uniform}[0,\bar{a}],
\qquad
\begin{aligned}
\tau&=\mu_D(l+1)\,\mathrm{uniform}(0.8,1.2),\\
\epsilon&\sim\mathrm{normal}\big(0,(\bar{a}/10)^2\big),
\end{aligned}
\label{eq: PIC action proposal}
\end{align}
where $\mathcal{U}(\cdot)$ denotes the law of its argument, $\sum_i s_i$ is the inventory position
of $s$, and $\mathrm{clip}(\cdot,0,\bar{a})$ truncates its argument to $[0,\bar{a}]$. The
perturbation $\epsilon$ prevents the shaped component from degenerating to a single point mass at
each state. Algorithm~\ref{alg: PIC proposal} collects the two proposals into a single draw. The
sampling distribution is
$\psi^{\mathrm{PIC}}=\mathbb{Q}^{\mathrm{PIC}}(\cdot\mid\mathcal{F}^{\mathrm{PIC}})$, obtained by
supplying Algorithm~\ref{alg: PIC proposal} and $\mathcal{F}^{\mathrm{PIC}}$ of
(\ref{eq: PIC F}) to Algorithm~\ref{alg: X sampling}.

\subsubsection{Expectations in the continuous-demand instances.} \label{oa: continuous demand settings}
The instances of Section~\ref{oa: PIC continuous demand} replace $p_D$ by a Gamma law clipped at $\bar{D}$ and moment-matched to the truncated pmf it replaces, with shape $\mu_D^2/\sigma_D^2=4.261$ and scale $\sigma_D^2/\mu_D=1.619$. The successor expectation is approximated by sample average approximation with $n_{\mathrm{SAA}}$ i.i.d.\ demand draws per state--action pair.

\subsection{Queueing Control}
\label{oa: sampling QC}

\subsubsection{States, actions, and the admissible set.}
A state $s=(s_1,\dots,s_J)\in\mathbb{Z}_+^{J}$ records the number of customers of each class
currently in service, so $\sum_{j}s_j$ is the system occupancy and $\kappa$ is both the number of
servers and the occupancy cap. An action $a=(a_1,\dots,a_J)\in\{0,1\}^{J}$ is an admission rule,
$a_j=1$ meaning that a class-$j$ arrival would be admitted. Since at most one arrival occurs per
decision epoch, (\ref{eq: QC action space}) requires only that admitting any single class leave
the system within capacity, and the admissible set is
\begin{align}
\mathcal{F}^{\mathrm{QC}}=\Big\{(s,a):\ s\in\mathbb{Z}_+^{J},\ a\in\{0,1\}^{J},\
\textstyle\sum_{j=1}^{J}s_j+\max_{j}a_j\le\kappa\Big\}.
\label{eq: QC F}
\end{align}
The state constraint (\ref{eq: QC state space}) is implied, since $\max_j a_j\ge 0$. At an
admissible state with $\sum_j s_j<\kappa$ every action in $\{0,1\}^{J}$ is admissible. At $\sum_j s_j=\kappa$ only the reject-all action is.

\subsubsection{Primitives entering the proposal.}
Class $j$ carries profit $p_j=j$ and service rate $\omega'_j=1+(j-1)/(J-1)$, so rates increase
linearly from $1$ to $2$ across classes. Arrival rates are obtained by splitting the offered load
across classes with a symmetric Dirichlet draw, $(d_1,\dots,d_J)\sim\mathrm{Dirichlet}(\mathbf{1}_J)$,
and setting $\lambda'_j=d_j\,\kappa\rho\,\omega'_j$, where $\rho$ is the service load reported
with each instance. Uniformization uses
$\vartheta'=\sum_{j}\lambda'_j+\kappa\max_{j}\omega'_j$ and the continuous-time discount rate
$\zeta=(0.01/0.99)\,\vartheta'$, which yields the discrete-time discount factor
$\vartheta=\vartheta'/(\zeta+\vartheta')=0.99$ used throughout. The Dirichlet draw is made once
per instance on the instance stream, so the rates are fixed before any sampling occurs.

\subsubsection{The proposal $\mathbb{Q}^{\mathrm{QC}}$.}
The state proposal allocates the server budget sequentially across classes: it sets
$\kappa_1=\kappa$ and, for $j=1,\dots,J$, draws $s_j\sim\mathrm{uniform}\{0,\dots,\kappa_j\}$ and
then, for $j<J$, reduces the remaining budget by a further uniform draw,
$\kappa_{j+1}=\kappa_j-\mathrm{uniform}\{0,\dots,s_j\}$. Because the budget is reduced by less
than the amount just allocated, the running total may exceed $\kappa$. The vector is then rescaled
componentwise to $\lfloor\kappa s_j/\sum_{i}s_i\rfloor$, which returns it to the capacity simplex.
Its coordinates are finally reversed with probability $1/2$, removing the ordering bias of the
sequential allocation, under which the classes drawn first receive systematically larger
allocations. The action proposal does not depend on the state,
\begin{align}
\mathbb{Q}^{\mathrm{QC}}_{\mathcal{A}}(\cdot\mid s)
=\bigotimes_{j=1}^{J}\mathrm{Bernoulli}(1/2)
=\mathrm{uniform}\{0,1\}^{J}.
\label{eq: QC action proposal}
\end{align}

Algorithm~\ref{alg: QC proposal} states the draw, and
$\psi^{\mathrm{QC}}=\mathbb{Q}^{\mathrm{QC}}(\cdot\mid\mathcal{F}^{\mathrm{QC}})$ follows from
Algorithm~\ref{alg: X sampling} with $\mathcal{F}^{\mathrm{QC}}$ of (\ref{eq: QC F}).

\begin{algorithm}[H]
\caption{One candidate draw from the QC proposal $\mathbb{Q}^{\mathrm{QC}}$.}
\label{alg: QC proposal}
\textbf{Input}: number of classes $J$, capacity $\kappa$.\\
\textbf{Output}: a candidate pair $(s,a)\sim\mathbb{Q}^{\mathrm{QC}}$.\\
\emph{State.} Set the residual budget $\kappa_1\leftarrow\kappa$.\\
\For{$j=1,\dots,J$}{
    Draw $s_j\sim\mathrm{uniform}\{0,1,\dots,\kappa_j\}$.\\
    \lIf{$j<J$}{$\kappa_{j+1}\leftarrow\kappa_j-\mathrm{uniform}\{0,1,\dots,s_j\}$}
}
\lIf{$\sum_{j}s_j>\kappa$}{$s_j\leftarrow\lfloor\kappa s_j/\sum_{i}s_i\rfloor$ for all $j$}
With probability $1/2$, reverse the coordinates: $s\leftarrow(s_J,\dots,s_1)$.\\
\emph{Action.} Draw $a_j\sim\mathrm{Bernoulli}(1/2)$ independently for $j=1,\dots,J$.\\
\Return{$(s,a)$}\\
\end{algorithm}

\paragraph{Where the rejection binds.} The rescaling step guarantees
$\sum_{j}s_j\le\kappa$, so every proposed state is admissible and rejection in
Algorithm~\ref{alg: X sampling} operates only through the joint condition
$\sum_{j}s_j+\max_{j}a_j\le\kappa$. It therefore discards exactly those pairs that would admit an
arrival into a full system, which is the capacity face of $\mathcal{F}^{\mathrm{QC}}$. 

\subsection{Construction of the Curated Sample \texorpdfstring{$\mathcal{Y}$}{Y}}
\label{oa: Y}

The sample $\mathcal{Y}$ is constructed once, before any realization of $\mathcal{X}^{0}$ and on
a random stream of its own, in accordance with the pre-commitment convention of
Section~\ref{sec: Constraint Sampling}. Corollary~\ref{cor: violations under Z} therefore applies
unchanged, and it applies with $M=|\mathcal{X}^{k}|$ rather than $|\mathcal{Z}^{k}|$, since
deterministic constraints cannot serve as randomly drawn support constraints. Both applications
follow the same template: $\mathcal{Y}_{\mathcal{S}}$ combines a chain-walk component with a
dispersed component, and $\mathcal{Y}_{\mathcal{A}}(s)$ is attached state by state. The
application-specific ingredients are collected in Table~\ref{tab: walk ingredients}.

\subsubsection{The chain-walk component.}
A fraction $\chi$ of the target $n_{\mathcal{S}}$ states is produced by
Algorithm~\ref{alg: chain walk}, which runs $n_{\mathrm{par}}$ parallel trajectories from a
nominal state $s^{0}$ under a randomized family of predefined policies
$\{\pi_\varsigma:\varsigma\in\mathcal{P}\}$, restarting every $t_{\mathrm{r}}$ steps with fresh
policy parameters and taking an exploratory action with probability $\varepsilon_{\mathrm{ex}}$
at each step. Transitions are drawn from $p(\cdot\mid s,a)$ of the application instance: one
demand draw per walker in PIC and one draw from the uniformized categorical law in QC.

The routine $\textsc{Subsample}(\mathcal{H},n)$ discards inadmissible states from the visited
multiset $\mathcal{H}$, applies a uniform random permutation to the remainder, removes duplicates
keeping first occurrences, and returns the first $n$ distinct states, or $\emptyset$ if fewer than
$n$ are available. The permutation is what makes first-occurrence order a draw weighted by visit
frequency, so that $\mathcal{Y}_{\mathcal{S}}$ reflects where the predefined policies spend their
time. Without it the routine would return a neighborhood of $s^{0}$, since every trajectory
restarts there.

\begin{algorithm}[H]
\caption{Chain-walk construction of the curated states.}
\label{alg: chain walk}
\textbf{Input}: parallel trajectories $n_{\mathrm{par}}$, restart period $t_{\mathrm{r}}$,
exploration probability $\varepsilon_{\mathrm{ex}}$, horizon $t_{\max}$, target
$n_{\mathrm{w}}=\mathrm{round}(\chi\,n_{\mathcal{S}})$, policy family
$\{\pi_\varsigma:\varsigma\in\mathcal{P}\}$, nominal state $s^{0}$, transition probability $p$.\\
\textbf{Output}: $n_{\mathrm{w}}$ distinct admissible states, or all distinct admissible states
visited if fewer are available.\\
Set $s^{b}=s^{0}$ and draw $\varsigma^{b}$ uniformly from $\mathcal{P}$, for
$b=1,\dots,n_{\mathrm{par}}$. Let $\mathcal{H}=\{s^{b}\}_{b\le n_{\mathrm{par}}}$ and $t=1$.\\
\While{$t\le t_{\max}$}{
    For each $b$, let $a^{b}=\pi_{\varsigma^{b}}(s^{b})$ and, independently with probability
    $\varepsilon_{\mathrm{ex}}$, replace $a^{b}$ by a uniform draw from the admissible actions at
    $s^{b}$.\\
    For each $b$, draw $s^{b}\sim p(\cdot\mid s^{b},a^{b})$ and append $s^{b}$ to $\mathcal{H}$.\\
    \If{$t$ is a multiple of $t_{\mathrm{r}}$}{
        Restart: set $s^{b}=s^{0}$ and redraw $\varsigma^{b}$ from $\mathcal{P}$, for all $b$.\\
    }
    \If{$t\ge t_{\mathrm{r}}$, $t$ is a multiple of $50$, and
        $\textsc{Subsample}(\mathcal{H},n_{\mathrm{w}})\neq\emptyset$}{
        \Return{$\textsc{Subsample}(\mathcal{H},n_{\mathrm{w}})$}
    }
    $t\leftarrow t+1$.\\
}
\Return{$\textsc{Subsample}(\mathcal{H},n_{\mathrm{w}})$ if nonempty, and the distinct admissible
states of $\mathcal{H}$ otherwise}\\
\end{algorithm}

\begin{table}[H]
\centering
\caption{\normalfont Application-specific ingredients of Algorithm~\ref{alg: chain walk}.}
\label{tab: walk ingredients}
\small
\begin{tabular}{|l|p{5.4cm}|p{5.4cm}|}
\hline
Ingredient & PIC & QC \\ \hline
Nominal state $s^{0}$
& Empty pipeline, $\mu_D$ units in every shelf bucket:
  $s^{0}=(0,\dots,0,\mu_D,\dots,\mu_D)$
& Empty system: $s^{0}=0$ \\ \hline
Policy parameter $\varsigma$
& $(\mathrm{kind},\overline{y},\underline{y})$ with $\mathrm{kind}\sim
  \mathrm{uniform}\{0,1,2\}$,
  $\overline{y}=\mu_D(l+1)\,\mathrm{uniform}(0.8,1.6)$,
  $\underline{y}=\mu_D(l+1)\,\mathrm{uniform}(0.2,0.8)$
& $(\underline{j},\bar{\kappa})$ with
  $\underline{j}\sim\mathrm{uniform}\{1,\dots,J+1\}$,
  $\bar{\kappa}\sim\mathrm{uniform}(0.6,1.15)$ \\ \hline
Policy $\pi_\varsigma(s)$
& $0$ if $\mathrm{kind}=0$; $\mathrm{clip}(\overline{y}-\sum_i s_i,0,a_{\max}(s))$ if
  $\mathrm{kind}=1$; the same quantity if $\mathrm{kind}=2$ and
  $\sum_i s_i\le\underline{y}$, and $0$ otherwise
& $\pi_\varsigma(s)_j=\mathbf{1}\{j\ge\underline{j}\}\,
  \mathbf{1}\{\sum_i s_i<\bar{\kappa}\kappa\}\,
  \mathbf{1}\{\sum_i s_i<\kappa\}$ \\ \hline
Exploratory action law
& $\mathrm{uniform}[0,a_{\max}(s)]$
& $\mathrm{uniform}\{0,1\}^{J}$ if $\sum_i s_i<\kappa$, and $a^{\mathrm{none}}$ otherwise \\ \hline
Admissibility filter in \textsc{Subsample}
& $\sum_{j}z_j+\sum_{i}x_i\le G$
& $\sum_{j}s_j\le\kappa$ \\ \hline
\end{tabular}
\end{table}

\subsubsection{Dispersed component and reconciliation.}
The remaining $n_{\mathcal{S}}-n_{\mathrm{w}}$ states are drawn from the state proposal
$\mathbb{Q}_{\mathcal{S}}$ of Section~\ref{oa: sampling PIC} or \ref{oa: sampling QC}, retained if
admissible, and deduplicated.
The two pools may overlap, so the merged set is deduplicated and any shortfall is made up by
further draws from $\mathbb{Q}_{\mathcal{S}}$, in batches sized by a running estimate of the
acceptance rate. Algorithm~\ref{alg: Y states} states the procedure.

\begin{algorithm}[H]
\caption{Assembling the curated states $\mathcal{Y}_{\mathcal{S}}$.}
\label{alg: Y states}
\textbf{Input}: state proposal $\mathbb{Q}_{\mathcal{S}}$, admissible state set
$\mathcal{S}_{\mathcal{F}}$, target $n_{\mathcal{S}}$, chain-walk fraction $\chi$, the inputs of
Algorithm~\ref{alg: chain walk}.\\
\textbf{Output}: $\mathcal{Y}_{\mathcal{S}}$, a set of $n_{\mathcal{S}}$ distinct admissible
states.\\
Set $n_{\mathrm{w}}\leftarrow\mathrm{round}(\chi n_{\mathcal{S}})$ and
$n_{\mathrm{d}}\leftarrow n_{\mathcal{S}}-n_{\mathrm{w}}$.\\
\emph{Chain-walk pool.} Obtain $\mathcal{W}$ from Algorithm~\ref{alg: chain walk} with target
$n_{\mathrm{w}}$.\\
\emph{Dispersed pool.} Set $\mathcal{D}\leftarrow\emptyset$.\\
\For{$r=1,\dots,50$}{
    Draw $\max\{4n_{\mathrm{d}},4096\}$ states from $\mathbb{Q}_{\mathcal{S}}$ and let
    $\mathcal{D}$ be the distinct states of
    $(\mathcal{D}\cup\{\text{draws}\})\cap\mathcal{S}_{\mathcal{F}}$.\\
    \lIf{$|\mathcal{D}|\ge n_{\mathrm{d}}$}{\textbf{break}}
}
\emph{Merge.} Let $\mathcal{Y}_{\mathcal{S}}$ be the distinct states of
$\mathcal{W}\cup\{\text{first }n_{\mathrm{d}}\text{ states of }\mathcal{D}\}$ and set
$\hat{r}\leftarrow 1$.\\
\emph{Top-up.}\\
\For{$r=1,\dots,200$}{
    \lIf{$|\mathcal{Y}_{\mathcal{S}}|\ge n_{\mathcal{S}}$}{\textbf{break}}
    Set $q\leftarrow n_{\mathcal{S}}-|\mathcal{Y}_{\mathcal{S}}|$ and
    $B\leftarrow\min\{\max\{4q/\max\{\hat{r},10^{-4}\},4096\},4\times10^{6}\}$.\\
    Draw $B$ states from $\mathbb{Q}_{\mathcal{S}}$ and let $\mathcal{K}$ be the admissible ones.\\
    Update the acceptance estimate
    $\hat{r}\leftarrow 0.5\,\hat{r}+0.5\max\{|\mathcal{K}|/B,\,10^{-6}\}$.\\
    Let $\mathcal{Y}_{\mathcal{S}}$ be the distinct states of
    $\mathcal{Y}_{\mathcal{S}}\cup\mathcal{K}$.\\
}
\Return{the first $n_{\mathcal{S}}$ states of $\mathcal{Y}_{\mathcal{S}}$}\\
\end{algorithm}

\paragraph{The dispersed component is not a draw from $\psi$.} States in the dispersed pool are
drawn from $\mathbb{Q}_{\mathcal{S}}$ and screened for state admissibility alone, whereas the
state marginal of $\psi$ is $\mathbb{Q}_{\mathcal{S}}$ conditioned on the \emph{joint} event
$\mathcal{F}$, which reweights states by the acceptance probability of their action sets.

\subsubsection{Action sets.}
Algorithm~\ref{alg: Y actions} attaches $\mathcal{Y}_{\mathcal{A}}(s)$ to each retained state.
For PIC the candidate set is an evenly spaced grid over $[0,a_{\max}(s)]$ containing both
endpoints. For QC it is a structured family of priority-threshold rules, padded with uniform
draws and then screened for admissibility at $s$. The QC candidate block is drawn once and shared
across states, so that any two states with the same residual capacity receive the same action set.

\begin{algorithm}[htbp!]
\caption{Attaching the candidate actions $\mathcal{Y}_{\mathcal{A}}(s)$.}
\label{alg: Y actions}
\textbf{Input}: curated states $\mathcal{Y}_{\mathcal{S}}$, actions per state $n_{\mathcal{A}}$.\\
\textbf{Output}: $\mathcal{Y}_{\mathcal{A}}(s)$ for every $s\in\mathcal{Y}_{\mathcal{S}}$, and
$\mathcal{Y}=\{(s,a):s\in\mathcal{Y}_{\mathcal{S}},\,a\in\mathcal{Y}_{\mathcal{A}}(s)\}$.\\
\emph{PIC.}\\
\ForEach{$s\in\mathcal{Y}_{\mathcal{S}}$}{
    Compute $a_{\max}(s)$ from (\ref{eq: PIC amax}).\\
    Set $\mathcal{Y}_{\mathcal{A}}(s)\leftarrow\big\{\tfrac{k}{n_{\mathcal{A}}-1}\,a_{\max}(s):
    k=0,1,\dots,n_{\mathcal{A}}-1\big\}$, retaining distinct values only.\\
}
\emph{QC.} For $\underline{j}=1,\dots,J+1$, let $a(\underline{j})\in\{0,1\}^{J}$ have
$a(\underline{j})_i=\mathbf{1}\{i\ge\underline{j}\}$, so that $a(1)=a^{\mathrm{all}}$ and
$a(J+1)=a^{\mathrm{none}}$, and set
$\mathcal{A}_{\mathrm{str}}=\{a(\underline{j}):\underline{j}=1,\dots,J+1\}$.\\
\uIf{$n_{\mathcal{A}}<|\mathcal{A}_{\mathrm{str}}|$}{
    Let $\mathcal{A}_{\mathrm{cand}}$ retain the $n_{\mathcal{A}}$ most selective thresholds,
    i.e.\ those with the largest $\underline{j}$.\\
}
\Else{
    Draw $n_{\mathcal{A}}-|\mathcal{A}_{\mathrm{str}}|$ actions independently from
    $\mathrm{Bernoulli}(1/2)^{\otimes J}$ and let $\mathcal{A}_{\mathrm{cand}}$ be
    $\mathcal{A}_{\mathrm{str}}$ together with these fills.\\
}
\ForEach{$s\in\mathcal{Y}_{\mathcal{S}}$}{
    Set $\mathcal{Y}_{\mathcal{A}}(s)\leftarrow\{a\in\mathcal{A}_{\mathrm{cand}}:
    \max_{i}a_i\le\kappa-\sum_{i}s_i\}$.\\
}
\Return{$\mathcal{Y}$}\\
\end{algorithm}

\subsection{Parameters}

\begin{table}[htbp!]
\centering
\caption{\normalfont Sampling parameters used in the sampled experiments of
Section~\ref{sec: Numerical Studies}.}
\label{tab: sampling parameters}
\scriptsize
\begin{tabular}{|l|cc|}
\hline
Quantity & PIC & QC \\ \hline
Sampled constraints per iteration, $M=|\mathcal{X}^{k}|$ & 50{,}000 & 50{,}000 \\
Shaped / dispersed weights in $\mathbb{Q}$ & $0.8/0.2$ & --- \\
Curated states, $n_{\mathcal{S}}$ & 10{,}000 & 10{,}000 \\
Candidate actions per curated state, $n_{\mathcal{A}}$ & 20 & 20 \\
Chain-walk fraction, $\chi$ & 0.7 & 0.7 \\
Parallel trajectories, $n_{\mathrm{par}}$ & 512 & 512 \\
Restart period, $t_{\mathrm{r}}$ & 200 & 200 \\
Exploration probability, $\varepsilon_{\mathrm{ex}}$ & 0.1 & 0.1 \\
Horizon, $t_{\max}$ & 4{,}000 & 4{,}000 \\
SAA draws, $n_{\mathrm{SAA}}$ (continuous-demand instances) & 500 & --- \\
Random seed & 0 & 0 \\
\hline
\end{tabular}
\end{table}

\section{Instance-Level Numerical Results}
\label{oa: full results}

Tables \ref{tab: PIC benchmark}--\ref{tab: QC Benchmark (sampling)} report the per-instance
results summarized in Table \ref{tab: summary} of the main text.

\begin{table}[H]\centering
\caption{\normalfont Conservative relative error bound $\frac{1}{(1-\gamma) \nu^\top\tilde{V}_k}\parallel \mathcal{T}\tilde{V}_k-\tilde{V}_k\parallel_\infty$ for all solution methods executed over 20 iterations in the PIC problem. The parameters $h$ and $\eta$ are fixed to 1 and 20, respectively. For the random approaches, five BFs are generated per iteration. } \label{tab: PIC benchmark}
\scriptsize

\begin{tabular}{|ccc|ccccc|}
\hline
\multicolumn{3}{|c|}{Instance parameters}                                             & \multicolumn{5}{c|}{Conservative relative error bound}                                                                                                         \\ \hline
\multicolumn{1}{|c|}{($l$, $m$)}                 & ($f$, $\theta$)                  & $\gamma$    & BOBF              & SBOBF(5) & Fixed    & \begin{tabular}[c]{@{}c@{}}Random\\ (ReLU)\end{tabular} & \begin{tabular}[c]{@{}c@{}}Random\\ (Fourier)\end{tabular} \\ \hline
\multicolumn{1}{|c|}{\multirow{9}{*}{(2,2)}} & \multirow{2}{*}{(10, 5)}   & \multicolumn{1}{c|}{0.95}  & \textbf{1.96e-07} & \textbf{3.12e-05} & 6.13e+00 & 5.48e+00                                                & \multicolumn{1}{c|}{1.44e+01}                                                   \\
\multicolumn{1}{|c|}{}                       &                          & \multicolumn{1}{c|}{0.99}  & \textbf{5.72e-07} & 5\textbf{.65e-05} & 6.71e+00 & 5.90e+00                                                & \multicolumn{1}{c|}{1.43e+01}                                                   \\
\multicolumn{1}{|c|}{}                       & \multirow{2}{*}{(10, 20)}  & \multicolumn{1}{c|}{0.95}  & \textbf{8.66e-08} & \textbf{5.24e-06} & 3.30e+00 & 1.33e+01                                                & \multicolumn{1}{c|}{5.02e+01}                                                   \\
\multicolumn{1}{|c|}{}                       &                          & \multicolumn{1}{c|}{0.99}  & \textbf{1.93e-07} & \textbf{2.13e-05} & 9.68e+00 & 1.51e+01                                                & \multicolumn{1}{c|}{4.99e+01}                                                   \\
\multicolumn{1}{|c|}{}                       & \multirow{2}{*}{(100, 5)}  & \multicolumn{1}{c|}{0.95}  & \textbf{1.55e-12} & \textbf{4.85e-08} & 9.68e-01 & 2.41e+00                                                & \multicolumn{1}{c|}{1.48e+01}                                                   \\
\multicolumn{1}{|c|}{}                       &                          & \multicolumn{1}{c|}{0.99}  & \textbf{5.19e-10} & \textbf{2.31e-08} & 1.42e+00 & 2.34e+00                                                & \multicolumn{1}{c|}{1.46e+01}                                                   \\
\multicolumn{1}{|c|}{}                       & \multirow{2}{*}{(100, 20)} & \multicolumn{1}{c|}{0.95}  & \textbf{5.11e-14} & \textbf{1.22e-09} & 1.71e+00 & 4.70e+00                                                & \multicolumn{1}{c|}{4.91e+01}                                                   \\
\multicolumn{1}{|c|}{}                       &                          & \multicolumn{1}{c|}{0.99}  & \textbf{9.22e-13} & \textbf{3.10e-09} & 2.58e+00 & 4.95e+00                                                & \multicolumn{1}{c|}{4.85e+01}                                                   \\ \cline{2-8} 
\multicolumn{1}{|c|}{}                       & \multicolumn{2}{c|}{Sub-group average}                & \textbf{1.31e-07} & \textbf{1.43e-05} & 4.06e+00 & 6.78e+00                                                & \multicolumn{1}{c|}{3.20e+01}                                                   \\ \hline
\multicolumn{1}{|c|}{\multirow{9}{*}{(1,4)}} & \multirow{2}{*}{(10, 5)}   & \multicolumn{1}{c|}{0.95}  & \textbf{7.46e-07} & \textbf{9.11e-05} & 2.05e+00 & 6.63e+00                                                & \multicolumn{1}{c|}{1.44e+01}                                                   \\
\multicolumn{1}{|c|}{}                       &                          & \multicolumn{1}{c|}{0.99}  & \textbf{8.04e-07} & 1.68e-04 & 2.24e+00 & 6.21e+00                                                & \multicolumn{1}{c|}{1.43e+01}                                                   \\
\multicolumn{1}{|c|}{}                       & \multirow{2}{*}{(10, 20)}  & \multicolumn{1}{c|}{0.95}  & \textbf{4.81e-08} & \textbf{7.00e-05} & 4.03e+00 & 1.02e+01                                                & \multicolumn{1}{c|}{4.35e+01}                                                   \\
\multicolumn{1}{|c|}{}                       &                          & \multicolumn{1}{c|}{0.99}  & \textbf{1.70e-07} & 3.08e-04 & 4.78e+00 & 1.18e+01                                                & \multicolumn{1}{c|}{4.33e+01}                                                   \\
\multicolumn{1}{|c|}{}                       & \multirow{2}{*}{(100, 5)}  & \multicolumn{1}{c|}{0.95}  & \textbf{1.41e-07} & \textbf{6.85e-05} & 1.70e+00 & 2.44e+00                                                & \multicolumn{1}{c|}{1.44e+01}                                                   \\
\multicolumn{1}{|c|}{}                       &                          & \multicolumn{1}{c|}{0.99}  & \textbf{8.45e-07} & \textbf{1.87e-05} & 3.02e+00 & 2.94e+00                                                & \multicolumn{1}{c|}{1.43e+01}                                                   \\
\multicolumn{1}{|c|}{}                       & \multirow{2}{*}{(100, 20)} & \multicolumn{1}{c|}{0.95}  & \textbf{6.85e-08} & 4.54e-04 & 2.63e+00 & 3.80e+00                                                & \multicolumn{1}{c|}{4.35e+01}                                                   \\
\multicolumn{1}{|c|}{}                       &                          & \multicolumn{1}{c|}{0.99}  & \textbf{4.45e-08} & 2.06e-04 & 4.23e+00 & 4.88e+00                                                & \multicolumn{1}{c|}{4.33e+01}                                                   \\ \cline{2-8} 
\multicolumn{1}{|c|}{}                       & \multicolumn{2}{c|}{Sub-group average}                & \textbf{3.58e-07} & 1.73e-04 & 3.09e+00 & 6.11e+00                                                & \multicolumn{1}{c|}{2.89e+01}                                                   \\ \hline
\multicolumn{3}{|c|}{Overall average}                                                                  & \textbf{2.45e-07} & \textbf{9.36e-05} & 3.57e+00 & 6.44e+00                                                & \multicolumn{1}{c|}{3.04e+01}                                                   \\ \hline
                                                                 
\end{tabular}
\end{table}

\begin{table}[H]
\centering
\caption{\normalfont Empirical relative error bound $\frac{1}{(1-\gamma) \nu^\top\tilde{V}_k}\parallel \mathcal{L}_{\mathcal{Y}}\tilde{V}_k - \tilde{V}_k \parallel_\infty$ for all solution methods executed over 30 iterations in the continuous PIC problem. The parameters $h$ and $\eta$ are fixed to 1 and 20, respectively. For the random approaches, 5 BFs are generated per iteration.}
\label{tab: PIC benchmark (sampling)}
\scriptsize
\begin{tabular}{|ccc|ccccc|}
\hline
\multicolumn{3}{|c|}{Instance parameters} & \multicolumn{5}{c|}{Empirical relative error bound} \\ \hline
\multicolumn{1}{|c|}{($l,m$)} & ($f, \theta$) & $\gamma$ & HBOBF & SHBOBF(5) & Fixed & \begin{tabular}[c]{@{}c@{}}Random\\ (ReLU)\end{tabular} & \begin{tabular}[c]{@{}c@{}}Random\\ (Fourier)\end{tabular} \\ \hline
\multicolumn{1}{|c|}{\multirow{9}{*}{(5,6)}} & \multirow{2}{*}{(10,5)} & 0.95 & \textbf{5.20e-05} & \textbf{5.20e-05} & 1.82e+01 & 3.12e+01 & 5.42e+01 \\
\multicolumn{1}{|c|}{} &  & 0.99 & \textbf{7.40e-05} & \textbf{7.40e-05} & 2.12e+01 & 3.52e+01 & 5.42e+01 \\
\multicolumn{1}{|c|}{} & \multirow{2}{*}{(10,20)} & 0.95 & \textbf{4.76e-05} & \textbf{4.76e-05} & 3.27e+01 & 1.08e+02 & 1.91e+02 \\
\multicolumn{1}{|c|}{} &  & 0.99 & \textbf{7.23e-05} & \textbf{7.23e-05} & 3.96e+01 & 1.25e+02 & 1.91e+02 \\
\multicolumn{1}{|c|}{} & \multirow{2}{*}{(100,5)} & 0.95 & \textbf{8.46e-15} & \textbf{5.42e-15} & 2.97e+01 & 1.83e+01 & 5.42e+01 \\
\multicolumn{1}{|c|}{} &  & 0.99 & \textbf{5.43e-14} & \textbf{1.32e-14} & 2.55e+01 & 1.74e+01 & 5.42e+01 \\
\multicolumn{1}{|c|}{} & \multirow{2}{*}{(100,20)} & 0.95 & \textbf{3.97e-14} & \textbf{9.92e-15} & 3.57e+01 & 5.20e+01 & 1.91e+02 \\
\multicolumn{1}{|c|}{} &  & 0.99 & \textbf{1.11e-12} & \textbf{1.03e-11} & 4.23e+01 & 5.72e+01 & 1.91e+02 \\ \cline{2-8}
\multicolumn{1}{|c|}{} & \multicolumn{2}{c|}{Sub-group average} & \textbf{3.07e-05} & \textbf{3.07e-05} & 3.06e+01 & 5.56e+01 & 1.23e+02 \\ \hline
\multicolumn{1}{|c|}{\multirow{9}{*}{(6,10)}} & \multirow{2}{*}{(10,5)} & 0.95 & \textbf{7.91e-06} & \textbf{7.91e-06} & 3.11e+01 & 3.14e+01 & 4.44e+01 \\
\multicolumn{1}{|c|}{} &  & 0.99 & \textbf{7.24e-05} & \textbf{7.24e-05} & 2.21e+01 & 3.45e+01 & 4.44e+01 \\
\multicolumn{1}{|c|}{} & \multirow{2}{*}{(10,20)} & 0.95 & \textbf{8.29e-05} & \textbf{8.29e-05} & 2.32e+01 & 1.16e+02 & 1.55e+02 \\
\multicolumn{1}{|c|}{} &  & 0.99 & \textbf{7.31e-08} & \textbf{7.31e-08} & 3.29e+01 & 1.22e+02 & 1.55e+02 \\
\multicolumn{1}{|c|}{} & \multirow{2}{*}{(100,5)} & 0.95 & \textbf{1.60e-14} & \textbf{5.33e-15} & 7.17e+00 & 2.54e+01 & 4.44e+01 \\
\multicolumn{1}{|c|}{} &  & 0.99 & \textbf{7.89e-14} & \textbf{1.32e-14} & 9.11e+00 & 2.09e+01 & 4.44e+01 \\
\multicolumn{1}{|c|}{} & \multirow{2}{*}{(100,20)} & 0.95 & \textbf{2.02e-14} & \textbf{2.02e-14} & 1.30e+01 & 7.47e+01 & 1.55e+02 \\
\multicolumn{1}{|c|}{} &  & 0.99 & \textbf{3.00e-14} & \textbf{2.60e-14} & 1.65e+01 & 6.59e+01 & 1.55e+02 \\ \cline{2-8}
\multicolumn{1}{|c|}{} & \multicolumn{2}{c|}{Sub-group average} & \textbf{2.04e-05} & \textbf{2.04e-05} & 1.94e+01 & 6.12e+01 & 9.99e+01 \\ \hline
\multicolumn{3}{|c|}{Overall average} & \textbf{2.56e-05} & \textbf{2.56e-05} & 2.50e+01 & 5.84e+01 & 1.11e+02 \\ \hline
\end{tabular}
\end{table}

\begin{table}[H]
\centering
\caption{\normalfont Conservative relative error bound $\frac{1}{(1-\gamma) \nu^\top\tilde{V}_k}\parallel \mathcal{T}\tilde{V}_k-\tilde{V}_k\parallel_\infty$ for all solution methods executed over 100 iterations in the QC problem. For the random methods, five BFs are generated per iteration.} \label{tab: QC Benchmark}
\scriptsize

\begin{tabular}{|cc|ccccc|}
\hline
\multicolumn{2}{|l|}{Instance parameters}                                                                     & \multicolumn{5}{c|}{Conservative relative error bound}                                                                                                                             \\ \hline
\multicolumn{1}{|c|}{($J$, $\kappa$)}                   & \multicolumn{1}{c|}{$\rho$}                                     & BOBF             & SBOBF(5) & Fixed  & \begin{tabular}[c]{@{}c@{}}Random \\ (ReLU)\end{tabular} & \begin{tabular}[c]{@{}c@{}}Random \\ (Fourier)\end{tabular} \\ \hline
\multicolumn{1}{|c|}{\multirow{5}{*}{(3, 15)}} & 0.8                                                          & \textbf{1.66e-08} & \textbf{9.07e-08}          & 1.00e+00 & 1.61e-01                                                 & 1.00e+00                                                    \\
\multicolumn{1}{|c|}{}                         & 1                                                            & \textbf{1.10e-08} & \textbf{4.62e-07}          & 6.41e-01 & 2.50e-01                                                 & 6.41e-01                                                    \\
\multicolumn{1}{|c|}{}                         & 3                                                            & \textbf{6.30e-07} & \textbf{5.05e-06} & 8.31e-01 & 2.64e-01                                                 & 7.40e-01                                                    \\
\multicolumn{1}{|c|}{}                         & 5                                                            & \textbf{9.07e-06} & \textbf{8.91e-05} & 1.50e+00 & 1.93e-01                                                 & 9.07e-01                                                    \\
\multicolumn{1}{|c|}{}                         & 10                                                           & \textbf{{6.45e-05}} & \textbf{5.01e-05} & 1.80e+00 & 1.40e-01                                                 & 1.00e+00                                                    \\ \cline{2-7} 
\multicolumn{1}{|c|}{}                         & \begin{tabular}[c]{@{}c@{}}Sub-group \\ average\end{tabular} & \textbf{1.48e-05} & \textbf{2.90e-05}          & 1.15e+00 & 2.02e-01                                                 & 8.58e-01                                                    \\ \hline
\multicolumn{1}{|c|}{\multirow{5}{*}{(4, 25)}} & 0.8                                                          & \textbf{7.97e-07} & \textbf{6.30e-06}          & 1.00e+00 & 1.00e+00                                                 & 1.00e+00                                                    \\
\multicolumn{1}{|c|}{}                         & 1                                                            & \textbf{7.91e-07} & \textbf{8.97e-06}          & 1.00e+00 & 1.00e+00                                                 & 1.00e+00                                                    \\
\multicolumn{1}{|c|}{}                         & 3                                                            & \textbf{7.94e-05} & 2.53e-04          & 1.37e+00 & 1.00e+00                                                 & 1.00e+00                                                    \\
\multicolumn{1}{|c|}{}                         & 5                                                            & {2.07e-03} & \textbf{1.26e-03} & 1.90e+00 & 1.00e+00                                                 & 1.00e+00                                                    \\
\multicolumn{1}{|c|}{}                         & 10                                                           & {7.71e-03} & \textbf{4.11e-03} & 2.36e+00 & 1.00e+00                                                 & 1.00e+00                                                    \\ \cline{2-7} 
\multicolumn{1}{|c|}{}                         & \begin{tabular}[c]{@{}c@{}}Sub-group \\ average\end{tabular} & {1.97e-03} & \textbf{1.13e-03} & 1.53e+00 & 1.00e+00                                                 & 1.00e+00                                                    \\ \hline
\multicolumn{2}{|c|}{Overall average}                                                                                 & {9.94e-04} & \textbf{5.78e-04} & 1.34e+00 & 6.01e-01                                                 & 9.29e-01                                                    \\ \hline
\end{tabular}

\end{table}

\begin{table}[H]
\centering
\caption{\normalfont Empirical relative error bound $\frac{1}{(1-\gamma) \nu^\top\tilde{V}_k}\parallel \mathcal{L}_{\mathcal{Y}}\tilde{V}_k - \tilde{V}_k \parallel_\infty$ for all solution methods executed over 50 iterations in the QC problem. For the random methods, 5 BFs are generated per iteration.}
\label{tab: QC Benchmark (sampling)}
\scriptsize
\begin{tabular}{|cc|ccccc|}
\hline
\multicolumn{2}{|c|}{Instance parameters} & \multicolumn{5}{c|}{Empirical relative error bound} \\ \hline
\multicolumn{1}{|c|}{($J, \kappa$)} & $\rho$ & HBOBF & SHBOBF(5) & Fixed & \begin{tabular}[c]{@{}c@{}}Random\\ (ReLU)\end{tabular} & \begin{tabular}[c]{@{}c@{}}Random\\ (Fourier)\end{tabular} \\ \hline
\multicolumn{1}{|c|}{\multirow{6}{*}{(8, 80)}} & 0.8 & \textbf{6.48e-14} & \textbf{6.48e-14} & 1.00e+00 & 1.00e+00 & 1.00e+00 \\
\multicolumn{1}{|c|}{} & 1 & \textbf{1.12e-13} & \textbf{1.12e-13} & 1.00e+00 & 1.00e+00 & 1.00e+00 \\
\multicolumn{1}{|c|}{} & 3 & \textbf{3.75e-05} & \textbf{7.11e-05} & 8.34e-01 & 7.99e-01 & 1.00e+00 \\
\multicolumn{1}{|c|}{} & 5 & \textbf{3.78e-05} & \textbf{6.85e-07} & 1.76e+00 & 1.00e+00 & 1.00e+00 \\
\multicolumn{1}{|c|}{} & 10 & \textbf{6.26e-05} & \textbf{9.20e-05} & 2.24e+00 & 1.00e+00 & 1.00e+00 \\ \cline{2-7}
\multicolumn{1}{|c|}{} & \begin{tabular}[c]{@{}c@{}}Sub-group\\ average\end{tabular} & \textbf{2.76e-05} & \textbf{3.28e-05} & 1.37e+00 & 9.60e-01 & 1.00e+00 \\ \hline
\multicolumn{1}{|c|}{\multirow{6}{*}{(10, 100)}} & 0.8 & \textbf{1.24e-13} & \textbf{1.24e-13} & 1.00e+00 & 1.00e+00 & 1.00e+00 \\
\multicolumn{1}{|c|}{} & 1 & \textbf{8.30e-14} & \textbf{8.30e-14} & 1.00e+00 & 1.00e+00 & 1.00e+00 \\
\multicolumn{1}{|c|}{} & 3 & \textbf{4.08e-05} & \textbf{9.36e-05} & 7.40e-01 & 7.08e-01 & 1.00e+00 \\
\multicolumn{1}{|c|}{} & 5 & \textbf{2.01e-05} & \textbf{5.21e-05} & 1.51e+00 & 9.35e-01 & 1.00e+00 \\
\multicolumn{1}{|c|}{} & 10 & \textbf{4.02e-05} & \textbf{9.58e-05} & 2.20e+00 & 9.73e-01 & 1.00e+00 \\ \cline{2-7}
\multicolumn{1}{|c|}{} & \begin{tabular}[c]{@{}c@{}}Sub-group\\ average\end{tabular} & \textbf{2.02e-05} & \textbf{4.83e-05} & 1.29e+00 & 9.23e-01 & 1.00e+00 \\ \hline
\multicolumn{2}{|c|}{Overall average} & \textbf{2.39e-05} & \textbf{4.05e-05} & 1.33e+00 & 9.42e-01 & 1.00e+00 \\ \hline
\end{tabular}
\end{table}

\subsection{Perishable Inventory Control under Continuous Demand}
\label{oa: PIC continuous demand}

The sampled PIC experiments of Section~\ref{sec: Numerical Studies} retain the truncated
negative binomial demand law. This subsection reports the same experiments under continuously
distributed demand, in the setting specified in
Section~\ref{oa: continuous demand settings}: demand follows the Gamma law moment-matched to $p_D$
and truncated at $\bar{D}$, states and order quantities are continuous, and the successor
expectation entering $\mathcal{L}_{\mathcal{Y}}$ is evaluated by sample average approximation
with $n_{\mathrm{SAA}}$ draws per state--action pair. The samples $\mathcal{X}^{k}$ and
$\mathcal{Y}$ are constructed exactly as in Sections~\ref{oa: psi} and \ref{oa: Y}, at the
parameter values of Table~\ref{tab: sampling parameters}.

\begin{figure}[H]
\FIGURE{
    \includegraphics[width=0.5\textwidth]{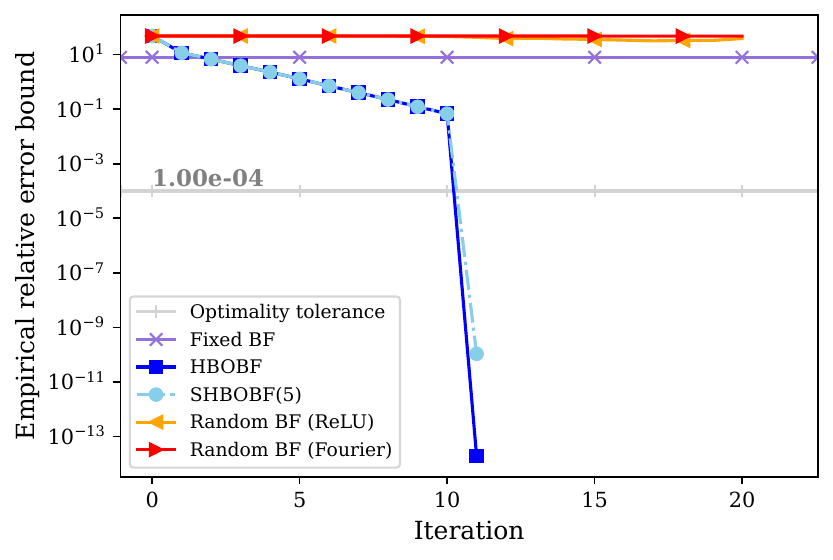}
}{\normalfont Empirical relative error bound $\frac{1}{(1-\gamma) \nu^\top\tilde{V}_k}\parallel \mathcal{L}_{\mathcal{Y}}\tilde{V}_k - \tilde{V}_k \parallel_\infty$ for all solution methods executed over 20 iterations in a PIC instance with Gamma demand and parameter settings $(f, h, \eta, \theta) = (100, 1, 20, 5)$, $(l, m) = (6, 10)$, and $\gamma=0.99$. For the random approaches, three BFs are generated per iteration. \label{fig: PIC continuous demand}}{}
\end{figure}

Figure~\ref{fig: PIC continuous demand} reports the empirical error bound for an instance with
$(l,m)=(6,10)$. Both HBOBF and SHBOBF(5) meet the tolerance and stop within 12 iterations, whereas the Fixed BF and the two random BF expansions remain at error
levels of order 10 after 20 iterations. Table~\ref{tab: PIC continuous demand} reports the
bounds across instance groups, averaged within each group. The ordering of the methods
reproduces that of the second block of Table~\ref{tab: summary}, obtained under discrete
demand, which indicates that neither the continuity of the state space nor the Monte Carlo
evaluation of the successor expectation disturbs the comparison.

\begin{table}[H]
\centering
\caption{\normalfont Empirical relative error bound $\frac{1}{(1-\gamma) \nu^\top\tilde{V}_k}\parallel \mathcal{L}_{\mathcal{Y}}\tilde{V}_k - \tilde{V}_k \parallel_\infty$ for all solution methods executed over 20 iterations in the continuous PIC problem. The parameters $h$ and $\eta$ are fixed to 1 and 20, respectively. For the random approaches, 3 BFs are generated per iteration.}
\label{tab: PIC continuous demand}
\scriptsize
\begin{tabular}{|ccc|ccccc|}
\hline
\multicolumn{3}{|c|}{Instance parameters} & \multicolumn{5}{c|}{Empirical relative error bound} \\ \hline
\multicolumn{1}{|c|}{($l,m$)} & ($f, \theta$) & $\gamma$ & HBOBF & SHBOBF(5) & Fixed & \begin{tabular}[c]{@{}c@{}}Random\\ (ReLU)\end{tabular} & \begin{tabular}[c]{@{}c@{}}Random\\ (Fourier)\end{tabular} \\ \hline
\multicolumn{1}{|c|}{\multirow{9}{*}{(5,6)}} & \multirow{2}{*}{(10,5)} & 0.95 & \textbf{3.47e-07} & \textbf{3.47e-07} & 2.43e+01 & 3.90e+01 & 4.58e+01 \\
\multicolumn{1}{|c|}{} &  & 0.99 & \textbf{4.93e-07} & \textbf{4.93e-07} & 2.46e+01 & 3.86e+01 & 4.58e+01 \\
\multicolumn{1}{|c|}{} & \multirow{2}{*}{(10,20)} & 0.95 & \textbf{1.09e-05} & \textbf{1.09e-05} & 3.15e+01 & 1.37e+02 & 1.63e+02 \\
\multicolumn{1}{|c|}{} &  & 0.99 & \textbf{1.58e-05} & \textbf{1.58e-05} & 3.62e+01 & 1.38e+02 & 1.63e+02 \\
\multicolumn{1}{|c|}{} & \multirow{2}{*}{(100,5)} & 0.95 & \textbf{6.32e-10} & \textbf{6.32e-10} & 2.63e+01 & 3.54e+01 & 4.58e+01 \\
\multicolumn{1}{|c|}{} &  & 0.99 & \textbf{9.56e-10} & \textbf{9.56e-10} & 3.04e+01 & 3.05e+01 & 4.58e+01 \\
\multicolumn{1}{|c|}{} & \multirow{2}{*}{(100,20)} & 0.95 & \textbf{5.80e-10} & \textbf{5.80e-10} & 3.73e+01 & 1.01e+02 & 1.63e+02 \\
\multicolumn{1}{|c|}{} &  & 0.99 & \textbf{9.36e-10} & \textbf{9.34e-10} & 3.75e+01 & 9.00e+01 & 1.63e+02 \\ \cline{2-8}
\multicolumn{1}{|c|}{} & \multicolumn{2}{c|}{Sub-group average} & \textbf{3.44e-06} & \textbf{3.44e-06} & 3.10e+01 & 7.63e+01 & 1.04e+02 \\ \hline
\multicolumn{1}{|c|}{\multirow{9}{*}{(6,10)}} & \multirow{2}{*}{(10,5)} & 0.95 & \textbf{6.31e-05} & \textbf{6.31e-05} & 1.82e+01 & 4.32e+01 & 4.76e+01 \\
\multicolumn{1}{|c|}{} &  & 0.99 & \textbf{3.61e-05} & \textbf{3.61e-05} & 2.19e+01 & 4.32e+01 & 4.76e+01 \\
\multicolumn{1}{|c|}{} & \multirow{2}{*}{(10,20)} & 0.95 & \textbf{8.98e-07} & \textbf{8.98e-07} & 2.36e+01 & 1.55e+02 & 1.67e+02 \\
\multicolumn{1}{|c|}{} &  & 0.99 & \textbf{1.29e-06} & \textbf{1.29e-06} & 3.51e+01 & 1.53e+02 & 1.67e+02 \\
\multicolumn{1}{|c|}{} & \multirow{2}{*}{(100,5)} & 0.95 & \textbf{1.60e-14} & \textbf{5.32e-15} & 8.70e+00 & 3.49e+01 & 4.76e+01 \\
\multicolumn{1}{|c|}{} &  & 0.99 & \textbf{1.95e-14} & \textbf{1.08e-10} & 7.96e+00 & 3.18e+01 & 4.76e+01 \\
\multicolumn{1}{|c|}{} & \multirow{2}{*}{(100,20)} & 0.95 & \textbf{2.85e-15} & \textbf{8.04e-14} & 1.47e+01 & 1.21e+02 & 1.67e+02 \\
\multicolumn{1}{|c|}{} &  & 0.99 & \textbf{1.04e-13} & \textbf{1.30e-14} & 1.47e+01 & 1.13e+02 & 1.67e+02 \\ \cline{2-8}
\multicolumn{1}{|c|}{} & \multicolumn{2}{c|}{Sub-group average} & \textbf{1.27e-05} & \textbf{1.27e-05} & 1.81e+01 & 8.68e+01 & 1.07e+02 \\ \hline
\multicolumn{3}{|c|}{Overall average} & \textbf{8.06e-06} & \textbf{8.06e-06} & 2.46e+01 & 8.15e+01 & 1.06e+02 \\ \hline
\end{tabular}
\end{table}

\subsubsection{Discussion on Sample Average Approximation}
\label{ec: saa}

Throughout the paper we assume that the expectation appearing in the constraints of each ALP and
RLP, namely $\mathbb{E}_{s'\sim p(\cdot\mid s,a)}[(\Phi w)(s')]$, can be evaluated exactly. This assumption is satisfied whenever
the transition kernel has modest support, so that the expectation reduces to a
finite sum over a manageable number of successor states. When the kernel is
dense, exact evaluation is no longer practical, and the
expectation must instead be approximated. The standard remedy is sample average
approximation (SAA), in which the expectation is replaced by an average over
independent draws of the successor state from $p(\cdot \mid s,a)$. This
substitution is
also standard practice in this literature: \citet{pakiman2025selfguided}
approximate the expectations in their self-guided ALP formulations in precisely
this manner.

Two features of our formulation make this approximation particularly well
behaved. First, the constraints depend on the transition kernel only through
expectations of bounded basis functions. It is therefore unnecessary to estimate
the kernel itself, which would require a sample size growing with the size of the
state space. It suffices to estimate a finite collection of scalar expectations,
and the sample size required to do so accurately is independent of the size of
the state space. This distinction was established for Markov decision processes
by \citet{kearns2002sparse}, who show via a standard concentration argument that
a sample may approximate the successor-state distribution poorly and nonetheless
yield an accurate estimate of the expectation of any fixed bounded function under it.
Second, because our value function approximation is linear in its weights, these
expectations do not depend on the decision variables. The sample may therefore be
drawn once, prior to solving the linear program, and only finitely many
expectations need to be estimated.

We do not analyze sample average approximation rigorously within our framework.  Quantifying the basis-generating operator and the scenario certificate under SAA is a separate undertaking that lies beyond the scope of this study and that the available space does not permit. The approximation itself, however, has been studied extensively, and the results most relevant to our setting are summarized below.

The error introduced by sample average approximation has been examined from several
directions. The analysis of \citet{petrik2009constraint} is closest to our
setting, in two respects. First, they study precisely the substitution described
above: an approximate linear program in which the constraint coefficients are
estimated from sampled transitions rather than computed exactly. They bound the
resulting error by combining a concentration inequality for the estimated
coefficients with a sensitivity argument for the linear program, and obtain a
bound on the deviation between the solutions of the exact and sampled programs
that depends on the number of samples per constraint, the number of basis
functions, and the discount factor, but not on the size of the state space. Second, their development is also
organized around an expansion, and the contrast with ours is instructive. They
enlarge the constraint set by replacing single-step constraints with multi-step,
or rolled-out, constraints. Because a generative model ordinarily produces only
single-step transitions, their expansion requires multi-step transition
quantities to be estimated from sampled trajectories, and the sampling burden
accordingly grows with the length of the roll-out. Our expansion instead enlarges
the approximation architecture while leaving the constraints intact, so every
expectation remains a single-step quantity however far the basis is expanded.
\citet{petrik2009constraint} further propose a relaxed formulation that permits a
limited number of constraints to be violated, thereby reducing the sensitivity of
the solution to any individual constraint.

Complementary results are available elsewhere in the literature.
\citet{azar2013minimax} establish matching upper and lower bounds on the number
of transition samples required under a generative model, showing that the
dependence on the target accuracy is minimax optimal and cannot be improved.
\citet{haskell2016empirical} study the same substitution within the dynamic
programming operator, developing a random-operator framework that yields
finite-sample guarantees for the resulting value and policy iteration schemes.
More broadly, the convergence of SAA is a foundational result in stochastic
programming, where the optimal solutions of the sampled problem are known to
converge to those of the true problem at an exponential rate in the sample size
\citep{kleywegt2002sample}. In implementation, the sample size may be selected in
the same manner as the number of sampled constraints, by increasing it until the
optimal objective value ceases to change appreciably. The realized approximation
error may additionally be certified from the sample itself using a
variance-adaptive concentration bound \citep{maurer2009empirical}.

\newpage
\putbib[references]
\end{bibunit}

\end{document}